\documentclass[11pt]{article}

\usepackage[preprint]{acl}

\usepackage{times}
\usepackage{latexsym}

\usepackage[T1]{fontenc}

\usepackage[utf8]{inputenc}

\usepackage{microtype}

\usepackage{inconsolata}

\usepackage{graphicx}
\usepackage{tcolorbox}
\usepackage{soul}
\usepackage{xcolor}
\usepackage{enumitem}
\tcbuselibrary{breakable}
\usepackage{longtable}
\usepackage{subcaption}
\usepackage{caption}
\usepackage{svg}
\usepackage{amsmath}
\usepackage{tabularx}
\usepackage{array}
\usepackage{xurl}
\usepackage{fvextra}
\usepackage{placeins}
\usepackage{algorithm}
\usepackage{algpseudocode}
\usepackage{listings}
\usepackage{multirow}
\usepackage[table]{xcolor}
\usepackage{float}
\usepackage{needspace}
\usepackage[T1]{fontenc}
\usepackage{booktabs}

\lstdefinestyle{python}{
    language=Python,
    basicstyle=\ttfamily\scriptsize,
    keywordstyle=\color{blue},
    stringstyle=\color{red!70!black},
    commentstyle=\color{green!50!black},
    showstringspaces=false,
    breaklines=true,
    frame=single,
    columns=fullflexible
}

\usepackage{caption}

\DeclareCaptionType{schema}[
  Schema
][List of Schemas]

\DefineVerbatimEnvironment{TemplateVerb}{Verbatim}{
  fontsize=\scriptsize,
  breaklines=true,
  breakanywhere=true
}

\newcommand\HotpotQA{{\texttt{HotpotQA}}}
\newcommand\BrowseCompPlus{{\texttt{BrowseComp-Plus}}}

\title{\textbf{DeepStress}: Stress-Testing Deep Search Agents}

\author{
  \textbf{Ismaël Rousseau\textsuperscript{1,2}},
  \textbf{Géraldine Damnati\textsuperscript{1}},
  \textbf{Frédéric Béchet\textsuperscript{2}}
  \\
  \textsuperscript{1}Orange Research, Lannion, France
  \\
  \textsuperscript{2}Aix Marseille Univ, CNRS, LIS, Marseille, France
  \\
  \texttt{\{first.last\}@orange.com},
  \texttt{\{first.last\}@lis-lab.fr}
}

\begin{document}
\maketitle
\begin{abstract}
While search agents demonstrate impressive capabilities in multi-step question answering, their robustness to poor-quality evidence remains under-explored. Such evidence is rare in benchmark corpora, and resulting failures are hard to reproduce. We propose \textbf{DeepStress}, a dynamic stress-testing framework that subjects agents to degraded evidence at controlled rates. It replaces the retrieval module with a synthetic environment where documents are generated dynamically for each query, under independent control of their trustworthiness, relevance, and factuality. To characterize behavior under stress, we go beyond binary accuracy with a five-way classification of final-answer outcomes and two new reliability-aware metrics, RAS and CoPRAS, which credit cautious behavior and account for computational cost. Evaluating 12 systems on questions from \HotpotQA~ and \BrowseCompPlus, we find that several open-weight agents answer confidently when sources contain obvious credibility red flags, and that many \HotpotQA~questions are answered from parametric knowledge even when all returned documents are factually wrong.
\end{abstract}

\section{Introduction}
In Retrieval-Augmented Generation (RAG), a model first retrieves passages relevant to the user query, then conditions its answer on the retrieved evidence. 
\emph{Search agents} extend this paradigm by making retrieval iterative: they interleave reasoning steps with calls to external search tools, gathering evidence before producing a final answer. Their performance depends not only on the language model used for synthesis, but also on how evidence is retrieved, how reliable it is, and how the agent adapts its search and reasoning strategy to the evidence it receives. However, standard search-agent evaluations often reduce performance to final-answer accuracy, hiding qualitatively different failure modes such as poor retrieval, misleading evidence, or ineffective search behavior.

Recent work, such as \BrowseCompPlus{} \cite{chen2025browsecompplus}, decouples retriever and model performance by utilizing a curated corpus where questions are paired with supportive and hard-negative documents. However, robustness failures are typically observed \textit{post-hoc}, and the low-signal conditions are difficult to reproduce. 
Our study focuses on how search agents behave when the quality of retrieved evidence is explicitly controlled: we analyze fine-grained final-answer outcomes and complement them with exploratory analyses of search and reasoning behavior.
Inspired by \textit{stress-testing} methodologies, we introduce \textbf{DeepStress}, a simulation environment designed to evaluate the robustness of deep search agents. Much like simulated environments used for autonomous vehicles, where extreme 
conditions are generated to test resilience, \textbf{DeepStress} places search agents in challenging scenarios where search results can be severely degraded. It intercepts the agent's search calls and replaces the retrieval module with a synthetic environment, allowing us to precisely manipulate the properties of the returned documents
along three
axes: \textit{trustworthiness}, \textit{relevance}, and \textit{factuality}.
\textbf{DeepStress} is not a replacement for existing benchmarks but acts as a controlled laboratory for stress testing of search-agent behavior with the following contributions:
\begin{enumerate}
    \item We introduce \textbf{DeepStress}, a controlled simulation stress test environment that evaluates search agents dynamically on
    synthetic documents modified along three axes: \emph{trustworthiness}, \emph{relevance}, and \emph{factuality}.
    \item We evaluate several agents across multiple experimental conditions and document substantial behavior differences when facing degraded documents and propose a novel score called RAS (\textit{Reliability-Aware Score}) that we evaluate against computational costs.
\end{enumerate}
\section{Related Work}

\paragraph{Search agents.}
Early work such as ReAct~\cite{yao2022react} and IRCoT~\cite{trivedi2023interleaving} showed that off-the-shelf LLMs can be prompted to interleave reasoning with external actions or retrieval, enabling multi-step information seeking without task-specific training. More recent work trains search behavior directly. Search-R1~\cite{jin2025searchr1trainingllmsreason}, for instance, uses reinforcement learning (RL) on multi-hop question-answering (QA) tasks, with rewards derived from final-answer correctness, to encourage models to issue useful search queries during reasoning. ZeroSearch~\cite{sun2025zerosearch} further removes the dependence on a live search engine during training by using a fine-tuned language model to generate documents that mimic search-engine outputs.
A parallel line of work targets \emph{deep-research agents}, which produce longer, citation-rich answers over multiple rounds of retrieval and synthesis. DR-Tulu~\cite{shao2026dr}, for example, uses evolving rubrics as RL rewards to improve long-form research answers.
These works focus on improving search-agent capabilities under a given retrieval process. In contrast, our goal is to evaluate how search agents behave under a controlled retrieval environment.

\paragraph{Search benchmarks.}
Search agents are commonly evaluated on their ability to gather relevant evidence through iterative reasoning and search. Multi-hop QA datasets require models to locate and combine evidence about multiple entities before answering. \HotpotQA{}~\cite{yang2018hotpotqa} introduced a diverse set of questions over Wikipedia entries, together with supporting sentence-level facts required for reasoning over the question. More recent benchmarks such as \texttt{BrowseComp}~\cite{weibrowsecomp} introduce harder questions answerable through web browsing, where the answer is short and easily verifiable but difficult to find, requiring persistent search over entangled constraints. \BrowseCompPlus~\cite{chen2025browsecompplus} further improves experimental control by using a fixed corpus with human-verified supporting documents and mined challenging negatives, allowing us to disentangle the retriever effect from that of the LLM.
Our work follows this line of controlled evaluation, but isolates a different factor: how agents react to changes in the quality of retrieved evidence. 

\paragraph{Adversarial environments.}
Recent work evaluates agents by perturbing the external observations they receive.
AgentNoiseBench~\cite{wang2026agentnoisebench} directly replaces or corrupts tool
outputs, while SafeSearch~\cite{dong2026safesearch} and
SearchGEO~\cite{chen2026much} expose search agents to controlled unreliable or
manipulated search results. 
WARP~\cite{zhang2026deepresearchagentspoisonedusergenerated}
poisons existing user-generated web pages, exploiting their repeated retrieval by
deep-research agents to steer their reports. Unlike these attack-oriented evaluations, DeepStress uses controlled semantic perturbations to study how search agents respond to changes in evidence trustworthiness, relevance, and factuality.

\paragraph{Beyond final-answer accuracy.}
Search agents can fail in many different ways:
an agent may give a wrong answer, express uncertainty, fail to produce a final answer, or exhaust its tool-call budget. Recent work on agent evaluation has argued for more detailed failure analysis, including taxonomies of reasoning, planning, tool-use, and system-execution errors~\cite{deshpande2025trail}.
Beyond final outcomes, the trajectory itself is also informative, as highlighted by TRACE~\cite{kim2026beyond}.
Prior work has studied search trajectories through query reformulation and evidence reuse~\cite{ning2026agentic}, and has shown that additional search calls can be helpful, wasteful, or harmful depending on retrieval conditions~\cite{shideepdiver,xie-etal-2026-searching}. Our work combines these two views: we report fine-grained final-answer outcomes and analyze whether agents detect and verbalize degradations in retrieved documents.

\paragraph{Robustness to knowledge conflict.}
By manipulating the quality of retrieved documents, our setting creates controlled cases where agents must reason from evidence that may be unreliable or conflicting, connecting our work to prior studies on knowledge conflict. 
DRAGged into Conflicts~\cite{cattan2025dragged} introduces a taxonomy of conflict situations in RAG, together with the expected model behavior for each category.
TRACK~\cite{feng-etal-2026-tracking} studies how LLMs propagate newly provided facts through multi-step reasoning when these facts conflict with the model's initial parametric knowledge. 
We differ from these studies by evaluating active search agents rather than static RAG or in-context reasoning pipelines. Instead of only asking whether a model can resolve a conflict once the evidence is given, we study how agents behave while searching under controlled degradations of the retrieved documents.
\section{Methodology}
\subsection{Simulation environment}

The core of the \textbf{DeepStress} methodology is to replace the standard search tool with a simulated environment. Documents are dynamically generated for each agent query rather than precomputed, since different agents, or the same agent under different conditions, may follow different search trajectories for the same question. This ensures that each simulated search result is tailored to the agent's actual query while preserving the controlled reliability attributes imposed by the scenario. Returning one controlled document per call differs from traditional retrieval settings, but is appropriate for our laboratory experiment, as it provides finer control over the evidence shown to the agent and isolates agent behavior from retrieval variability.

\vspace{0.2cm}
Let \(\{D_1, D_2, \ldots, D_n\}\) be the set of dimensions under study. A scenario \(\gamma\) specifies a probability for each dimension, chosen by the experimenter:

\vspace{-0.4cm}
\begin{equation}
\boldsymbol{\pi}_\gamma =
\bigl(\pi_\gamma(D_1), \ldots, \pi_\gamma(D_n)\bigr)
\end{equation}
\vspace{-0.6cm}

For each simulated search call, we sample one binary label per dimension from the scenario probabilities. These labels define the target properties of the returned document. In the experiments reported in this paper, they are translated into dimension-specific prompt instructions, although the framework itself is not tied to prompt-based generation.

 For example, if \(D_k\) corresponds to factuality and \(\pi_\gamma(D_k)=80\%\), then the document is sampled as \textit{factually correct} with probability 80\% and \textit{factually incorrect} with probability 20\%.
 In our implementation\footnote{The code for this flexible environment is publicly available and the repository link will be provided after  peer-review.
}, this label determines whether the dynamic prompt includes supporting evidence or factual errors, both for extreme values \(\pi_\gamma(D_k)=0\%\) or \(100\%\) or more balanced situations. 


\subsection{Simulating document reliability}
In this paper, we focus on a controlled experimental setup where we manipulate three source attributes: 

\textbf{Trustworthiness} (T): whether the source appears credible based on its style, structure and URL, regardless of its factuality; \textbf{Relevance} (R): whether the source meaningfully addresses the model-generated query rather than covering unrelated content; \textbf{Factuality} (F): whether the claims contained in the source are factually correct.



\vspace{0.1cm}

We generate documents with an LLM, using the sampled values of \(T\), \(R\), and \(F\) to control the output. For each search call, the simulator generates one synthetic document with a title and URL; its length is chosen to approximate a Wikipedia paragraph. The agent's search query is passed to the LLM as the query the document should address, while the sampled labels are converted into dimension-specific prompt instructions. Full prompt construction, generation details as well as document examples are provided in Appendix~\ref{sec:synthetic_document_generation}.

\paragraph{\textit{Factually correct documents} ($F=1$),} are grounded in dataset-provided evidence: gold Wikipedia paragraphs for \HotpotQA{} and human-verified evidence documents for \BrowseCompPlus{}. This reduces accidental errors on details such as dates or locations. \textit{Factually incorrect} documents ($F=0$) are generated with instructions to introduce severe factual errors.

\paragraph{\textit{Trustworthy} documents (T=1),} are generated with a plausible reliable URL and page title. For \textit{low-trust} results (T=0), we sample a domain from the OpenSources list\footnote{\scriptsize{\url{https://github.com/OpenSourcesGroup/opensources}}} of misleading and outright fake websites, generate a plausible page title for that domain, and include specific instructions in the content generation prompt.

\paragraph{\textit{Irrelevant} documents (R=0),} are generated by replacing the generation target with an unrelated question sampled from the dataset and use its associated supporting context. To mimic a realistic partial web search, the query passed to the generator is a document title sampled from that unrelated support context. 
This produces a coherent and grounded document that is useful for another information need, but irrelevant to the original question asked to the agent.

\vspace{-0.1cm}
\subsection{Sanity check}
\vspace{-0.1cm}

Because our framework relies on synthetic documents generated under controlled conditions, we conduct a human evaluation to verify that the intended labels are reflected in the resulting documents.
We manually annotate a random sample of 50 documents with three annotators\footnote{Examples are provided in Appendix~\ref{sec:data_generation_examples_debussy} and details of the  protocol and full results in Appendix~\ref{sec:human_eval}}. Agreement with the intended labels is high for \textit{trustworthiness} (96.4\%), with strong inter-annotator agreement. Agreement remains strong but is lower for \textit{factuality} (85.5\%), suggesting that this attribute is more difficult to assess consistently for human annotators: a qualitative feedback further suggests that the generated factual errors can be easy to recognize with prior knowledge of the topic in some cases, while remaining plausible to non-specialists for other queries.
We did not ask annotators to check for \textit{relevance} as this dimension is accurate by construction by selecting irrelevant documents from human-verified evidence from the datasets.
Although limited in size, this manual sanity check validates our generation methods, showing that documents generally reflect their intended attributes.



\subsection{Metrics}
\label{sec:metrics}
In order to verify how agents react under degraded conditions, we use several metrics:

\vspace{-0.2cm}

\paragraph{Final-answer outcomes.}
Let \(\mathcal{D}\) be a set of evaluated runs. Each run \(i \in \mathcal{D}\) is associated with a final answer \(a_i\) and a document-quality condition \(\gamma_i\).
We analyze final answers through five outcomes, going beyond a binary correct/incorrect evaluation:
\vspace{-0.2cm}
\begin{itemize}
    \item \textbf{Correct} \((C_i)\): the final answer is factually equivalent to the ground truth. \vspace{-0.2cm}
    \item \textbf{Incorrect} \((I_i)\): the model gives a confident but factually wrong answer. \vspace{-0.2cm}
    \item \textbf{Abstention} \((A_i)\): the model states that it doesn't know or have enough information. \vspace{-0.2cm}
    \item \textbf{Exceed tool budget} \((B_i)\): the agent reaches its tool-call  limit before answering. \vspace{-0.2cm}
    \item \textbf{Missing answer tag} \((M_i)\): the run terminates without a closing answer tag (\textit{e.g.} formatting error or premature \texttt{<eos>} token) \vspace{-0.2cm}
\end{itemize}

Runs that exceed the tool budget or miss the answer tag are assigned directly to their corresponding category. For runs that contain a final answer, we use \texttt{GPT-4.1} as an LLM-as-a-judge to classify the answer as \(C_i\), \(I_i\) or \(A_i\). We adapt the original \texttt{BrowseComp} evaluation prompt to include the abstention category\footnote{Additional evaluation details are provided in Appendix~\ref{sec:llm_judge_config}}.

These outcome categories provide a detailed view of how different systems succeed and fail, but they do not directly yield a single score for ranking systems. We 
complement them with aggregate metrics such as TPC to measure cost efficiency and novel metrics called RAS and CoPRAS.

\vspace{-0.2cm}
\paragraph{Tokens Per Correctness (TPC).}
\citet{xie-etal-2026-searching} define the cost of run \(i\) as a function of the number of generated tokens \(g_i\) and input tokens \(x_i\), and the sequence of search calls \(S_i\):

\vspace{-0.4cm}
\begin{equation}
\mathrm{Cost}_i = g_i + \lambda x_i + \mu \lvert S_i \rvert,
\end{equation}
\vspace{-0.6cm}

where,  \(\lambda=0.25\) and \(\mu=500\) weight the relative cost of input tokens and search calls.
They introduce TPC, which estimates how much computation cost is spent per correct answer:
\begin{equation}
\mathrm{TPC}(\mathcal{D}) =
\frac{\sum_{i\in\mathcal{D}} \mathrm{Cost}_i}
{\sum_{i\in\mathcal{D}} C_i}.
\end{equation}

\paragraph{Reliability-Aware Score (RAS).}
We address this distinction by introducing RAS, 
which accounts for the reliability of the evidence available to the agent when evaluating its final outcome.
In our setting, every question is answerable in principle, but the simulator can make the available evidence unreliable or uninformative. In the extreme case where \textit{relevance} or \textit{factuality} is set to 0\%, the agent should not be expected to recover the answer from the provided documents. RAS therefore gives full credit to correct answers, no credit to incorrect or malformed answers, and \textbf{partial credit} to abstention-like behavior under unreliable conditions.

For a given scenario \(\gamma\), under conditions \(\{\pi_\gamma(T), \pi_\gamma(R), \pi_\gamma(F)\}\) for respectively  trustworthiness, relevance, and factuality, we observe the realized rate as the real proportion of documents for each condition: \(\{\hat{\pi}_\gamma(T), \hat{\pi}_\gamma(R), \hat{\pi}_\gamma(F)\}\). We define the reliability for a run \(i\) under condition \(\gamma_i\) as the product:
\begin{equation}
\rho_{\gamma_i} = \hat{\pi}_{\gamma_i}(T)\times\hat{\pi}_{\gamma_i}(R)\times\hat{\pi}_{\gamma_i}(F).
\end{equation}


The reliability-aware score (RAS) for run \(i\) is:
\begin{equation} 
\mathrm{RAS}_i =
\begin{cases}
1, & C_i, \\
1 - \rho_{\gamma_i}, & A_i \vee B_i, \\
0, & I_i \vee M_i.
\end{cases}
\end{equation}


Correct answers always receive full credit, while incorrect and malformed
answers receive none. Explicit abstention and tool-budget exhaustion receive
partial credit when reliable evidence is unlikely. Of course these two last behaviors are not equivalent and are reported separately in the final-answer outcomes, however grouping them in RAS is a
simplifying design choice which allows us to compare systems at different tool-budgets. Notably, a system that always abstains would get a RAS score of $1-\rho_\gamma$: it reaches the maximum RAS when documents reliability is at 0, but gets a null score under fully reliable evidence. 

Including trustworthiness in $\rho_\gamma$ reflects
the epistemic perspective adopted in this work: an agent should not be required
to rely on a source that appears unreliable, even when its content happens to
be factually correct. RAS still gives full credit when such evidence is used
successfully, but also treats refusal as acceptable when observable credibility
cues are poor.

\paragraph{Cost per Reliability-Aware Score.}
To obtain an analogous cost-normalized metric, we also define \textit{Cost per Reliability-Aware Score} (CoPRAS):
\begin{equation}
\mathrm{CoPRAS}(\mathcal{D}) =
\frac{\sum_{i\in\mathcal{D}} \mathrm{Cost}_i}
{\sum_{i\in\mathcal{D}} \mathrm{RAS}_i}. 
\end{equation}
If the denominator is zero, CoPRAS is set to \(+\infty\). 
As with any aggregate metric, RAS and CoPRAS reflect a particular scoring choice; we discuss alternative choices and weightings in Appendix~\ref{sec:aggregate_metrics_discussion}.

\vspace{-0.1cm}
\section{Experimental Protocol}
\vspace{-0.1cm}

\subsection{Datasets}
We draw questions from two QA datasets that differ in both question difficulty and exposure to LLM pretraining. \textbf{\HotpotQA{}} questions typically require a few reasoning hops over Wikipedia passages and are answerable from a small set of supporting paragraphs. \textbf{\BrowseCompPlus{}}, derived from \texttt{BrowseComp}, instead contains queries constructed by inversion from a target entity under multiple constraints; answering them requires iterative search and aggregation across many documents, and state-of-the-art retrievers achieve low recall on its corpus of adversarially mined distractors.

The two datasets are also complementary in their dependence on supporting documents. \HotpotQA{} answers often involve general knowledge that models may store in their parameters, and the corpus itself has likely been seen during pretraining. By controlling the factuality of the supporting documents, we can therefore study how a model handles conflicts between its parametric knowledge and the dynamic knowledge returned by search. \BrowseCompPlus{} questions, in contrast, cannot be answered without supporting documents, as the corpus is too recent to be known by current LLMs. With no such conflict, we can study instead how contradictory supporting documents affect the reasoning strategy of a search agent.



\vspace{-0.2cm}
\subsection{Evaluated systems}
\vspace{-0.1cm}
We have selected systems to cover a diverse set of search-agent designs: closed-weight and open-weight models, different model sizes, and different training procedures such as supervised fine-tuning and RL under different reward objectives.
A system denotes a model together with its harness, \textit{i.e.}, the orchestration code and prompts that determine how context is assembled, how tools are invoked, and how model outputs are parsed. To keep the comparison controlled, simulated documents are formatted according to the conventions expected by each harness\footnote{Additional implementation details, including model versions and references, tool-use templates, and training procedures, are provided in Appendix~\ref{sec:systems_implementation_technical_details}.}. 
In total, twelve systems are evaluated.
\textbf{GPT systems} (\texttt{gpt-5-nano} and \texttt{gpt-5-mini}), each run with \texttt{low}, \texttt{medium}, and \texttt{high} reasoning effort, provide six closed-weight tool-calling baselines.
\textbf{SpiqaDR}\footnote{URL will be provided after peer review.} is a 3B Qwen2.5-based search agent trained with RL
for scientific literature search; a detailed description is given in Appendix~\ref{sec:spiqa_training}.
\textbf{Search-R1}~\cite{jin2025searchr1trainingllmsreason} is a 7B Qwen2.5-based search agent trained with RL to interleave reasoning with multi-turn search.
\textbf{AutoRefine}~\cite{shi2026search} is a 3B Qwen2.5-based retrieval-augmented reasoning model that alternates search with explicit evidence-refinement steps before answering.
\textbf{ASearcher}~\cite{gao2026unlocking} is a 7B Qwen2.5-based search agent trained with asynchronous RL for long-horizon information seeking; we use its local-RAG variant rather than the web-search version.
\textbf{DR-Tulu}~\cite{shao2026dr} is an 8B open deep-research model trained with RL and evolving rubrics to produce long-form, citation-grounded answers.
We also evaluate \textbf{DR-Tulu (no reasoning)}, which uses the same checkpoint with a modified orchestration wrapper: whenever the model opens a \texttt{<think>} block, the wrapper immediately closes it, preventing explicit reasoning while keeping the same tool-use and answer format.
\vspace{-0.1cm}
\subsection{Experimental setup}
\begin{figure*}[t]
    \centering

    \captionsetup[subfigure]{skip=0pt}

    \includegraphics[
        width=\textwidth,
        height=1.2em,
        keepaspectratio
    ]{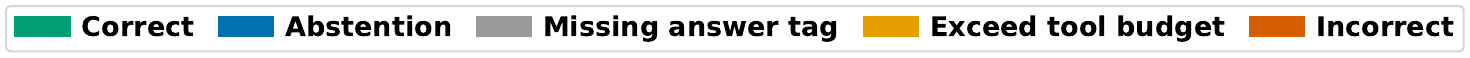}
    
    \vspace{0.2em}

    \begin{minipage}[t]{0.31\textwidth}
        \centering
        \includegraphics[width=\linewidth]{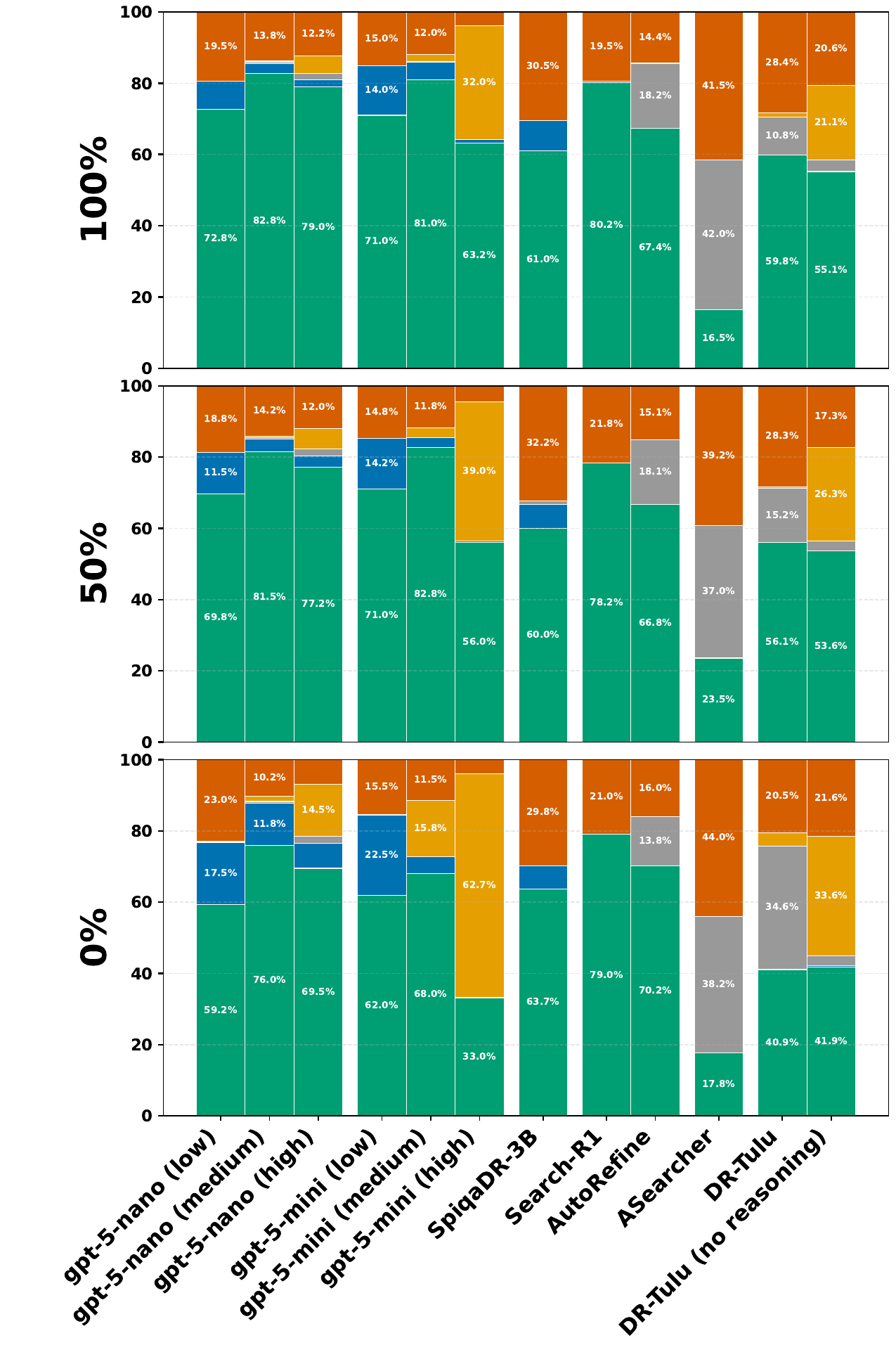}
    \end{minipage}
    \hfill
    \begin{minipage}[t]{0.31\textwidth}
        \centering
        \includegraphics[width=\linewidth]{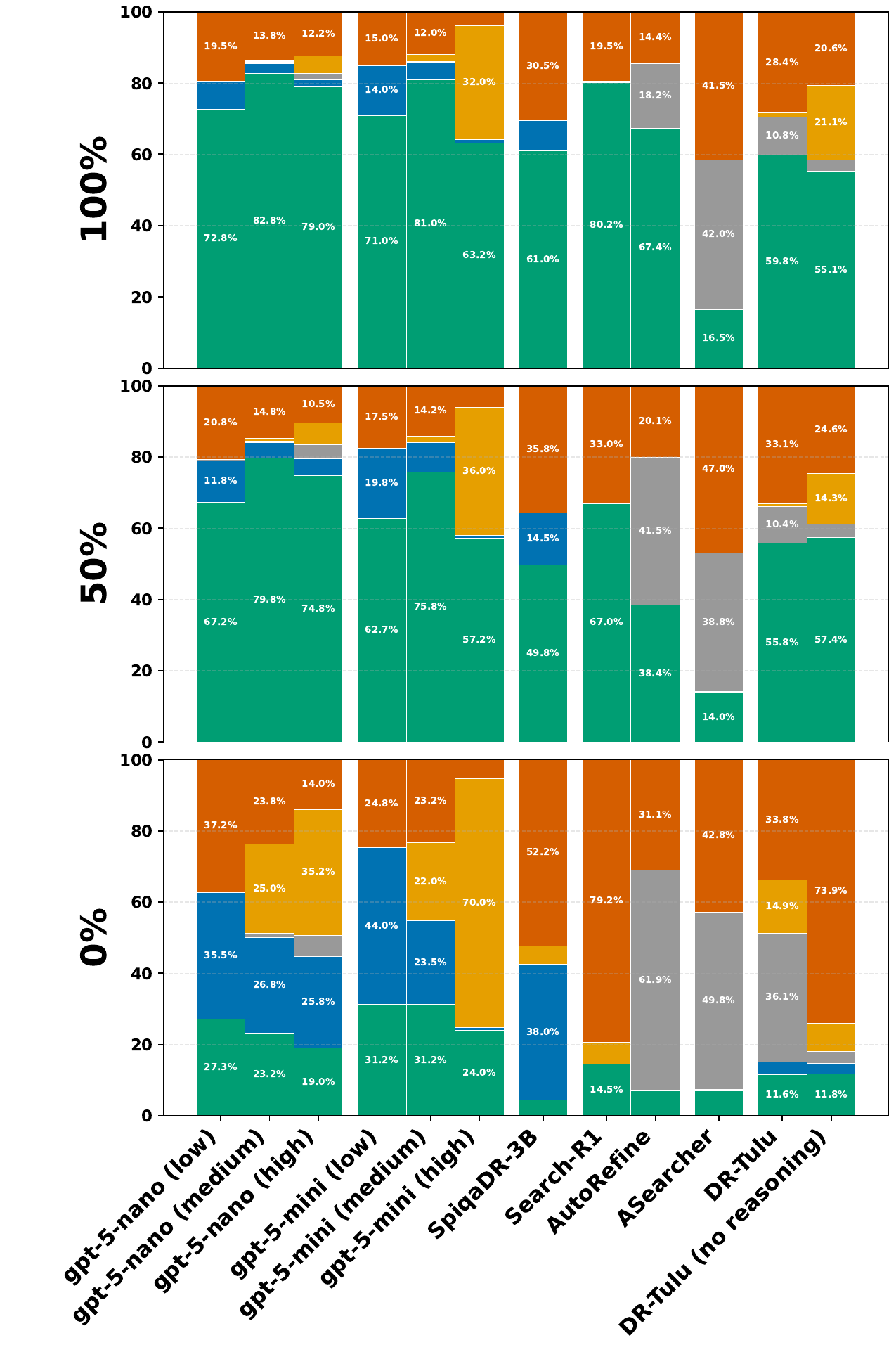}
    \end{minipage}
    \hfill
    \begin{minipage}[t]{0.31\textwidth}
        \centering
        \includegraphics[width=\linewidth]{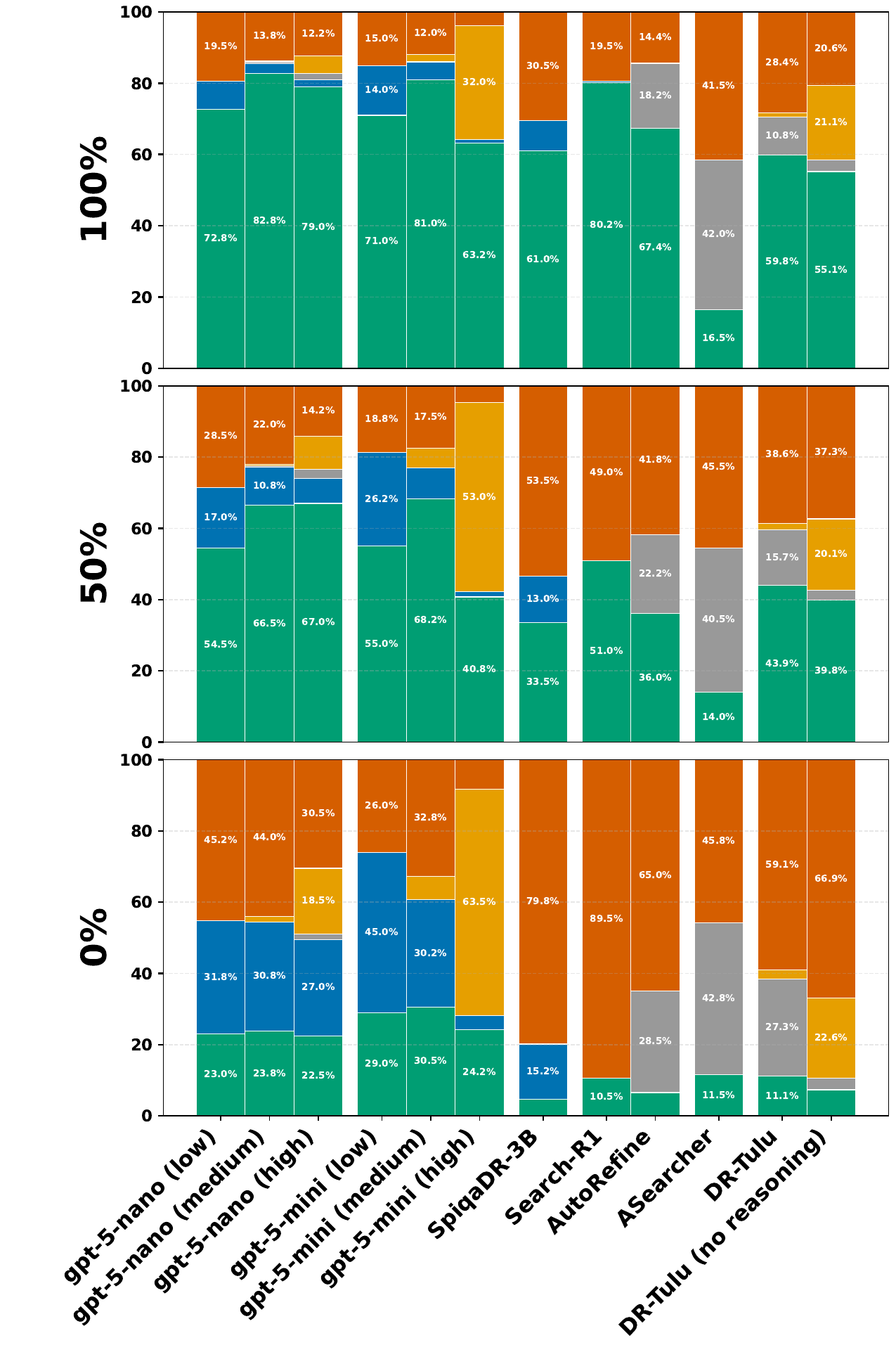}
    \end{minipage}

    \par\vspace{-1.2em}

    \subcaptionbox{
        Trustworthiness
        \label{fig:outcomes_agg_trust}
    }[0.32\textwidth]{}
    \hfill
    \subcaptionbox{
        Relevance
        \label{fig:outcomes_agg_rel}
    }[0.32\textwidth]{}
    \hfill
    \subcaptionbox{
        Factuality
        \label{fig:outcomes_agg_fact}
    }[0.32\textwidth]{}

    \caption{Final-answer outcomes aggregated over \HotpotQA{} and \BrowseCompPlus{}.
    Outcome rates across three quality levels (0\%, 50\%, and 100\%) for each
    document-quality axis: Trustworthiness, Relevance, and Factuality. Raw values and 95\% confidence intervals are reported in Appendix~\ref{sec:confidence_intervals}.}
    \label{fig:aggregated_outcomes}
    \vspace{-0.1cm}
\end{figure*}
\vspace{-0.1cm}


We evaluate 12 systems on \HotpotQA{} and \BrowseCompPlus{}, with a maximum budget of 100 tool calls per question. For each dataset, we randomly sample 200 questions. 
We vary one reliability dimension at a time while keeping the two others fixed at 100\%. For each dimension, we consider three quality levels (0\%, 50\%, 100\%) covering 7 scenarios in total. Having a finer-grained set of quality levels would be interesting but we focus on extreme cases and an intermediate condition.
This design keeps the evaluation cost manageable, balancing the cost of dynamic document generation and agent execution without sacrificing the statistical validity of our robustness analysis.
Overall, our evaluation comprises 12 systems $\times$ 2 datasets $\times$ 200 questions $\times$ 7 scenarios, resulting in 33,600 traces.
We use \texttt{Gemma-4-31B} to generate our synthetic documents. Implementation details for document generation are provided in Appendix \ref{sec:annex_inference_datagen}.
\vspace{-0.1cm}
\section{Results}
\vspace{-0.1cm}
\subsection{Analyzing system behaviors}
\vspace{-0.1cm}
\label{sec:system_behaviors}

\paragraph{Final-answer outcomes.}
Figure~\ref{fig:aggregated_outcomes} summarizes final-answer outcomes by system and degraded document-quality dimension, one at a time, aggregated over both datasets. The top histograms correspond to the fully reliable setting and are repeated for reference.
\vspace{0.1cm}

When degrading \textit{trustworthiness} (Figure~\ref{fig:outcomes_agg_trust}), 
several open-weight search agents, including \texttt{SpiqaDR}, \texttt{Search-R1}, \texttt{AutoRefine}, and \texttt{ASearcher}, show little or no change.
For example, \texttt{Search-R1} remains between 78\% and 80\% correct across all trustworthiness levels, with no clear change in its abstention rate. These models rely heavily on the information provided in their context, even when sources have clear surface cues of low credibility. By contrast, \texttt{DR-Tulu} becomes less reliable as trustworthiness decreases: its correct-answer rate drops from 59.8\% to 40.9\%, while the rate of \textit{no-answer} 
increases from 10.8\% to 34.6\%. This suggests that lower-quality documents can make this system more brittle with respect to following a specified format.

GPT-based systems show a different pattern: when trustworthiness is degraded, they more often avoid giving a final answer, either by abstaining or by exhausting the tool-call budget. For instance, \texttt{GPT-5-mini (low)} increases its abstention rate from 14.0\% to 22.5\%, while \texttt{GPT-5-mini (high)} increases its tool-budget exhaustion rate from 32.0\% to 62.7\%. Overall, even though the low-trust documents contain very clear, sometimes almost caricatural low-credibility signals, all systems still give non-abstaining final answers in many runs at 0\% trustworthiness. This suggests that agents may continue to rely on evidence even when the source clearly looks unreliable.
\begin{table*}[t]
\centering
\scriptsize
\setlength{\tabcolsep}{2pt}
\resizebox{\textwidth}{!}{%
\begin{tabular}{lccccccccccccccc}
\toprule
\textbf{system} & \multicolumn{5}{c}{\textbf{0\%}} & \multicolumn{5}{c}{\textbf{50\%}} & \multicolumn{5}{c}{\textbf{100\%}} \\
\cmidrule(lr){2-6} \cmidrule(lr){7-11} \cmidrule(lr){12-16}
 & \textbf{Cost} & \textbf{Correct} & \textbf{RAS} & \textbf{TPC} & \textbf{CoPRAS} & \textbf{Cost} & \textbf{Correct} & \textbf{RAS} & \textbf{TPC} & \textbf{CoPRAS} & \textbf{Cost} & \textbf{Correct} & \textbf{RAS} & \textbf{TPC} & \textbf{CoPRAS} \\
\midrule
gpt-5-nano (low) & 2.5k & 0.36 & 0.83 & 6.8k & \textbf{3.0k} & 1.6k & 0.64 & 0.78 & 2.5k & \textbf{2.0k} & 1.2k & 0.73 & 0.73 & 1.7k & 1.7k \\
gpt-5-nano (medium) & 22.7k & 0.41 & 0.84 & 55.3k & 27.0k & 10.2k & \textbf{0.76} & 0.81 & 13.4k & 12.6k & 6.5k & \textbf{0.83} & \textbf{0.83} & 7.8k & 7.8k \\
gpt-5-nano (high) & 39.6k & 0.37 & 0.84 & 107.1k & 47.1k & 21.4k & 0.73 & 0.80 & 29.3k & 26.8k & 15.2k & 0.79 & 0.79 & 19.3k & 19.3k \\
gpt-5-mini (low) & 3.0k & 0.41 & 0.90 & 7.3k & 3.3k & 2.1k & 0.63 & 0.80 & 3.4k & 2.7k & 1.7k & 0.71 & 0.71 & 2.4k & 2.4k \\
gpt-5-mini (medium) & 26.4k & \textbf{0.43} & 0.88 & 61.0k & 30.0k & 14.1k & 0.76 & \textbf{0.83} & 18.7k & 17.1k & 8.9k & 0.81 & 0.81 & 11.0k & 11.0k \\
gpt-5-mini (high) & 55.6k & 0.27 & \textbf{0.96} & 205.4k & 58.0k & 41.9k & 0.51 & 0.74 & 81.6k & 56.8k & 34.3k & 0.63 & 0.63 & 54.3k & 54.3k \\
SpiqaDR-3B & 4.5k & 0.24 & 0.48 & 18.6k & 9.5k & 2.1k & 0.48 & 0.55 & 4.4k & 3.8k & 1.7k & 0.61 & 0.61 & 2.7k & 2.7k \\
Search-R1 & 5.2k & 0.35 & 0.37 & 14.9k & 14.0k & 2.9k & 0.65 & 0.65 & 4.5k & 4.5k & 2.8k & 0.80 & 0.80 & 3.5k & 3.5k \\
AutoRefine & \textbf{1.1k} & 0.28 & 0.28 & \textbf{3.8k} & 3.8k & \textbf{1.1k} & 0.47 & 0.47 & \textbf{2.3k} & 2.3k & \textbf{1.1k} & 0.67 & 0.67 & \textbf{1.6k} & \textbf{1.6k} \\
ASearcher & 3.4k & 0.12 & 0.12 & 28.0k & 27.5k & 2.7k & 0.17 & 0.17 & 15.7k & 15.5k & 2.6k & 0.17 & 0.17 & 15.8k & 15.8k \\
DR-Tulu & 29.0k & 0.21 & 0.30 & 138.3k & 97.6k & 12.8k & 0.51 & 0.52 & 24.8k & 24.6k & 10.6k & 0.59 & 0.60 & 17.8k & 17.8k \\
DR-Tulu (no reasoning) & 53.0k & 0.20 & 0.44 & 261.7k & 121.8k & 47.1k & 0.50 & 0.60 & 93.9k & 78.2k & 43.9k & 0.55 & 0.55 & 79.8k & 79.8k \\
\midrule
\textbf{Average} & 20.5k & 0.30 & 0.60 & 75.7k & 36.9k & 13.3k & 0.57 & 0.64 & 24.5k & 20.6k & 10.9k & 0.66 & 0.66 & 18.1k & 18.1k \\
\bottomrule
\end{tabular}
}
\caption{Reliability-aware score (RAS; $\nearrow$), tokens per correctness (TPC; $\searrow$), and cost per RAS (CoPRAS; $\searrow$), aggregated by isolated document quality level. 
}
\label{tab:aggregate_metrics}
\end{table*}


When degrading \textit{relevance} (Figure~\ref{fig:outcomes_agg_rel}), the effect is stronger: the correct-answer rate drops sharply for all systems. Systems also fail in different ways. \texttt{Search-R1} mostly fails by producing factually incorrect answers, with its incorrect-answer rate increasing from 19.5\% to 79.2\%. \texttt{SpiqaDR} shows a mixed failure pattern, with more incorrect answers (from 30.5\% to 52.2\%) and more abstentions (from 8.5\% to 38.0\%). GPT-based models mostly fail by either abstaining or exhausting tool-call budget: low-effort models tend to abstain, while high-effort models tend to keep searching until they reach the budget limit. \texttt{DR-Tulu} fails to produce a valid answer tag as documents become less relevant, whereas \texttt{DR-Tulu without reasoning} mostly outputs factually incorrect answers. This suggests that \texttt{DR-Tulu}'s reasoning trace may help prevent some incorrect final answers, but at the cost of more format failures. \texttt{AutoRefine} also shows a specific brittleness to relevance degradation: its rate of missing answer tags increases from 18.2\% to 61.9\%, a much sharper increase than what we observe for the other degradation axes.


Finally, \textit{factuality} degradation (Figure~\ref{fig:outcomes_agg_fact}) increases the trends seen for relevance degradation.~
\texttt{SpiqaDR} abstains much less at 0\% factuality than at 0\% relevance (15.2\% versus 38.0\%), and \texttt{DR-Tulu} reaches 59.1\% incorrect answers at 0\% factuality, compared with 33.8\% at 0\% relevance. This suggests that irrelevant documents are easier to reject than relevant but false documents.
\vspace{0.1cm}

These differences in behavior suggest that model robustness is heavily influenced by training methods. Specifically, narrow RL-based rewards tend to produce brittle models, whereas the scale and diversity of GPT training enable more robust generalization.

\vspace{-0.1cm}
\paragraph{Aggregate metrics.} 

Table ~\ref{tab:aggregate_metrics} reports aggregate metrics (correctness, RAS, and cost-normalized variants TPC and CoPRAS) at 0\%, 50\%, and 100\% document quality. Each value is averaged across three settings, where one dimension is varied while the others remain at 100\%.
Degrading document quality increases the average cost from 10.9k to 20.5k. This is consistent with \citet{xie-etal-2026-searching}, who also observe that noisier search conditions lead agents to issue more searches, although the effect is highly system-dependent: the cost of \texttt{GPT-5-nano (medium)} increases by a factor of 3.5 between 100\% and 0\% document quality, whereas \texttt{AutoRefine} shows almost no change.

Correctness and RAS give different views of behavior under degradation. At 0\% quality, average correctness drops to 0.30 while average RAS stays at 0.61.
This gap shows that not all incorrect outcomes are equal, some agents avoid unsupported answers when evidence is unreliable. For instance, \texttt{GPT-5-nano (medium)} maintains a stable RAS despite a drop in correctness (0.83 to 0.41). Conversely, \texttt{Search-R1} drops in both correctness (0.80 to 0.35) and RAS (0.80 to 0.37), receiving little reliability credit. Thus, RAS differentiates between robust non-answering and overconfident failure in low-quality settings.



Cost-normalized metrics show that TPC increases sharply from 18.1k to 75.7k as quality drops to 0\%, as it ignores reliability and rewards only correctness. CoPRAS increases less severely (to 36.8k) by incorporating RAS. This shifts the ranking at 0\% quality: \texttt{AutoRefine} is favored by TPC for its low cost, while \texttt{GPT-5-nano (low)} and \texttt{GPT-5-mini (low)} are favored by CoPRAS for combining low cost with robustness. Ultimately, TPC rewards cheap correctness, while CoPRAS rewards cost-efficient reliability.

We provide a more detailed discussion in Appendix~\ref{sec:aggregate_metrics_discussion}, including alternative cost and scoring choices and the relation between maximum tool-call budget and model rankings stability.

\vspace{-0.2cm}
\subsection{Breakdown by dataset}
\label{sec:dataset_analysis}
\vspace{-0.1cm}
The previous analysis aggregates \HotpotQA{} and \BrowseCompPlus{}, but the two datasets reveal very different behaviors under factuality degradation\footnote{A detailed breakdown across all  dimensions is given in Appendix~\ref{sec:detailed_outcomes_dataset}}. 

\begin{figure}[t]
    \centering
    \includegraphics[width=\linewidth]{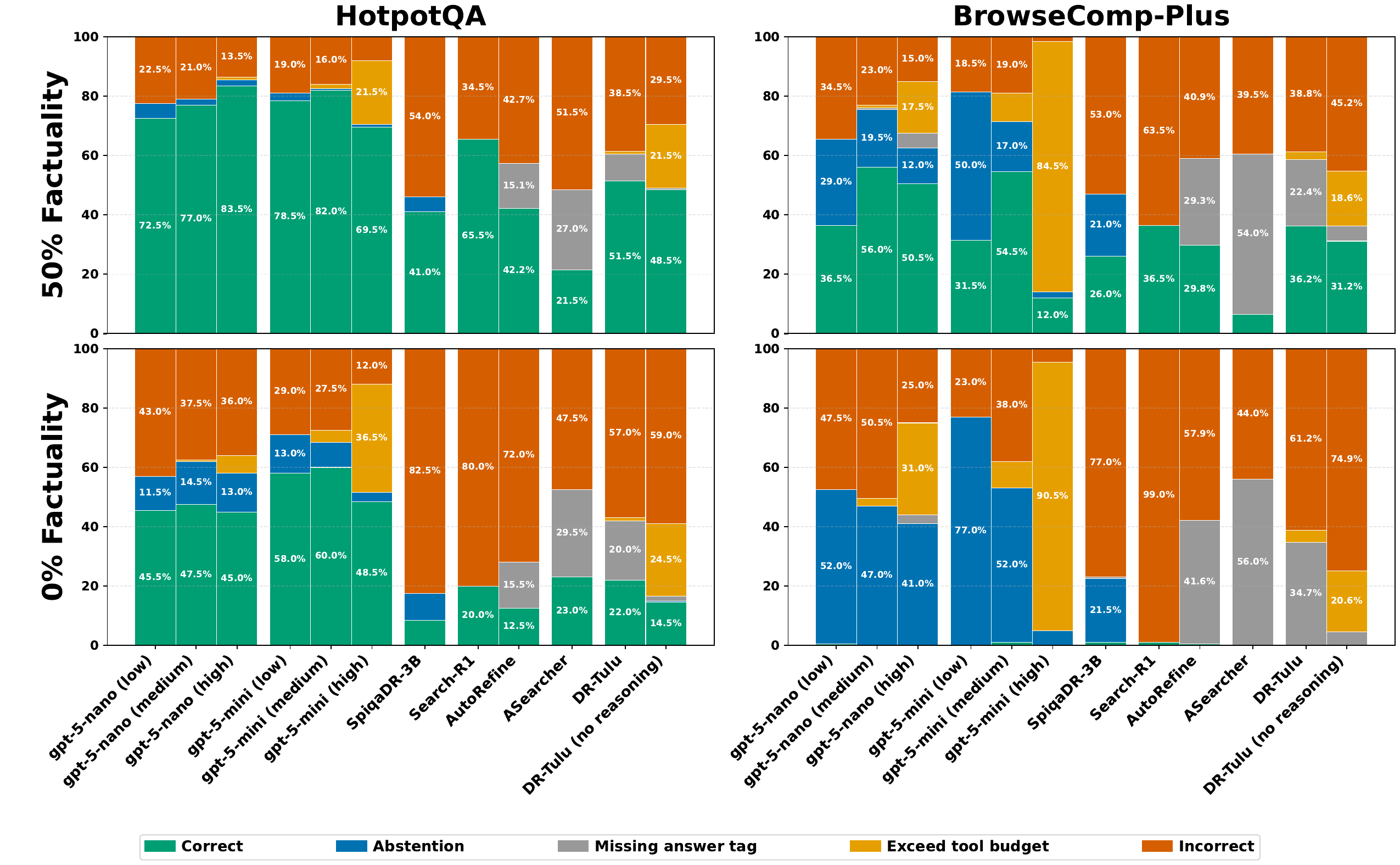}
    \caption{Outcomes at factuality levels 0\% and 50\% on \HotpotQA{} and \BrowseCompPlus{}.}
    \label{fig:datasets_outcomes}
    \vspace{-0.25cm}
\end{figure}

Figure~\ref{fig:datasets_outcomes} compares final-answer outcomes on \HotpotQA{} and \BrowseCompPlus{} under two stress-test settings: 0\% and 50\% factuality.

\paragraph{\textit{0\% factuality}:} in this setting, all relevant documents are incorrect, testing whether models rely on parametric knowledge, abstain, or follow misleading evidence.
Even though the simulator never provides accurate evidence for the target fact, agents still answer many \HotpotQA{} questions correctly. For example, \texttt{gpt-5-mini (medium)} reaches 60\% correctness, and all systems obtain at least 10\%. By contrast, performance on \BrowseCompPlus{} drops almost to zero. This suggests that many \HotpotQA{} questions can be answered from parametric knowledge, while \BrowseCompPlus{} depends more directly on evidence from the context.
To test this hypothesis, we have run a direct-inference baseline in which each model receives only the question, with no search tool and no retrieved documents. The results are reported in appendix~\ref{sec:direct_inference_results} 
confirming the same pattern: all systems answer a non-trivial fraction of \HotpotQA{} questions correctly, while
direct-inference performance on \BrowseCompPlus{} is almost always close to zero.~
As a result, \HotpotQA{} should not be used to evaluate search agents without a direct-inference baseline.

\paragraph{\textit{50\% factuality}:} in this setting, agents receive a mix of correct and incorrect documents to test their ability to resolve conflicting evidence, each search result has a 50\% chance of being generated from the correct supporting context, while the remaining results contain factual errors. Importantly, incorrect documents are not constructed to agree with each other, whereas factual documents are grounded in the same evidence. As a result, an agent that performs multiple searches can in principle identify the stable signal across documents and discount inconsistent false claims. This setting is especially informative on \BrowseCompPlus{}, where agents cannot rely on parametric knowledge. Correct answers in this condition suggest that the agent is able to aggregate evidence across multiple tool calls, even if some may yield contradictory information.

\vspace{-0.2cm}
\section{Conclusion}
\vspace{-0.2cm}

We have introduced \textbf{DeepStress}, a controlled simulation framework for stress-testing Deep Search agents under degraded evidence.
Our experiments reveal substantial differences in agents' robustness, uncertainty expression, and search adaptation, leading to three main conclusions. First, source reliability should be explicitly considered in the design of search agents, so that they can assess retrieved evidence and adapt their behavior accordingly. Second, robustness to conflicts between parametric knowledge and retrieved evidence should be systematically evaluated: producing correct answers despite consistently incorrect retrieved evidence may indicate robustness, but also benchmark contamination.
Third, aggregate measures such as the proposed RAS metrics provide a useful operational characterization of system performance by capturing robustness across different levels of evidence degradation. Finally, although our framework relies on synthetic interventions, the conditions it models reflect genuine deployment risks, as agents operating on the open Web can encounter erroneous or deliberately manipulated content~\cite{zhang2026deepresearchagentspoisonedusergenerated}.
Given these findings, we recommend integrating controlled stress testing into the evaluation framework for Deep Search systems.

Two directions follow from this work.
First, by examining query formulation and reasoning steps, we move toward a process-based evaluation that complements outcome-level results with an analysis of agent search behavior and evidence integration.
Second, \textbf{DeepStress} could also support the training of more robust agents. The brittleness of systems trained with narrow outcome-based rewards motivates reliability-aware training objectives, for which the controlled document labels and RAS could provide a useful starting point.


\section*{Limitations}
Our simulator is designed as a controlled environment for studying how document quality affects agent behavior. Each search call returns one generated document. This lets us control the evidence seen by the agent, but makes the search setting simpler than real search. In particular, we do not model cases where a single search call returns several documents with conflicting, incomplete, or irrelevant content.

We study simple search agents that use a single search loop and one type of search tool. This helps isolate the effect of document quality and keeps the results easy to interpret. However, our findings mainly apply to this type of agent. They may not directly apply to more complex systems that use summarization, reranking, query reformulation, or sub-agents.

We vary the three reliability dimensions independently and use LLM-generated documents. This gives us precise control over each dimension, but we do not study documents where several reliability issues appear at the same time. Synthetic documents may also be cleaner, less noisy, and less diverse than real search results. Our control experiments and human evaluation suggest that the setup is still useful for studying agent behavior, but these documents are not a perfect substitute for real web content.

Finally, our document generation pipeline requires each question to have supporting evidence. We use this evidence to ground factually correct documents, such as Wikipedia paragraphs from HotpotQA or reference documents from BrowseComp-Plus, and reduce hallucinations from the generator. Benchmarks without such evidence can still be used, but their factually correct documents may contain generator errors. This would make the factuality axis noisier and the conclusions less reliable.


\section*{Ethical concerns}
This work is mainly dedicated to evaluation and diagnosis, and is intended to assess the behavior of models with respect to unreliable documents that could be retrieved by search engines. We believe that the subject of the study itself has a significant impact towards trustworthy and explainable Artificial Intelligence. We are aware that for the sake of evaluation we generate unreliable synthetic documents, including untrustworthy and non-factual documents, but these documents are generated dynamically for the sake of the experiment and are not meant to be distributed. When presenting samples of such documents, we will systematically add a warning notice mentioning that they contain unreliable content. We are however also aware that the methodology could be misused with the objective of generating such documents for illicit purposes. We will include a disclaimer on the code repository discouraging such misuse. 
Regarding the human annotation process for evaluating the "quality" of synthetic documents (quality meaning here that they correspond to the purpose they were designed for), the annotators were internal colleagues that contributed on a voluntary basis, and there was no specific compensation. Annotators were fully aware that they were likely to be exposed to unreliable documents. 

\bibliography{custom}

@article{weibrowsecomp,
  title={BrowseComp: A Simple Yet Challenging Benchmark for Browsing Agents},
  author={Wei, Jason and Sun, Zhiqing and Papay, Spencer and McKinney, Scott and Han, Jeffrey and Fulford, Isa and Chung, Hyung Won and Passos, Alex Tachard and Fedus, William and Glaese, Amelia},
  url={https://arxiv.org/abs/2504.12516},
  journal={arXiv preprint arXiv:2504.12516},
  year={2025}
}

@inproceedings{yang2018hotpotqa,
  title={{H}otpot{QA}: A dataset for diverse, explainable multi-hop question answering},
  author={Yang, Zhilin and Qi, Peng and Zhang, Saizheng and Bengio, Yoshua and Cohen, William and Salakhutdinov, Ruslan and Manning, Christopher D},
  url = "https://aclanthology.org/D18-1259/",
  booktitle={Proceedings of the 2018 conference on {E}mpirical {M}ethods in {N}atural {L}anguage {P}rocessing},
  year={2018}
}

@article{cattan2025dragged,
  title={Dragged into conflicts: Detecting and addressing conflicting sources in search-augmented llms},
  author={Cattan, Arie and Jacovi, Alon and Ram, Ori and Herzig, Jonathan and Aharoni, Roee and Goldshtein, Sasha and Ofek, Eran and Szpektor, Idan and Caciularu, Avi},
  url={https://arxiv.org/abs/2506.08500},
  journal={arXiv preprint arXiv:2506.08500},
  year={2025}
}

@inproceedings{shideepdiver,
  title={DeepDiver: Adaptive Web-Search Intensity Scaling via Reinforcement Learning},
  author={Shi, Wenxuan and Tan, Haochen and Kuang, Chuqiao and Li, Xiaoguang and Chen, Hanting and Ren, Xiaozhe and Wang, Yasheng and Hou, Lu and Shang, Lifeng},
  url={https://proceedings.neurips.cc/paper_files/paper/2025/file/180d4373aca26bd86bf45fc50d1a709f-Paper-Conference.pdf},
  booktitle={The Thirty-ninth Annual Conference on Neural Information Processing Systems},
  year={2025}
}

@inproceedings{xie-etal-2026-searching,
    title = "Over-Searching in Search-Augmented Large Language Models",
    author = "Xie, Roy  and
      Gopinath, Deepak  and
      Qiu, David  and
      Lin, Dong  and
      Sun, Haitian  and
      Potdar, Saloni  and
      Dhingra, Bhuwan",
    editor = "Demberg, Vera  and
      Inui, Kentaro  and
      Marquez, Llu{\'i}s",
    booktitle = "Proceedings of the 19th Conference of the {E}uropean Chapter of the {A}ssociation for {C}omputational {L}inguistics",
    month = mar,
    year = "2026",
    address = "Rabat, Morocco",
    url = "https://aclanthology.org/2026.eacl-long.361/",
    doi = "10.18653/v1/2026.eacl-long.361",
    ISBN = "979-8-89176-380-7"
}

@inproceedings{feng-etal-2026-tracking,
    title = "Tracking the Limits of Knowledge Propagation: How {LLM}s Fail at Multi-Step Reasoning with Conflicting Knowledge",
    author = "Feng, Yiyang  and
      Chen, Zeming  and
      Wu, Haotian  and
      Zhou, Jiawei  and
      Bosselut, Antoine",
    editor = "Demberg, Vera  and
      Inui, Kentaro  and
      Marquez, Llu{\'i}s",
    booktitle = "Proceedings of the 19th Conference of the {E}uropean Chapter of the {A}ssociation for {C}omputational {L}inguistics",
    month = mar,
    year = "2026",
    address = "Rabat, Morocco",
    url = "https://aclanthology.org/2026.eacl-long.273/",
    doi = "10.18653/v1/2026.eacl-long.273",
    ISBN = "979-8-89176-380-7",
}

@misc{jin2025searchr1trainingllmsreason,
  title={Search-R1: Training LLMs to Reason and Leverage Search Engines with Reinforcement Learning},
  author={Jin, Bowen and Zeng, Hansi and Yue, Zhenrui and Yoon, Jinsung and Arik, Sercan O and Wang, Dong and Zamani, Hamed and Han, Jiawei},
  url={https://openreview.net/forum?id=Rwhi91ideu#discussion},
  booktitle={Second Conference on Language Modeling},
  year={2025}
}

@inproceedings{
shao2026dr,
title={{DR} Tulu: Reinforcement Learning with Evolving Rubrics for Deep Research},
author={Rulin Shao and Akari Asai and Shannon Zejiang Shen and Hamish Ivison and Varsha Kishore and Jingming Zhuo and Xinran Zhao and Molly Park and Samuel G. Finlayson and David Sontag and Tyler Murray and Sewon Min and Pradeep Dasigi and Luca Soldaini and Faeze Brahman and Wen-tau Yih and Tongshuang Wu and Luke Zettlemoyer and Yoon Kim and Hannaneh Hajishirzi and Pang Wei Koh},
booktitle={Forty-third International Conference on Machine Learning},
year={2026},
url={https://openreview.net/forum?id=97NEP1pyS3}
}

@inproceedings{
chen2025browsecompplus,
title={BrowseComp-Plus: A More Fair and Transparent Evaluation  Benchmark of Deep-Research Agent},
author={Zijian Chen and Xueguang Ma and Shengyao Zhuang and Ping Nie and Kai Zou and Sahel Sharifymoghaddam and Andrew Liu and Joshua Green and Kshama Patel and Ruoxi Meng and Mingyi Su and Yanxi Li and Haoran Hong and Xinyu Shi and Xuye Liu and Nandan Thakur and Crystina Zhang and Luyu Gao and Wenhu Chen and Jimmy Lin},
booktitle={First Workshop on Multi-Turn Interactions in Large Language Models},
year={2025},
url={https://openreview.net/forum?id=YJAA2PzfDi}
}

@article{ning2026agentic,
  title={Agentic Search in the Wild: Intents and Trajectory Dynamics from 14M+ Real Search Requests},
  author={Ning, Jingjie and Coelho, Jo{\~a}o and Kong, Yibo and Long, Yunfan and Martins, Bruno and Magalh{\~a}es, Jo{\~a}o and Callan, Jamie and Xiong, Chenyan},
  journal={arXiv preprint arXiv:2601.17617},
  url={https://arxiv.org/abs/2601.17617},
  year={2026}
}

@article{sun2025zerosearch,
  title={Zerosearch: Incentivize the search capability of llms without searching},
  author={Sun, Hao and Qiao, Zile and Guo, Jiayan and Fan, Xuanbo and Hou, Yingyan and Jiang, Yong and Xie, Pengjun and Zhang, Yan and Huang, Fei and Zhou, Jingren},
  journal={arXiv preprint arXiv:2505.04588},
  url={https://arxiv.org/abs/2505.04588},
  year={2025}
}

@inproceedings{trivedi2023interleaving,
  title={Interleaving retrieval with chain-of-thought reasoning for knowledge-intensive multi-step questions},
  author={Trivedi, Harsh and Balasubramanian, Niranjan and Khot, Tushar and Sabharwal, Ashish},
  booktitle={Proceedings of the 61st annual meeting of the Association for Computational Linguistics},
  url={https://aclanthology.org/2023.acl-long.557/},
  year={2023}
}

@inproceedings{yao2022react,
  title={ReAct: Synergizing Reasoning and Acting in Language Models},
  author={Yao, Shunyu and Zhao, Jeffrey and Yu, Dian and Shafran, Izhak and Narasimhan, Karthik R and Cao, Yuan},
  booktitle={NeurIPS 2022 Foundation Models for Decision Making Workshop},
  url={https://openreview.net/forum?id=tvI4u1ylcqs},
  year={2022}
}

@article{deshpande2025trail,
  title={Trail: Trace reasoning and agentic issue localization},
  author={Deshpande, Darshan and Gangal, Varun and Mehta, Hersh and Krishnan, Jitin and Kannappan, Anand and Qian, Rebecca},
  journal={arXiv preprint arXiv:2505.08638},
  url={https://arxiv.org/abs/2505.08638},
  year={2025}
}

@article{pramanick2024spiqa,
  title={Spiqa: A dataset for multimodal question answering on scientific papers},
  author={Pramanick, Shraman and Chellappa, Rama and Venugopalan, Subhashini},
  journal={Advances in Neural Information Processing Systems},
  volume={37},
  year={2024},
  url={https://openreview.net/forum?id=h3lddsY5nf}
}

@inproceedings{huang-etal-2025-towards-multi,
    title = "Towards Multi-Document Question Answering in Scientific Literature: Pipeline, Dataset, and Evaluation",
    author = "Huang, Hui  and
      Velcin, Julien  and
      Kessaci, Yacine",
    editor = "Christodoulopoulos, Christos  and
      Chakraborty, Tanmoy  and
      Rose, Carolyn  and
      Peng, Violet",
    booktitle = "Findings of the Association for Computational Linguistics: EMNLP 2025",
    month = nov,
    year = "2025",
    address = "Suzhou, China",
    publisher = "Association for Computational Linguistics",
    url = "https://aclanthology.org/2025.findings-emnlp.576/",
    doi = "10.18653/v1/2025.findings-emnlp.576",
    ISBN = "979-8-89176-335-7"
}

@inproceedings{gao2026unlocking,
  title={Unlocking Long-Horizon Agentic Search with Large-Scale End-to-End RL},
  author={Gao, Jiaxuan and Fu, Wei and Xie, Minyang and Xu, Shusheng and He, Chuyi and Mei, Zhiyu and Zhu, Banghua and Wu, Yi},
  booktitle={The Fourteenth International Conference on Learning Representations},
  year={2026},
  url={https://openreview.net/forum?id=MfPDdPUGKi}
}

@inproceedings{
shi2026search,
title={Search and Refine During Think: Facilitating Knowledge Refinement for Improved Retrieval-Augmented Reasoning},
author={Yaorui Shi and Sihang Li and Chang Wu and Zhiyuan Liu and Junfeng Fang and Hengxing Cai and An Zhang and Xiang Wang},
booktitle={The Thirty-ninth Annual Conference on Neural Information Processing Systems},
year={2026},
url={https://openreview.net/forum?id=rBlWKIUQey}
}

@misc{
kim2026beyond,
title={Beyond the Final Answer: Evaluating the Reasoning Trajectories of Tool-Augmented Agents},
author={Wonjoong Kim and Sangwu Park and Yeonjun In and Sein Kim and Dongha Lee and Chanyoung Park},
year={2026},
url={https://openreview.net/forum?id=chLlLbI7de}
}

@misc{
dong2026safesearch,
title={SafeSearch: Automated Red-Teaming for the Safety of {LLM}-Based Search Agents},
author={Jianshuo Dong and Sheng Guo and Hao Wang and Xun Chen and Zhuotao Liu and Tianwei Zhang and Ke Xu and Minlie Huang and Han Qiu},
year={2026},
url={https://openreview.net/forum?id=f2ZiFnZEQA}
}

@inproceedings{
wang2026agentnoisebench,
title={AgentNoiseBench: Benchmarking Robustness of Tool-Using {LLM} Agents Under Noisy Condition},
author={Ruipeng Wang and Yuxin Chen and Yukai Wang and Chang Wu and Junfeng Fang and Xiaodong Cai and Qi GU and Hui Su and An Zhang and Xiang Wang and Xunliang Cai and Tat-Seng Chua},
booktitle={Forty-third International Conference on Machine Learning},
year={2026},
url={https://openreview.net/forum?id=ORVJCJczoG}
}

@article{chen2026much,
  title={How Much Can We Trust LLM Search Agents? Measuring Endorsement Vulnerability to Web Content Manipulation},
  author={Chen, Yimeng and Ren, Zhe and Laakom, Firas and Li, Yu and Guo, Dandan and Schmidhuber, J{\"u}rgen},
  journal={arXiv preprint arXiv:2606.16821},
  year={2026}
}

@misc{zhang2026deepresearchagentspoisonedusergenerated,
      title={Deep-Research Agents Can Be Poisoned via User-Generated Content}, 
      author={Tingwei Zhang and Harold Triedman and Vitaly Shmatikov},
      year={2026},
      eprint={2605.24245},
      archivePrefix={arXiv},
      primaryClass={cs.CR},
      url={https://arxiv.org/abs/2605.24245}, 
}
\newtcolorbox{promptbox}[1][]{%
  colback      = green!5!white,
  colframe     = green!75!black,
  fonttitle    = \bfseries,
  colbacktitle = green!85!black,
  fontupper    = \ttfamily\fontsize{9pt}{9pt}\selectfont,
  arc=2mm,
  boxrule=0.4mm,
  left=2mm, right=2mm, top=1.5mm, bottom=1.5mm,
  breakable,
  #1
}

\newtcolorbox{systempromptbox}[1][]{%
  colback=white,
  colframe=black!65,
  fonttitle=\bfseries,
  colbacktitle = black!65,
  fontupper    = \ttfamily\fontsize{9pt}{9pt}\selectfont,
  arc=2mm,
  boxrule=0.4mm,
  left=2mm, right=2mm, top=1.5mm, bottom=1.5mm,
  breakable,
  #1
}

\fvset{breaklines=true,
       breakanywhere=true,
       breaksymbolleft={},
       breaksymbolright={}}

\appendix

\section{Details about synthetically generated documents}
Synthetic document generation is central to our stress-testing setup. This section provides additional information about the quality of the generated documents, including human evaluation results, statistics, and examples.

\subsection{Human evaluation}
\label{sec:human_eval}
To validate that our generation pipeline follows the intended document attributes, we conduct a human evaluation with three annotators. The sample focuses on relevant documents: in our generation pipeline, relevance is controlled systematically by swapping the agent query, whereas the LLM is mainly responsible for following the trustworthiness and factuality constraints. Annotators are shown an agent query, the corresponding synthetic document generated by our pipeline, and the supporting context associated with the original question from \HotpotQA{} or \BrowseCompPlus{}. They are then asked to independently judge whether the document appears trustworthy, whether its claims are factually correct (in this setting, relevance is always annotated as True).

We run this annotation process in \texttt{LabelStudio}\footnote{\url{https://labelstud.io/}} on 50 randomly selected tool calls. Three annotators are involved, and each document is annotated by at least 2 annotators, resulting in 100 annotations. In Table \ref{tab:human_eval}, we report raw accuracy and the binary correlation coefficient $\phi$, and Cohen's $\kappa$ for interannotator agreement. 

Overall, the annotations show strong accuracy with respect to the intended synthetic labels and high inter-annotator agreement. Agreement is highest for trustworthiness, while factuality is more challenging, reflecting the greater ambiguity of judging whether all claims in a generated document are correct. Further analysis in table~\ref{tab:factuality_confusion} revealed that annotators tended to judge supposedly factual documents as non-factual, while supposedly non-factual documents were nearly always assessed as so. Evaluating factuality might require finer-grained assessment as some documents may be only partially correct, but overall, this evaluation confirms that our synthetic generation pipeline is consistent. 

\begin{table}[h]
\centering
\footnotesize
\setlength{\tabcolsep}{3pt}
\begin{tabularx}{\columnwidth}{Xccc}
\toprule
\textbf{Dimension} & \textbf{Labeling} & \multicolumn{2}{c}{\textbf{Inter-ann.}} \\
 & \textbf{Acc. (\%)} & \textbf{Agr. (\%)} & \textbf{$\kappa$} \\
\midrule
Trustworthiness & 96.4  & 93.8 & 0.86 \\
Factuality & 85.5 & 81.2 & 0.64 \\
\midrule
\textbf{Overall} & \textbf{92.8} & \textbf{90.6} & \textbf{0.77} \\
\bottomrule
\end{tabularx}
\caption{Human agreement with synthetic labels and between annotators.}
\label{tab:human_eval}
\end{table}

\begin{table}[h]
\centering
\footnotesize
\setlength{\tabcolsep}{4pt}
\begin{tabular}{lcc}
\toprule
\textbf{Label} & \multicolumn{2}{c}{\textbf{Human}} \\
\cmidrule(lr){2-3}
 & \textbf{Not Factual} & \textbf{Factual} \\
\midrule
\textbf{Not Factual} & 30 & 1 \\
\textbf{Factual} & 13 & 56 \\
\bottomrule
\end{tabular}
\caption{Factuality confusion matrix}
\label{tab:factuality_confusion}
\end{table}

\begin{figure*}[t]
    \centering
    \includegraphics[width=\linewidth]{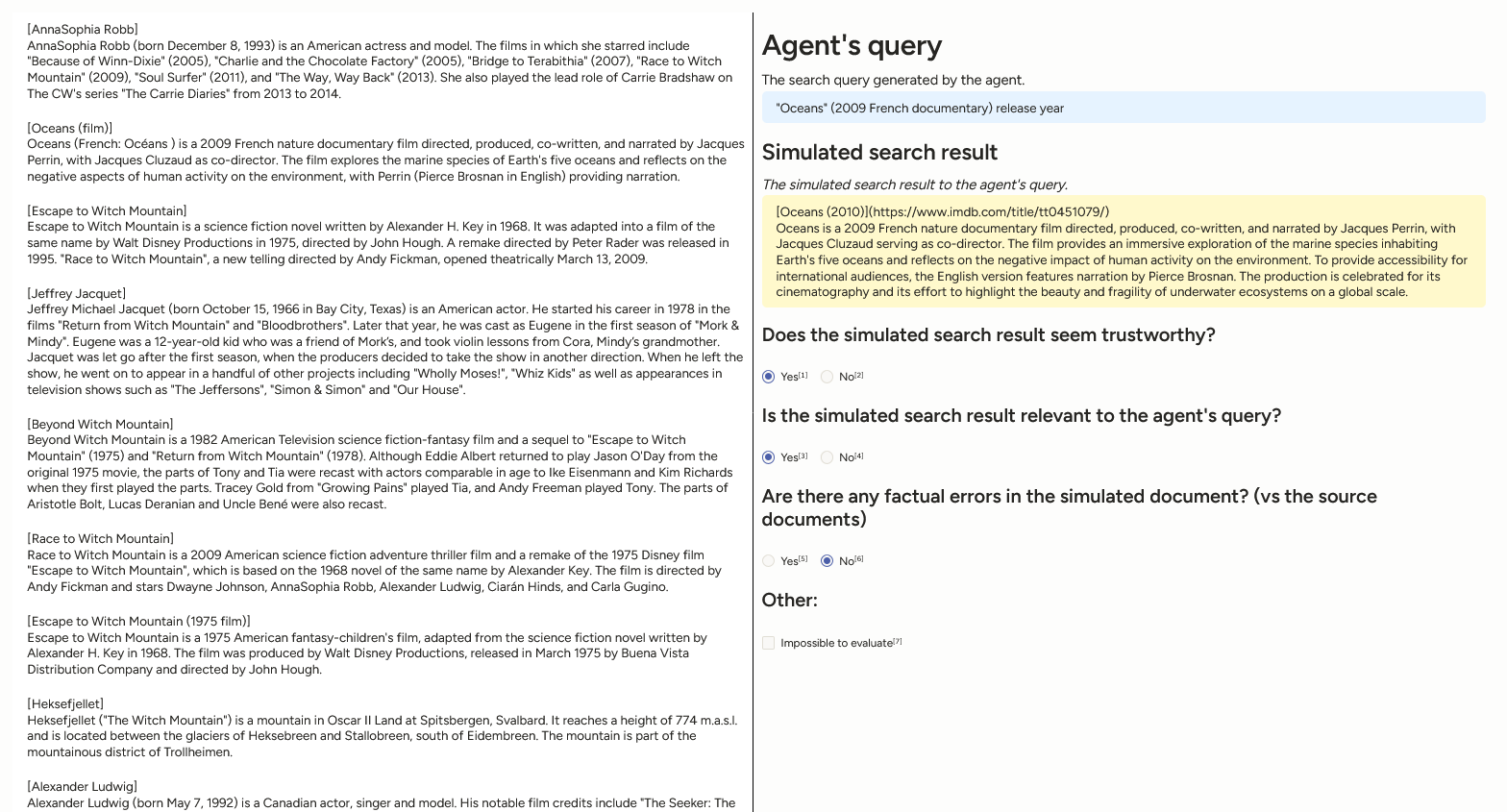}
    \caption{Human evaluation in our LabelStudio interface}
    \label{fig:human_evaluation_labelstudio}
\end{figure*}

\subsection{Documents statistics}
In Table \ref{tab:document_word_stats}, we present statistics regarding the length of the documents generated. 

\begin{table}[!h]
\centering
{
\begin{tabular}{lrrr}
\toprule
\textbf{Dataset} & \textbf{Mean} & \textbf{Median} & \textbf{Std} \\
\midrule
HotpotQA & 102.7 & 104.0 & 13.4 \\
BrowseComp-Plus & 106.4 & 107.0 & 13.9 \\
\bottomrule
\end{tabular}
}
\caption{Length of generated documents in words for \HotpotQA{} and \BrowseCompPlus{}.}
\label{tab:document_word_stats}
\end{table}

\subsection{Examples}
\label{sec:data_generation_examples_debussy}
Table~\ref{tab:debussy_generated_documents} shows examples of synthetic documents produced by our generation pipeline under different reliability conditions. All examples use the same query, ``\textit{Debussy innovations orchestral color impressionism}'', and are generated following the procedure described in Appendix~\ref{sec:synthetic_document_generation}.

\begin{table*}[h]
\centering
\scriptsize
\setlength{\tabcolsep}{4pt}
\renewcommand{\arraystretch}{1.08}
\begin{tabularx}{\textwidth}{p{0.23\textwidth}X}
\toprule
\textbf{Condition} & \textbf{Generated document} \\
\midrule
Trustworthy, relevant, factual &
Claude Debussy was a French composer whose innovations in orchestral color and harmony were central to the development of musical impressionism. He moved away from the traditional structural constraints of the Romantic era, instead focusing on the atmospheric quality of sound and the evocative use of instrumental timbres. By utilizing unconventional scales and fluid rhythms, Debussy created a sense of luminosity and spatial depth in his orchestrations. His approach emphasized the immediate sensory impact of a chord or a melodic fragment, effectively treating the orchestra as a palette of colors to depict nature and mood, which profoundly influenced the trajectory of twentieth-century classical music. \\
\midrule
Trustworthy, relevant, non-factual &
Claude Debussy was a Baroque composer from the 17th century who is credited with inventing the strict rules of the fugue. He is best known for his rigid adherence to traditional Germanic counterpoint and his refusal to use orchestral color, preferring instead to write exclusively for the harpsichord. By eliminating harmony and rhythm from his works, Debussy established a style of mathematical precision that avoided all forms of impressionism. His innovations focused primarily on reducing the size of the orchestra to a single percussion instrument, ensuring that his music remained devoid of atmosphere or sensory imagery, which served as a direct precursor to the minimalist movement of the Renaissance. \\
\midrule
Trustworthy, irrelevant, factual &
The Siege of Sparta was a significant military engagement that occurred in 272 BC, pitting the forces of Epirus against a strategic alliance. The Epirotes were led by King Pyrrhus, who sought to expand his influence, while the defense of the city was managed by an alliance consisting of Macedon and Sparta. The Spartan forces were under the direct command of King Areus I and his heir, Acrotatus. The battle took place at Sparta and ultimately concluded with a decisive victory for the Spartan-Macedonian alliance, effectively ending the campaign of King Pyrrhus in the region and securing the city from the Epirote invasion. \\
\midrule
Untrustworthy, relevant, factual &
OMG you won't believe how Debussy totally changed the game with his mind-blowing approach to orchestral color! This French rebel basically invented impressionism in music, throwing out all the boring old rules to create these shimmering, dreamy soundscapes that feel like a literal trip. While other composers were stuck in the past, Debussy was using innovative textures and wild harmonies to paint pictures with sound. It is absolutely scandalous how he manipulated the orchestra to get those hazy, ethereal vibes that still make modern listeners gasp today. If you think you know music, you are wrong until you dive into the sheer audacity of his orchestral innovations! \\
\bottomrule
\end{tabularx}
\caption{Examples of generated search-result documents for the query ``debussy innovations orchestral color impressionism'' using the synthetic generation pipeline. Disclaimer: the second line contains non-factual information that may not be trusted.}
\label{tab:debussy_generated_documents}
\end{table*}

\section{Aggregate metrics discussion}
\label{sec:aggregate_metrics_discussion}
In Section~\ref{sec:metrics}, we describe the \textit{Tokens Per Correctness} (TPC) metric from \citet{xie-etal-2026-searching} and introduce our reliability-aware metrics: \textit{Reliability-Aware Score} (RAS) and \textit{Cost per RAS} (CoPRAS). These aggregate metrics depend on several design choices, including how correctness and cost are defined, and how the maximum tool-call budget is set. In this section, we discuss the motivation for these choices and illustrate their impact empirically.

\subsection{Design choices}
\paragraph{Cost.}
To make our results comparable with prior work, we reuse the cost definition from \citet{xie-etal-2026-searching}, which combines generated tokens, input tokens, and the number of search calls. Following their setup, generated tokens have weight \(1\), input tokens have weight \(0.25\), and each search call has weight \(500\). These values are motivated by the relative cost of input and output tokens, and by the estimated cost of a web-search API call expressed in equivalent output tokens.
This choice is reasonable for web-search agents, but it is not universal. In particular, systems that search over a local document collection, or use a cheaper internal retriever rather than a public search API, would require different weights. We therefore treat this cost function as a useful reference point rather than an absolute measure of computational cost.

\paragraph{Scoring.}
Aggregate scores are useful because they allow us to rank systems directly. However, they also compress several different outcomes into a single number. This choice is not neutral: it depends on what we want to evaluate. From a user perspective, it may be enough that the system avoids giving a wrong answer. From a system-design perspective, we may want to distinguish more carefully between a model that says ``I don't know'', a model that keeps searching until the tool budget is exhausted, a model that fails to produce a valid answer tag, and a model that gives an incorrect answer.
RAS defines one possible scoring choice. It gives full credit to correct answers, no credit to incorrect or malformed answers, and partial credit to abstention or tool-budget exhaustion when reliable evidence is unlikely. This reflects the idea that, under degraded document conditions, not answering can be better than giving an unsupported answer. However, this is not the only possible design. For example, one could give less credit to tool-budget exhaustion than to explicit abstention, such as \(0.5-\rho_{\gamma_i}\) instead of \(1-\rho_{\gamma_i}\), because clearly stating uncertainty may be preferable to searching until the budget is reached. This weighting is necessarily a design choice. CoPRAS partly captures this difference through cost, since long searches are more expensive than early abstention.
We therefore treat RAS and CoPRAS as summary metrics, not as replacements for the detailed outcome distributions. They are useful for comparing systems from a reliability-oriented perspective, but other scoring choices could emphasize different behaviors.

\subsection{Impact of tool-call budget}

For the main experiments, we set the tool-call budget to 100. This choice directly affects our aggregate metrics. A lower budget penalizes systems that usually need many searches before answering, since they are more likely to stop before producing a final answer. It also affects cost-based metrics, because the number of tool calls is part of the cost definition. Changing the budget can therefore change both metric values and system rankings. We analyze this effect by recomputing the metrics under different tool-call budgets.

\begin{figure*}[t]
    \centering
    \includegraphics[width=\linewidth]{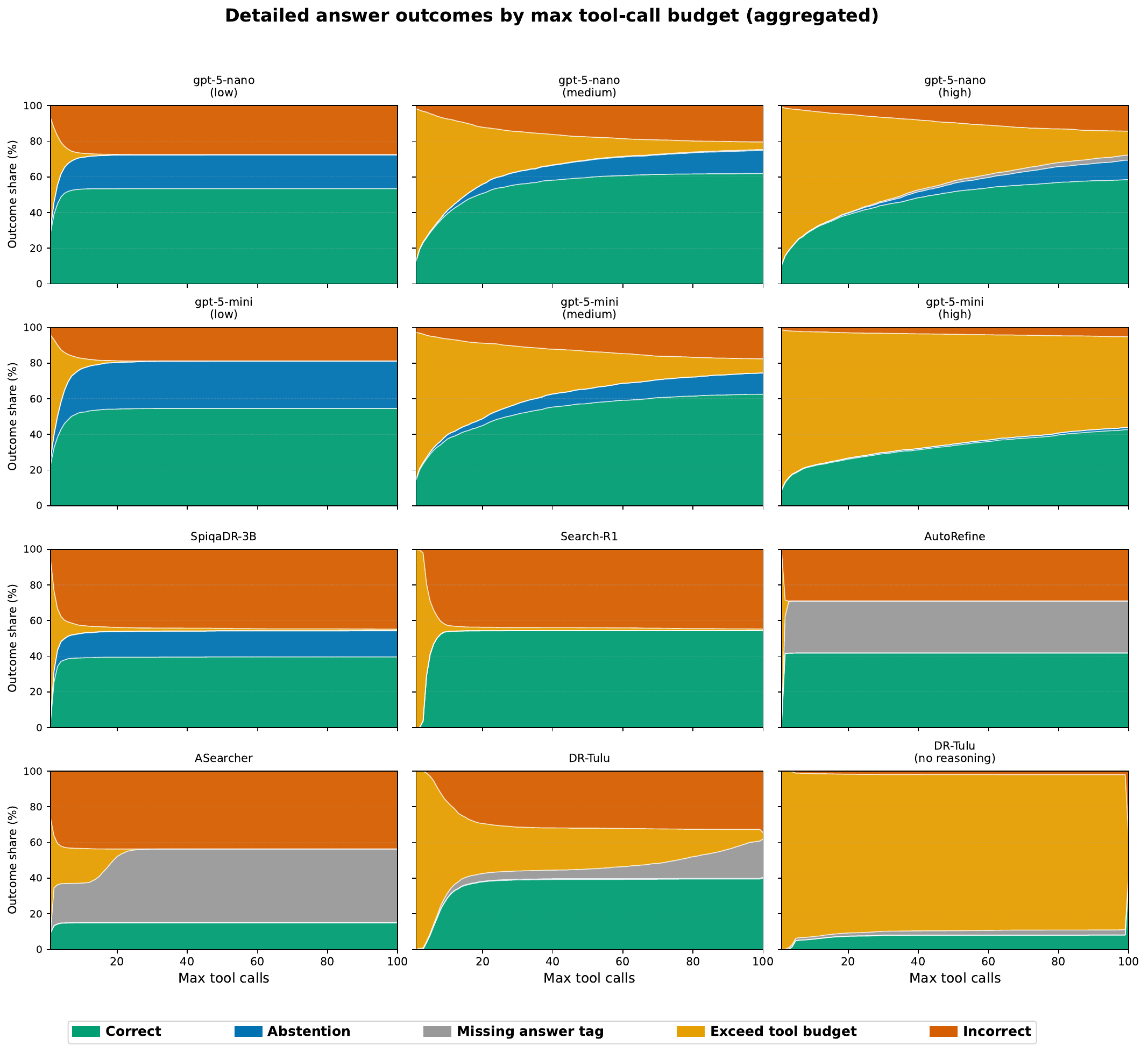}
    \caption{Detailed outcomes vs tool-calls budget}
    \label{fig:detailed_outcomes_vs_budget}
\end{figure*}

\begin{figure*}[p]
    \centering

    \begin{subfigure}[t]{0.48\textwidth}
        \centering
        \includegraphics[width=\linewidth]{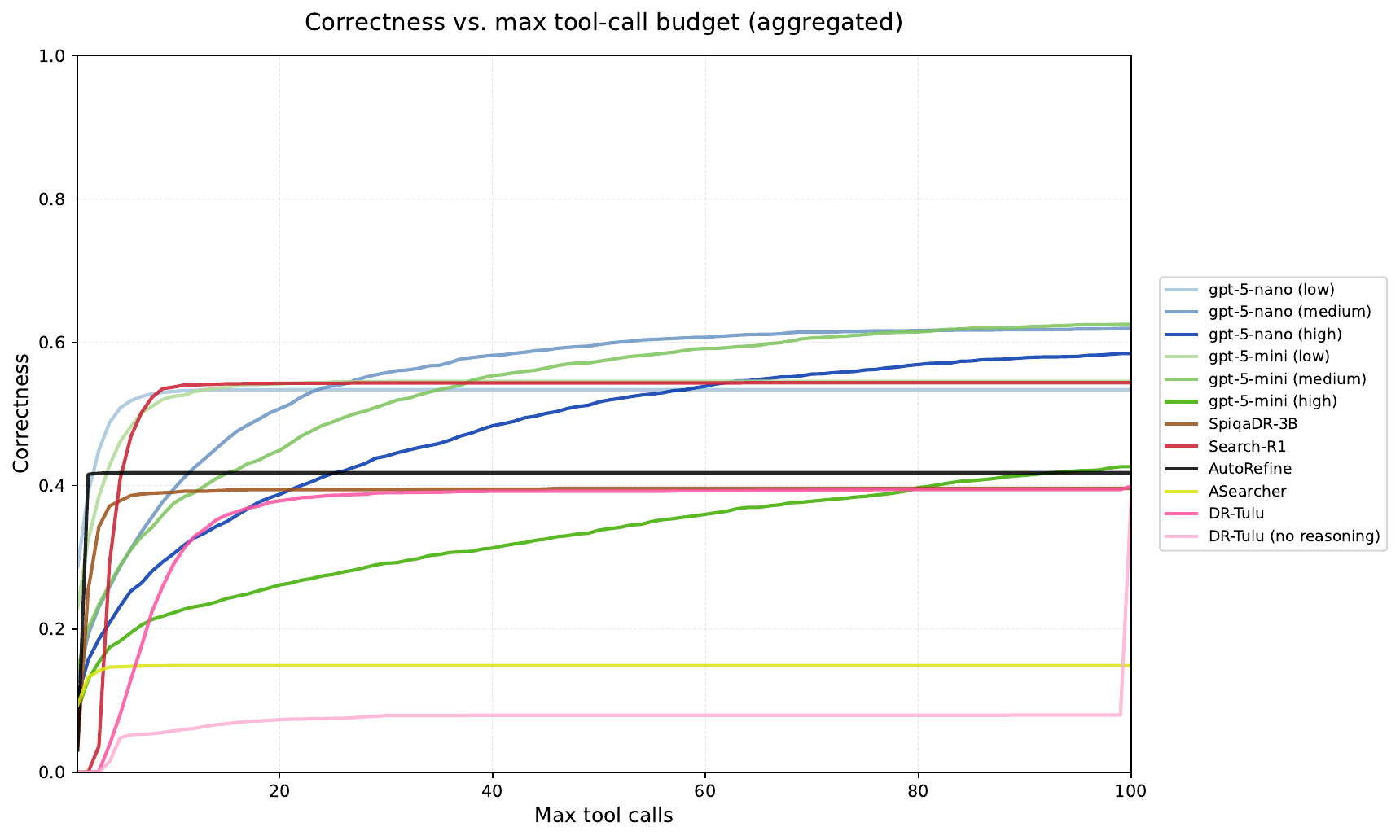}
        \caption{Correctness score evolution}
        \label{fig:correctness_score_evol}
    \end{subfigure}
    \hfill
    \begin{subfigure}[t]{0.48\textwidth}
        \centering
        \includegraphics[width=\linewidth]{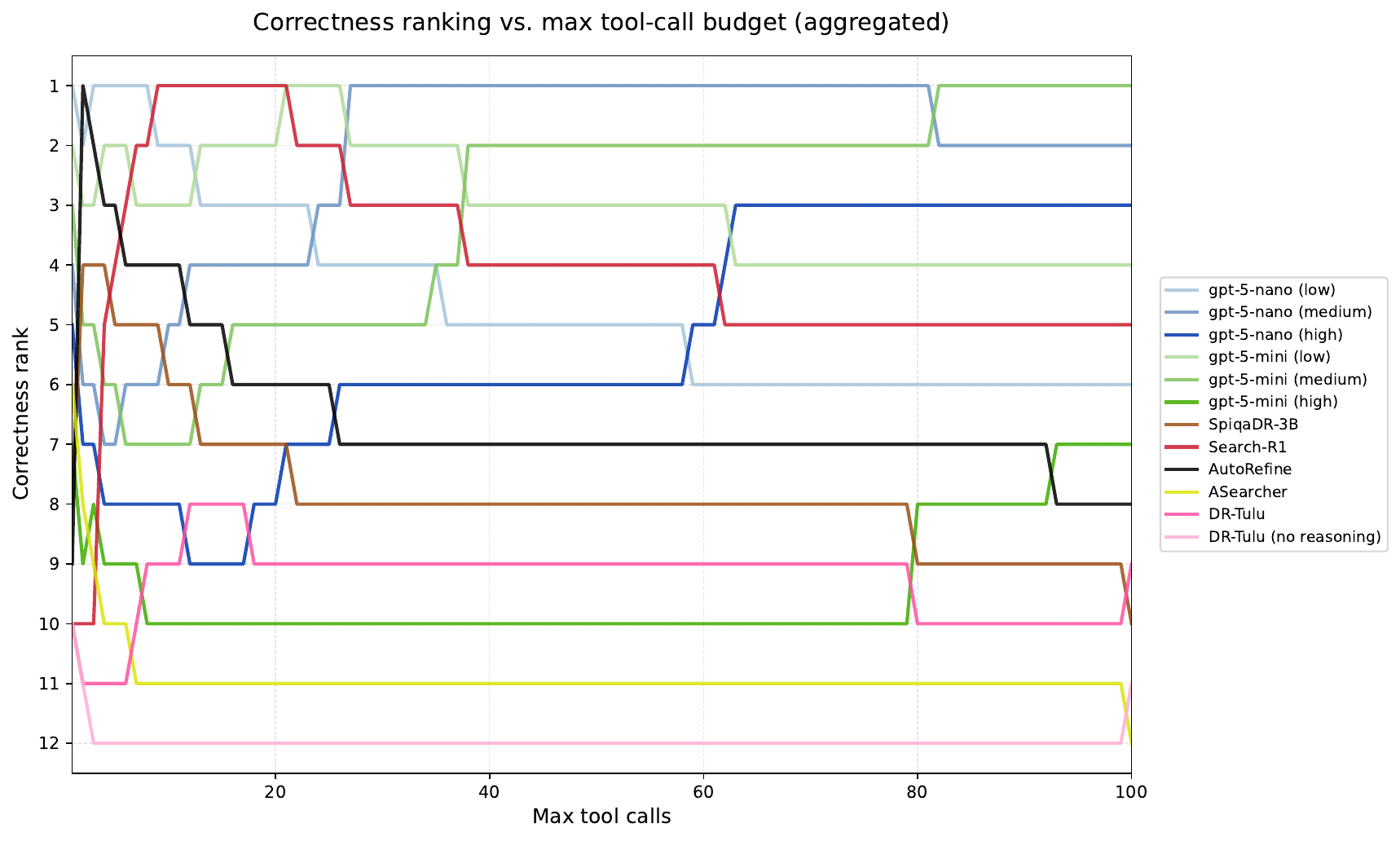}
        \caption{Correctness rankings evolution}
        \label{fig:correctness_ranks_evol}
    \end{subfigure}

    \vspace{0.6em}

    \begin{subfigure}[t]{0.48\textwidth}
        \centering
        \includegraphics[width=\linewidth]{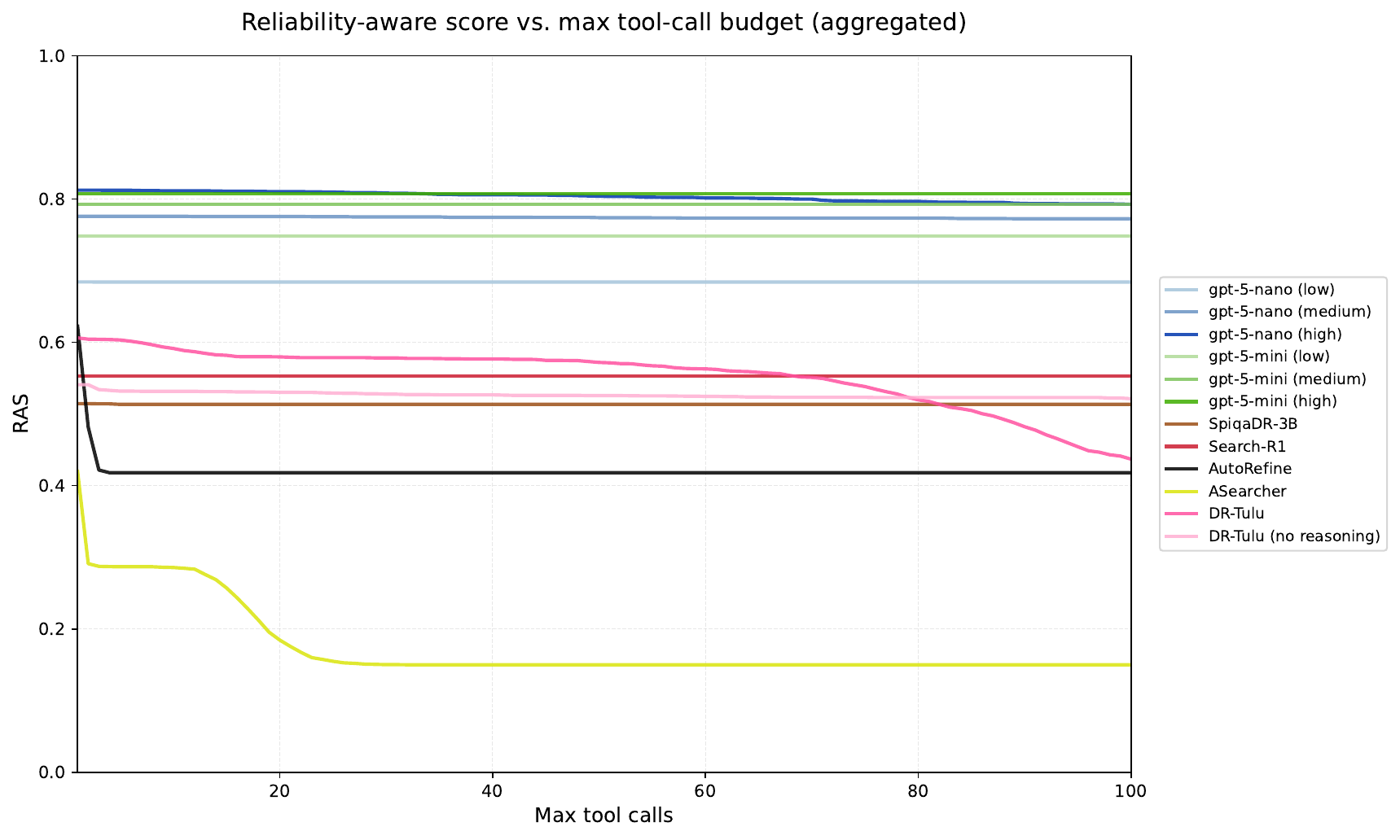}
        \caption{RAS score evolution}
        \label{fig:ras_score_evol}
    \end{subfigure}
    \hfill
    \begin{subfigure}[t]{0.48\textwidth}
        \centering
        \includegraphics[width=\linewidth]{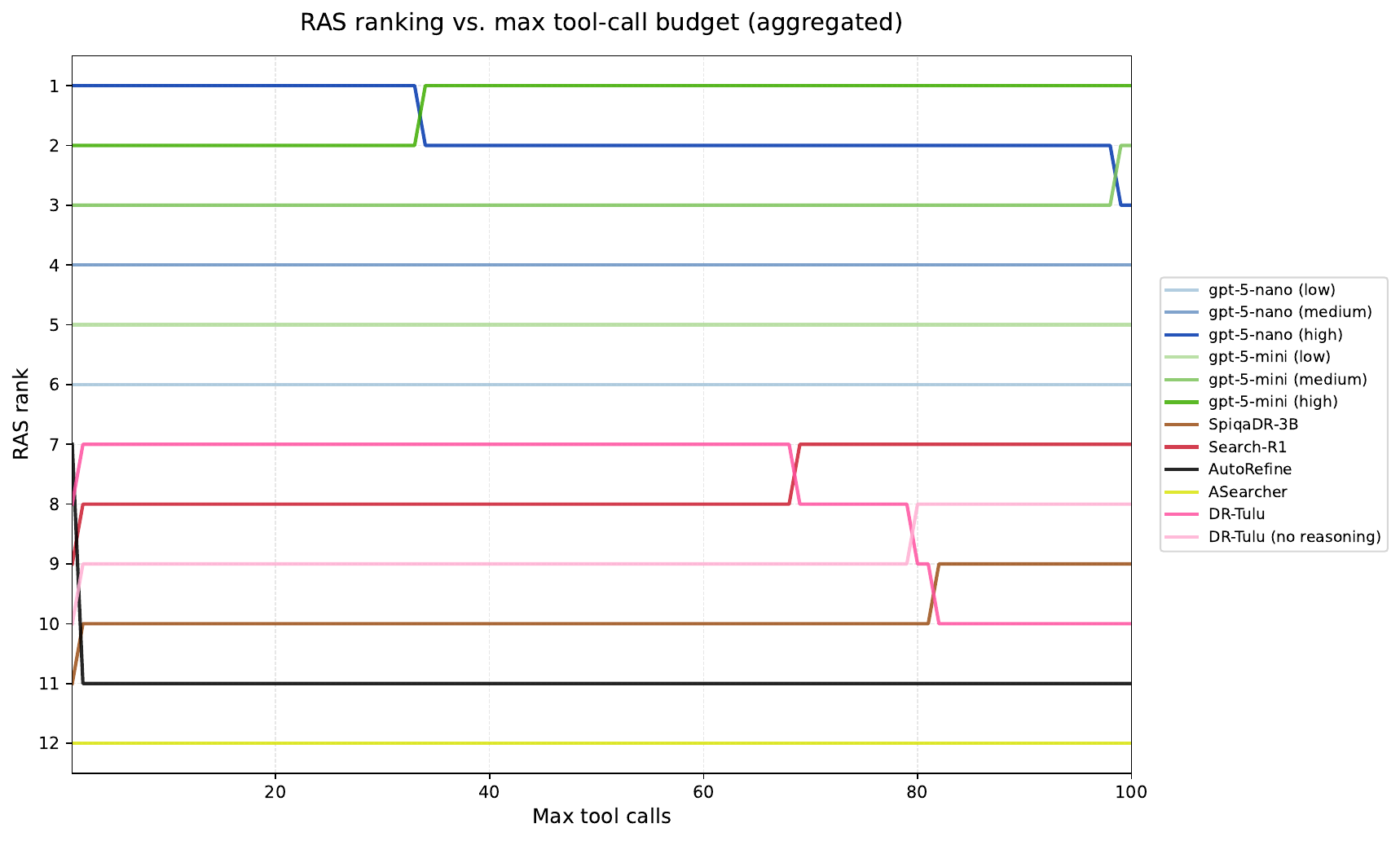}
        \caption{RAS rankings evolution}
        \label{fig:ras_ranks_evol}
    \end{subfigure}

    \vspace{0.6em}

    \begin{subfigure}[t]{0.48\textwidth}
        \centering
        \includegraphics[width=\linewidth]{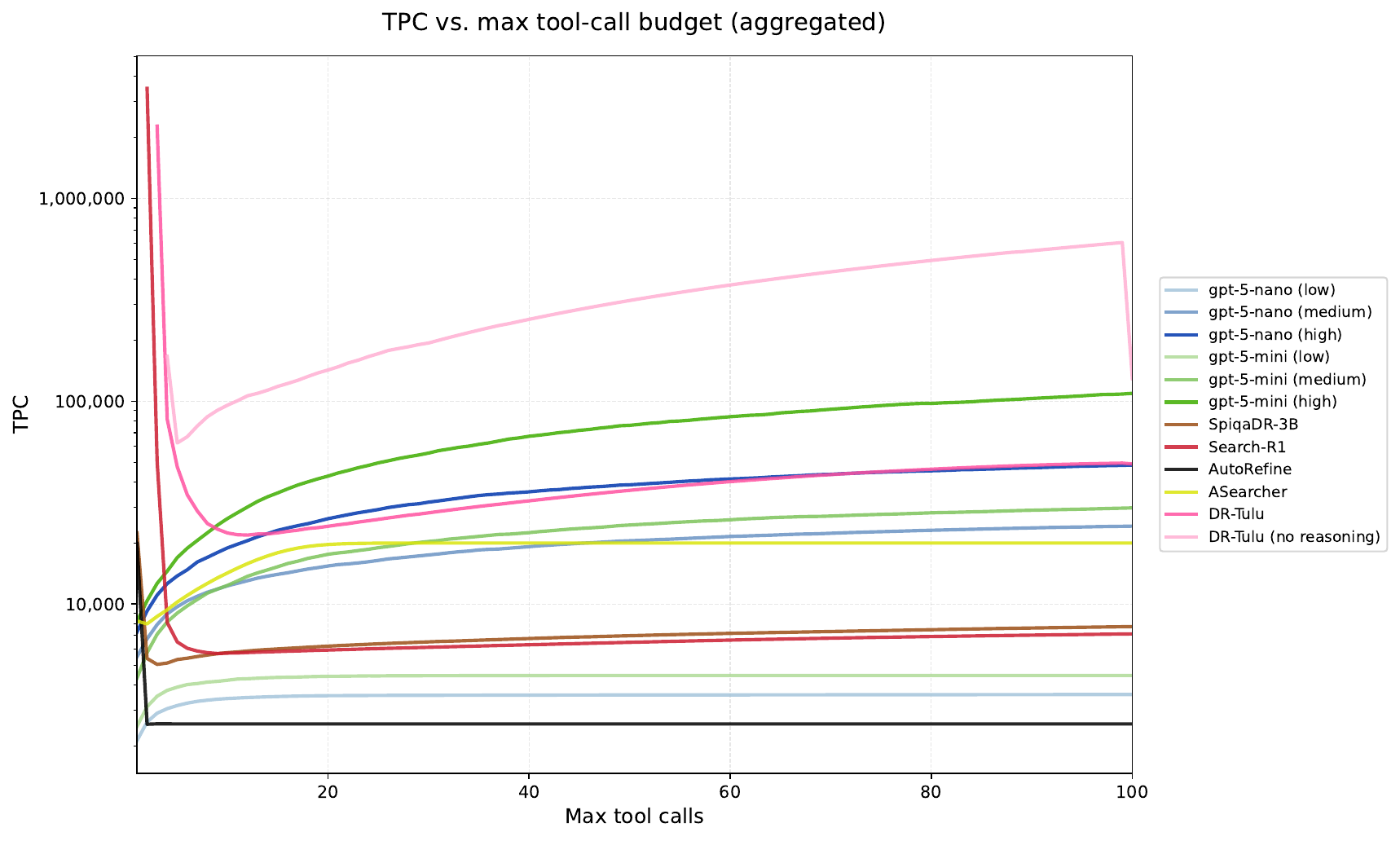}
        \caption{TPC score evolution}
        \label{fig:tpc_score_evol}
    \end{subfigure}
    \hfill
    \begin{subfigure}[t]{0.48\textwidth}
        \centering
        \includegraphics[width=\linewidth]{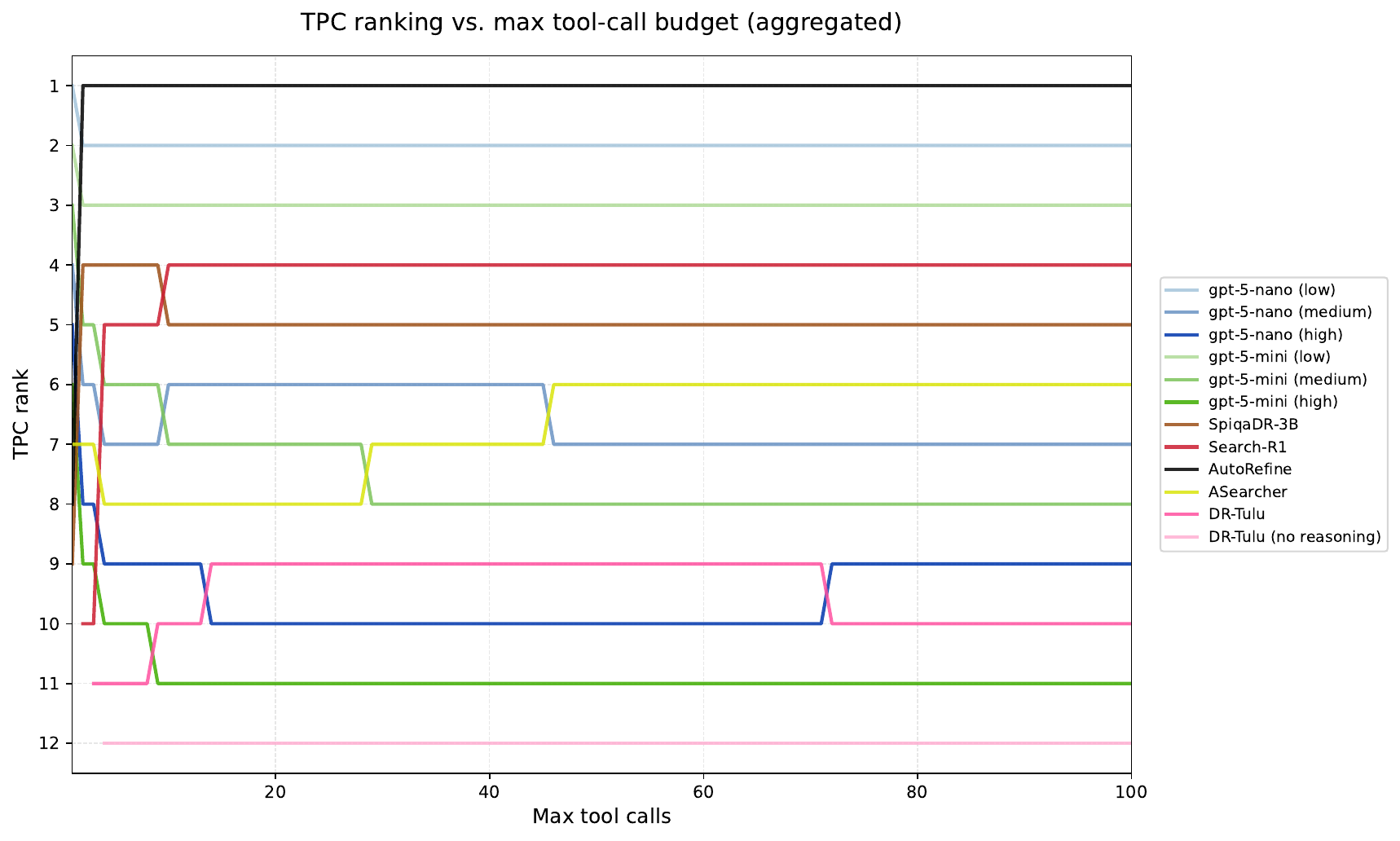}
        \caption{TPC rankings evolution}
        \label{fig:tpc_ranks_evol}
    \end{subfigure}

    \vspace{0.6em}

    \begin{subfigure}[t]{0.48\textwidth}
        \centering
        \includegraphics[width=\linewidth]{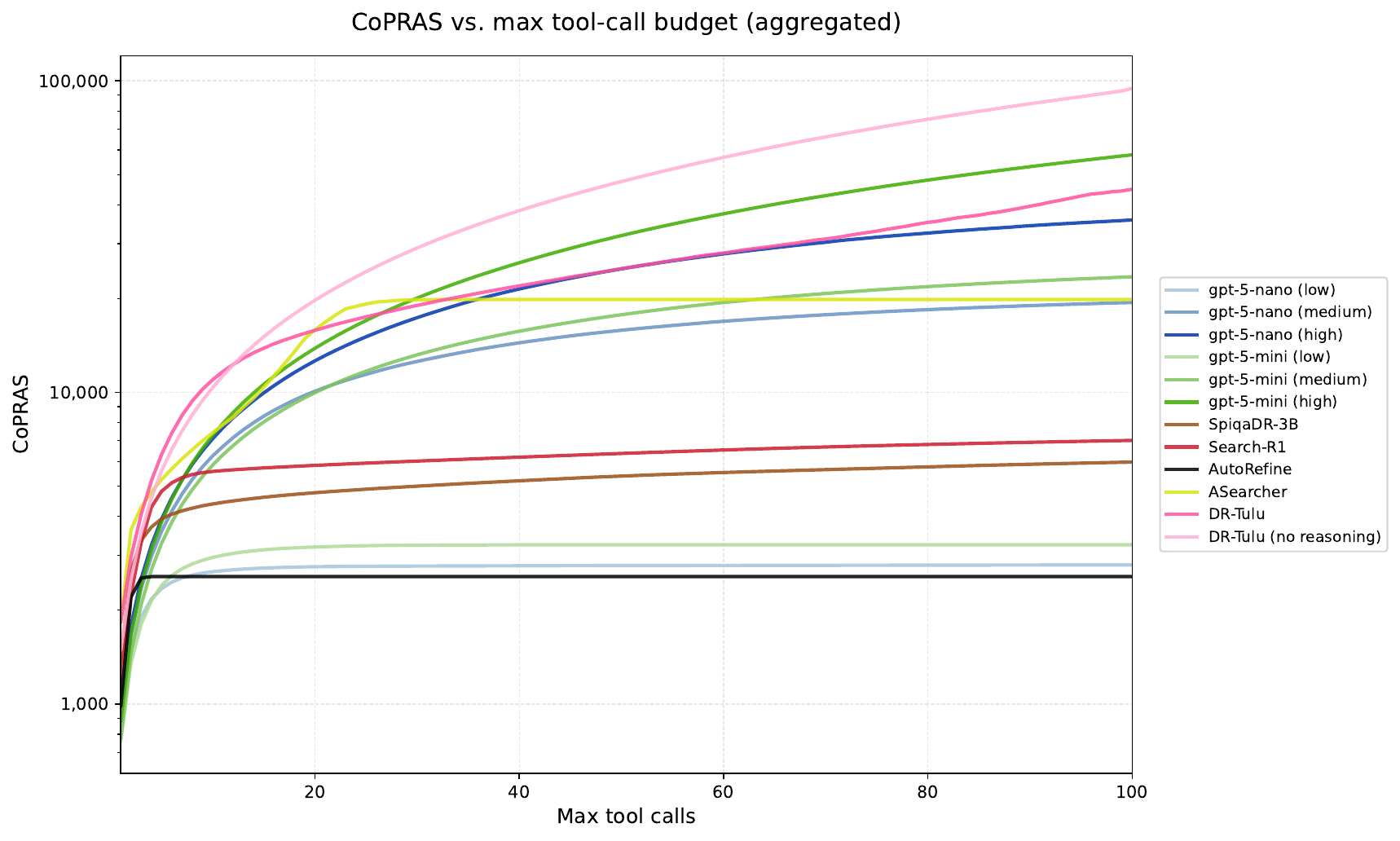}
        \caption{CoPRAS score evolution}
        \label{fig:copras_score_evol}
    \end{subfigure}
    \hfill
    \begin{subfigure}[t]{0.48\textwidth}
        \centering
        \includegraphics[width=\linewidth]{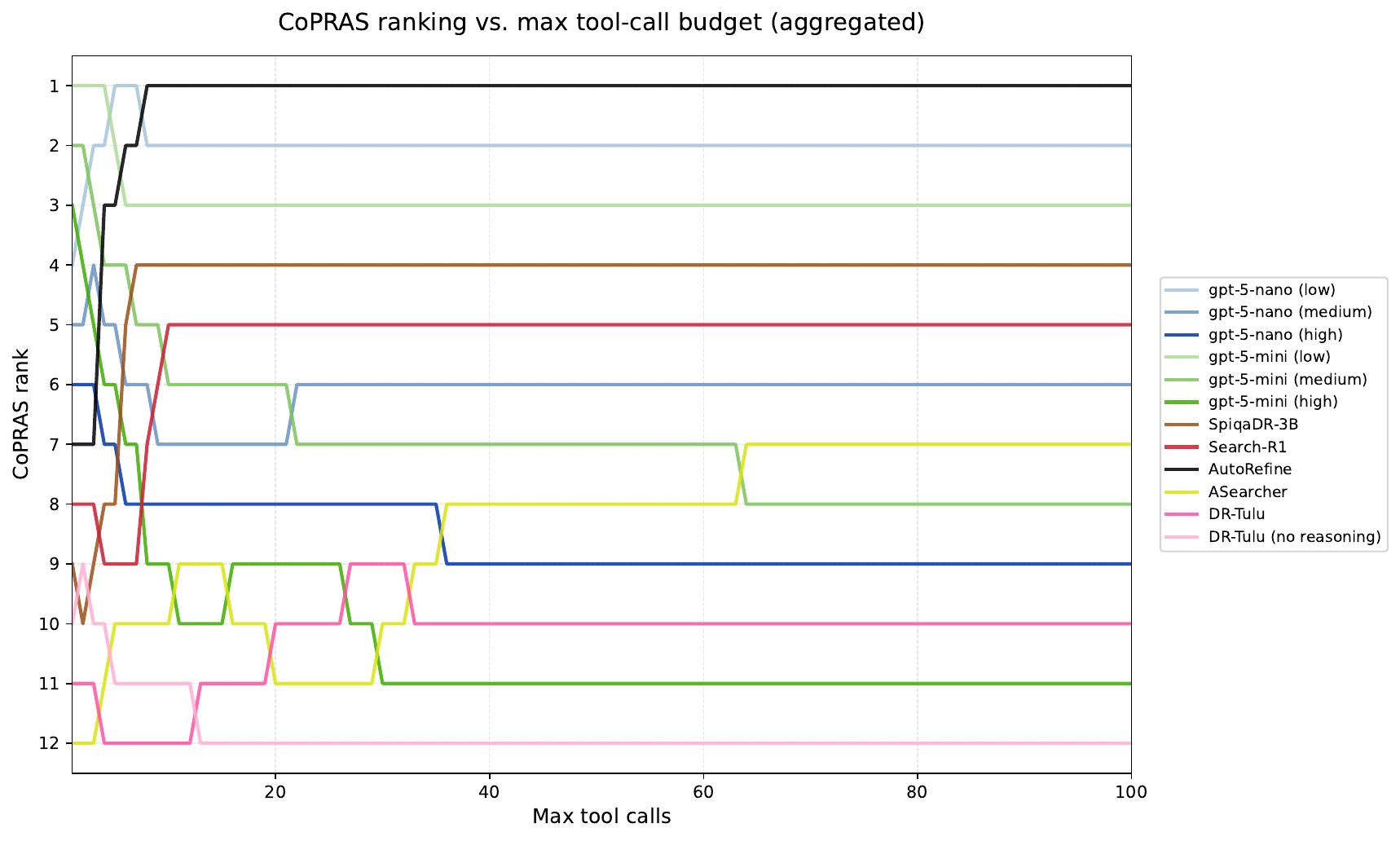}
        \caption{CoPRAS rankings evolution}
        \label{fig:copras_ranks_evol}
    \end{subfigure}

    \caption{Effect of tool budget on aggregate metrics: correctness, RAS, TPC, and CoPRAS.}
    \label{fig:all_metrics_vs_budget}
\end{figure*}

Figure~\ref{fig:detailed_outcomes_vs_budget} shows how final-answer outcomes change as the budget increases. Figure~\ref{fig:all_metrics_vs_budget} shows the corresponding evolution of \textbf{Correctness}, \textbf{RAS}, \textbf{TPC}, and \textbf{CoPRAS}. Overall, correctness increases with the budget, but the rate of improvement differs across systems. Models with narrow tool-call distributions reach their final performance quickly, while models with more spread-out distributions, such as \texttt{GPT-5-nano (high)} and \texttt{GPT-5-mini (high)}, improve more gradually. As a result, correctness-based rankings become reasonably stable only after around 50 tool calls, although high-reasoning models continue to benefit from larger budgets.

RAS is more stable with respect to the tool-call budget. Unlike correctness, it does not always increase with additional turns, because abstention and budget exhaustion can receive partial credit under unreliable document conditions. Several systems therefore plateau after only a small number of tool calls. The main exceptions are \texttt{DR-Tulu} and \texttt{ASearcher}, which often terminate without producing a valid final answer tag; increasing the budget can expose more of these incomplete interactions and penalize them in RAS. Overall, the relative stability of RAS suggests that smaller budgets, around 10 tool calls, may already be sufficient to obtain informative RAS-based rankings, while substantially reducing evaluation cost.

TPC and CoPRAS show broadly similar trends as the budget increases, since both normalize cost by a performance score. The main difference appears at very small budgets: some systems have near-zero correctness in this regime, which makes TPC extremely large or infinite. CoPRAS is less sensitive to this issue when systems receive reliability-aware credit through RAS, but it still increases when additional tool calls raise the cost without improving the score.

\section{Additional qualitative experiments}
\label{sec:additional_experiments}
We report additional experiments that complement the main results with a more qualitative view of agent behavior. We first examine the volume and content of search queries produced by different systems, and whether these queries change under degraded document conditions. We then analyze the agents reasoning steps to measure whether they explicitly identify low trustworthiness, low relevance, or low factuality of the returned documents in their chain-of-thoughts. These experiments are intended as exploratory process-level analyses and point to future work on more systematic evaluation of search-agent behavior beyond final-answer accuracy.

\subsection{Search queries}
\label{sec:search_queries}
\begin{figure*}[t]
    \centering
    \begin{subfigure}[t]{0.49\textwidth}
        \centering
        \includegraphics[width=\linewidth]{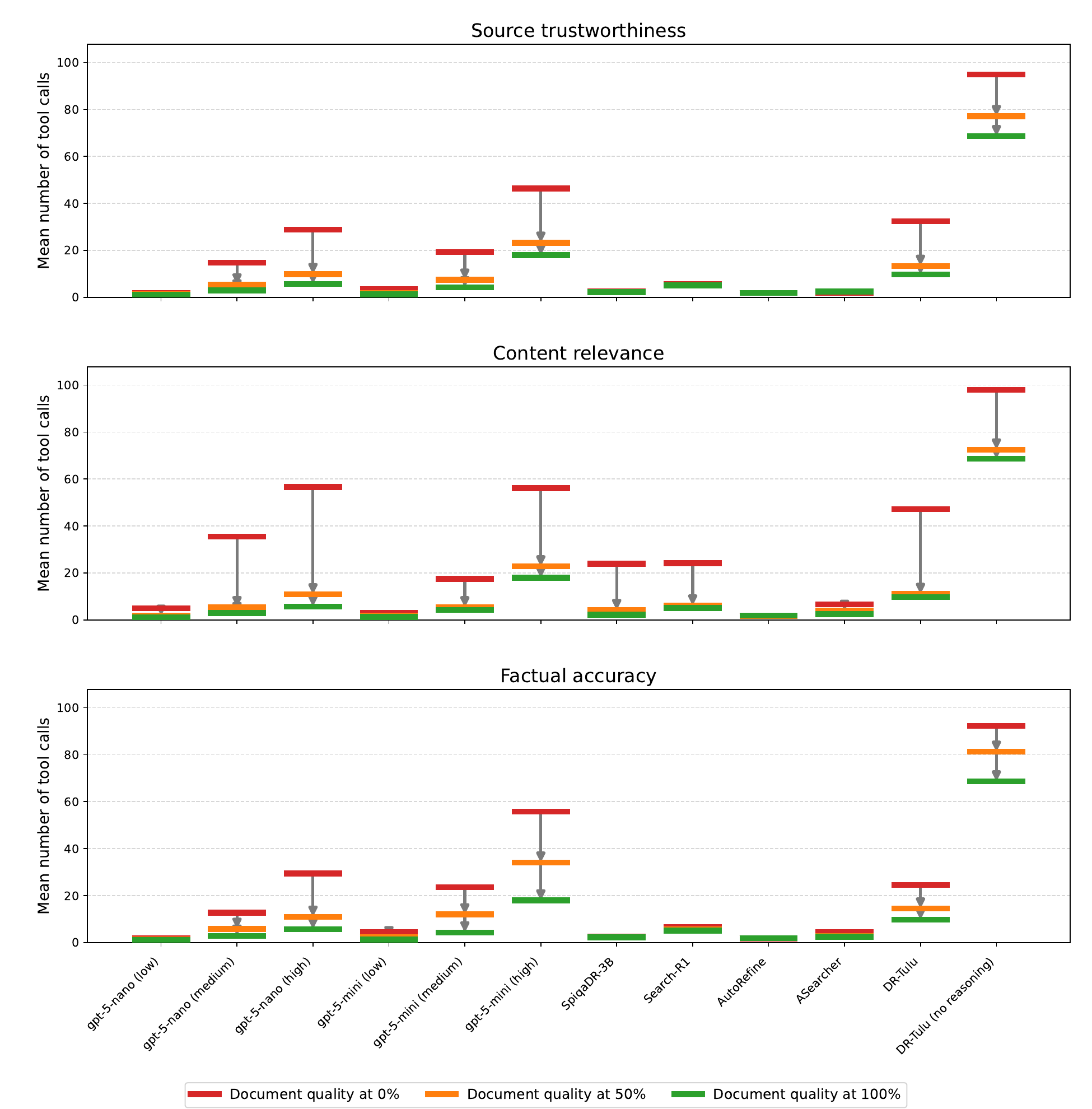}
        \caption{HotpotQA}
        \label{fig:hpqa_n_tool_calls}
    \end{subfigure}
    \hfill
    \begin{subfigure}[t]{0.49\textwidth}
        \centering
        \includegraphics[width=\linewidth]{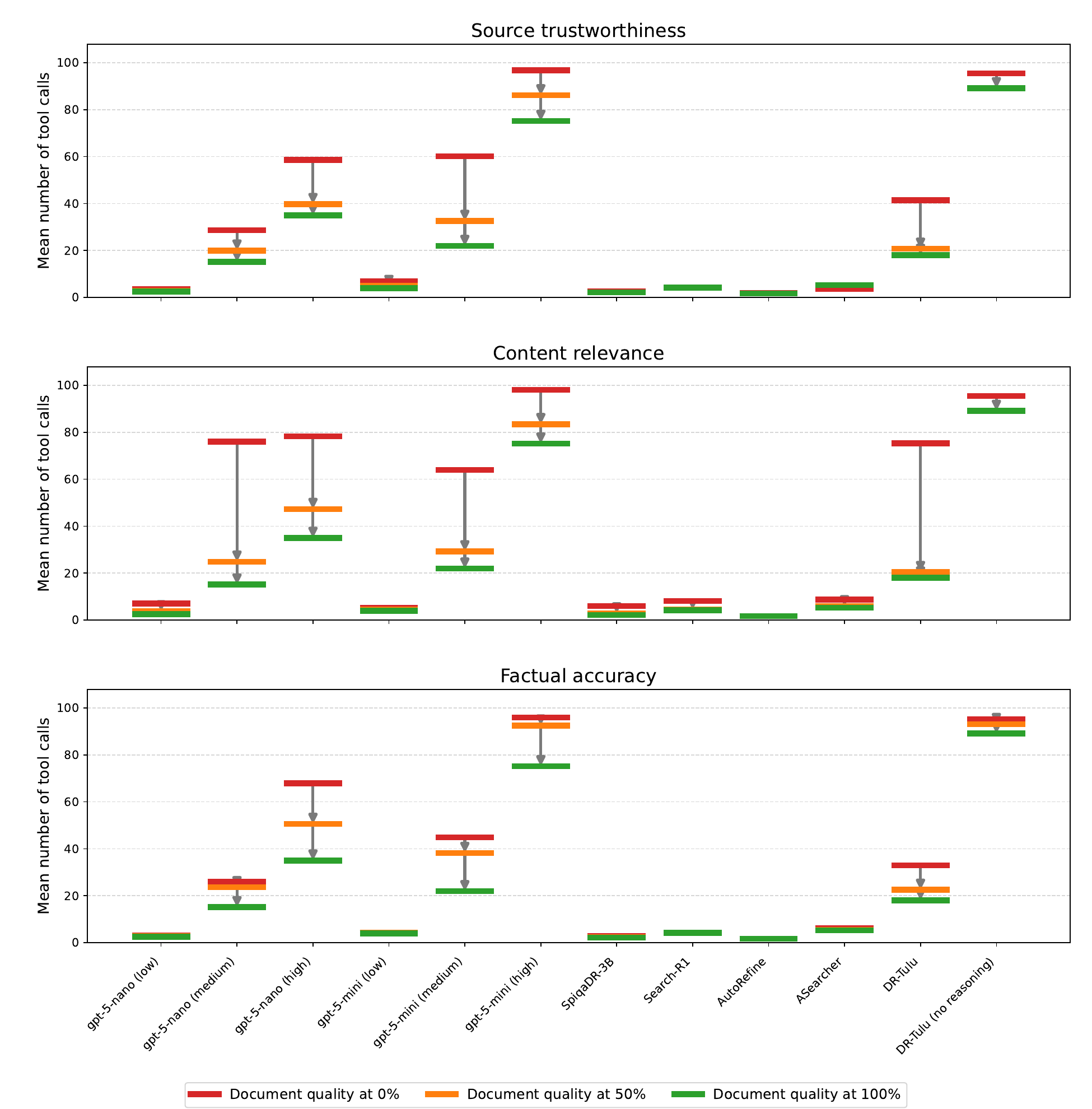}
        \caption{BrowseComp-Plus}
        \label{fig:bc_n_tool_calls}
    \end{subfigure}

    \caption{Mean number of tool calls per dataset, per model, for values at 0\%, 50\% and 100\% on all axes.}
    \label{fig:n_tool_calls}
\end{figure*}

\subsubsection{Number of searches}
\label{sec:search_volume}
As shown in Figure~\ref{fig:n_tool_calls}, we observe large differences in the number of tool calls made by each agent. Since our setup exposes only one tool, \textit{search}, this is equivalent to the number of search queries issued by the agent.

Most specialized open-weight models make few tool calls, with limited variation as document quality decreases. The main exceptions are DR-Tulu, and SpiqaDR and Search-R1 when relevance is set to 0 on \HotpotQA{}. GPT-based agents show a clearer scaling with reasoning effort: low-effort runs are often close to open-weight models, while high-effort runs search much more extensively. For these models, document degradation generally increases the number of searches, especially at higher reasoning effort.

Agents also search more on \BrowseCompPlus{} than on \HotpotQA{} on average. This may reflect the role of parametric knowledge in \HotpotQA{}, which can reduce the need for search. It may also reflect task structure: \BrowseCompPlus{} questions require persistent search over multiple constraints, whereas \HotpotQA{} questions usually involve shorter multi-hop reasoning over Wikipedia-style facts.

\begin{figure*}[t]
    \centering

    \includegraphics[width=\linewidth]{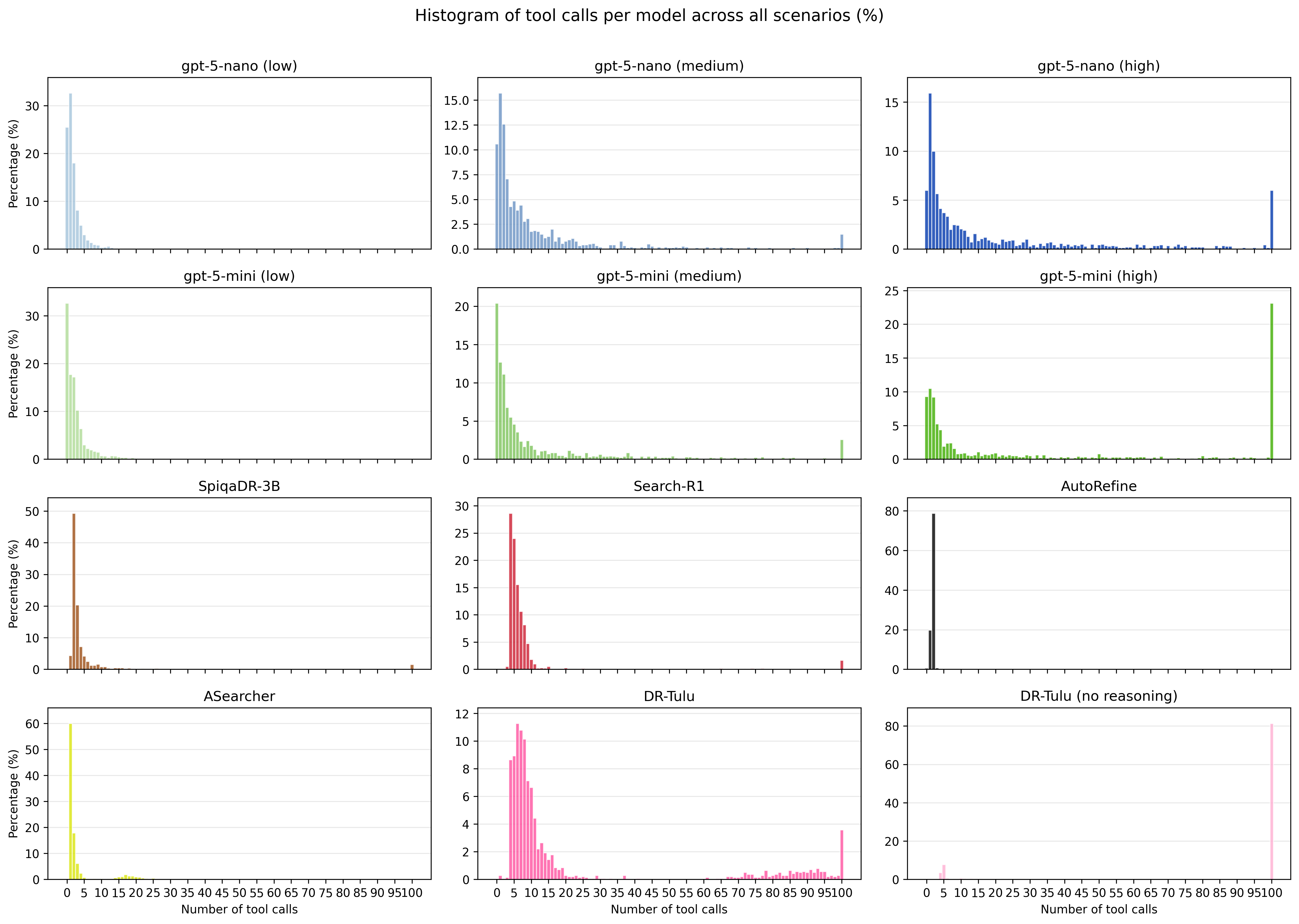}
    \caption{Tool calls distributions: \HotpotQA{}}
    \label{fig:hpqa_tool_calls_histograms}

    \vspace{1em}

    \includegraphics[width=\linewidth]{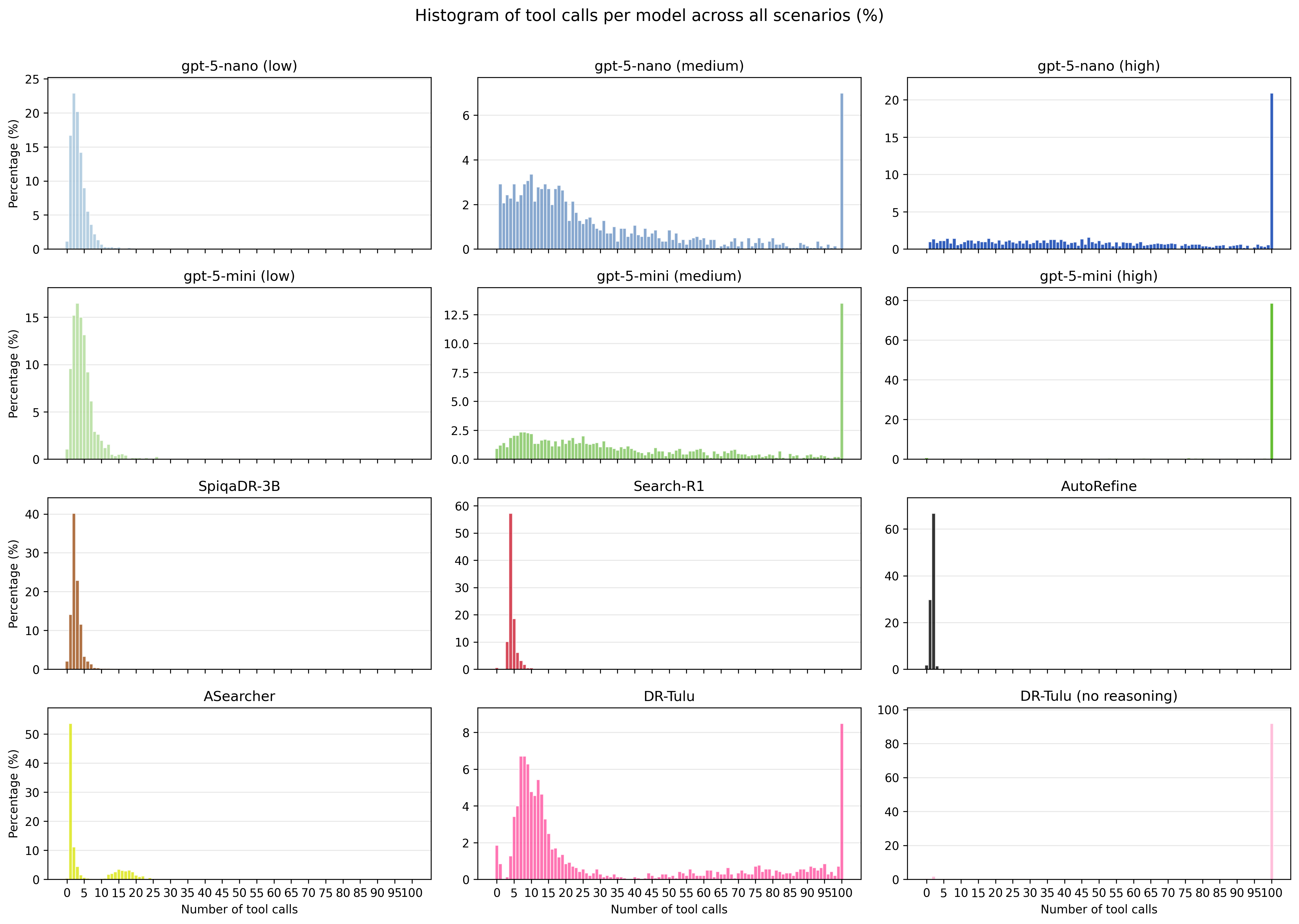}
    \caption{Tool calls distributions: \BrowseCompPlus{}}
    \label{fig:bc_tool_calls_histograms}
\end{figure*}

\subsubsection{Queries content}
\label{sec:queries_content}

Agents also differ in how they adapt the content of their search queries under degraded document conditions. We track three query-level strategies that reflect increasingly constrained or source-aware search behavior:
\begin{itemize}
    \item \textbf{Exact-Match Operator}: the query contains quoted text, indicating that the agent searches for an exact entity name, phrase, or clue span.
    \item \textbf{Specify Source}: the query targets a particular source, domain, or source family, for example using \texttt{site:} operators or naming resources such as Wikipedia, IMDb, news outlets, archives, or domain-specific databases.
    \item \textbf{Authority Cues}: the query explicitly seeks authoritative evidence, for example by using terms such as ``official'' or ``reliable source'', or by targeting institutional domains such as government or university websites.
\end{itemize}

\begin{figure*}[!h]
    \centering
    \includegraphics[width=\linewidth]{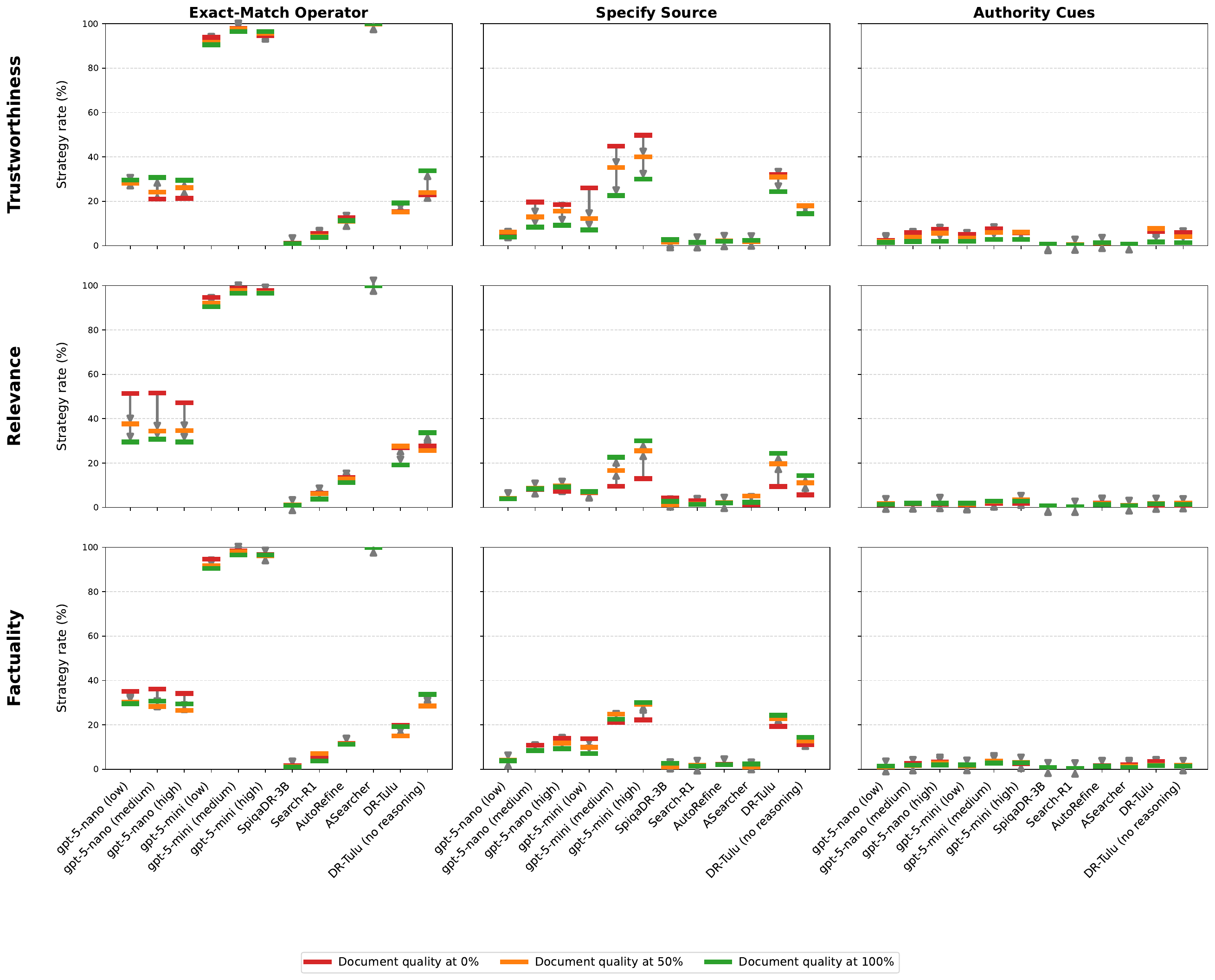}
    \caption{Evolution of query strategy usage on \BrowseCompPlus{} and \HotpotQA{} at 0\%, 50\%, and 100\% document quality.}
    \label{fig:queries}
\end{figure*}

Figure~\ref{fig:queries} shows how often these strategies appear as document quality varies. We observe substantial differences across systems and degraded dimensions.

First, systems differ strongly in their use of exact-match operators, even when document quality is 100\%. Some agents almost never use quoted queries, such as \texttt{SpiqaDR}. This is consistent with its training setup, where search is based on dense-vector retrieval rather than keyword search operators. By contrast, \texttt{ASearcher} and \texttt{gpt-5-mini}-based agents use exact-match queries much more frequently. \texttt{gpt-5-nano} behaves differently: it uses exact-match operators in only about 30\% of queries under perfect document conditions, but increases this rate to about 50\% when relevance is degraded to 0\%. This suggests that some agents respond to irrelevant results by making their queries more constrained.

Second, trustworthiness degradation leads some agents to issue more source-aware queries. GPT-based systems and \texttt{DR-Tulu} more often specify a specific source or domain when trustworthiness is low, suggesting that they try to redirect search toward more credible sources. We observe a similar, though weaker, pattern for authority cues: these systems slightly increase the use of terms such as ``official'' or ``reliable source'' under low-trust conditions. Interestingly, degrading relevance has the opposite effect for the same systems, reducing the use of source-specific queries. This suggests that agents adapt differently depending on the perceived failure mode: low trustworthiness encourages source verification, whereas low relevance encourages query reformulation around the target content.

Overall, query-content features reveal process-level differences that are not visible from final answers alone. Some agents react to degraded documents by making queries more constrained or more source-aware, while others show little adaptation. These results are exploratory, but they suggest that query reformulation is an important component of robust search-agent behavior.

\subsection{Verbalization of degradation}
\label{sec:verbalization_degradation}
We also measure whether agents explicitly notice document degradation in their available reasoning traces. We use an LLM-as-a-judge to detect whether each reasoning step verbalizes a concern about trustworthiness, relevance, or factuality. Since GPT-based agents expose only summaries of their internal reasoning, while open-weight agents expose full reasoning traces, comparisons between these families should be interpreted with caution. The evaluation prompt and hyperparameter details are provided in Appendix~\ref{sec:llm_judge_config}.

Applying this analysis to all reasoning steps would be costly, as our traces contain 335,274 steps in total. We therefore randomly sample 100 reasoning steps per system, scenario, and dataset, for a total of 15,400 steps. We do not restrict samples to steps that immediately follow a degraded document, so the resulting rate measures how often degradation concerns appear in traces, rather than a direct reaction to each retrieved document.

\begin{figure*}[t]
    \centering
    \begin{subfigure}[t]{0.49\textwidth}
        \centering
        \includegraphics[width=\linewidth]{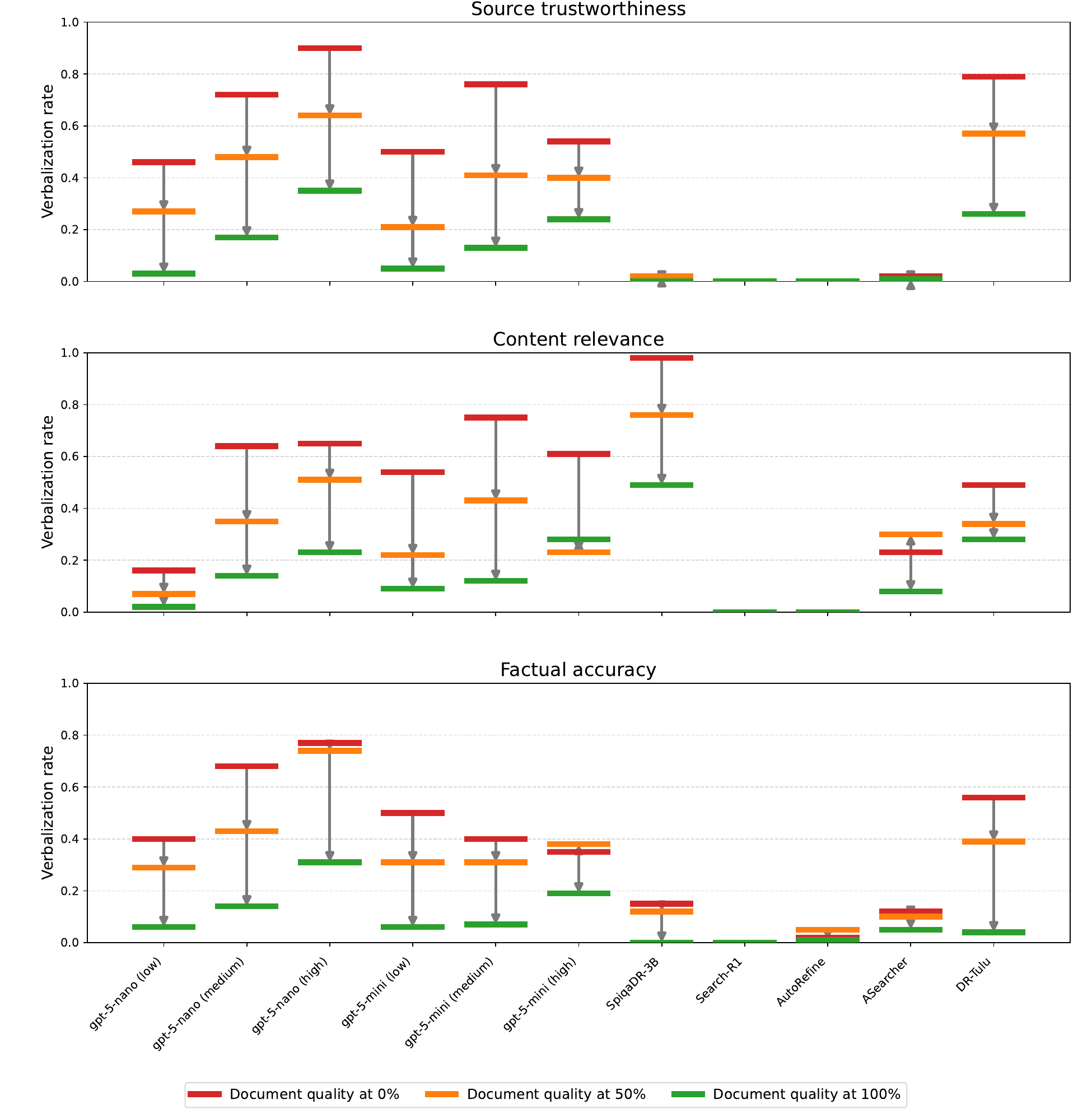}
        \caption{\HotpotQA}
        \label{fig:hpqa_verbalization}
    \end{subfigure}
    \hfill
    \begin{subfigure}[t]{0.49\textwidth}
        \centering
        \includegraphics[width=\linewidth]{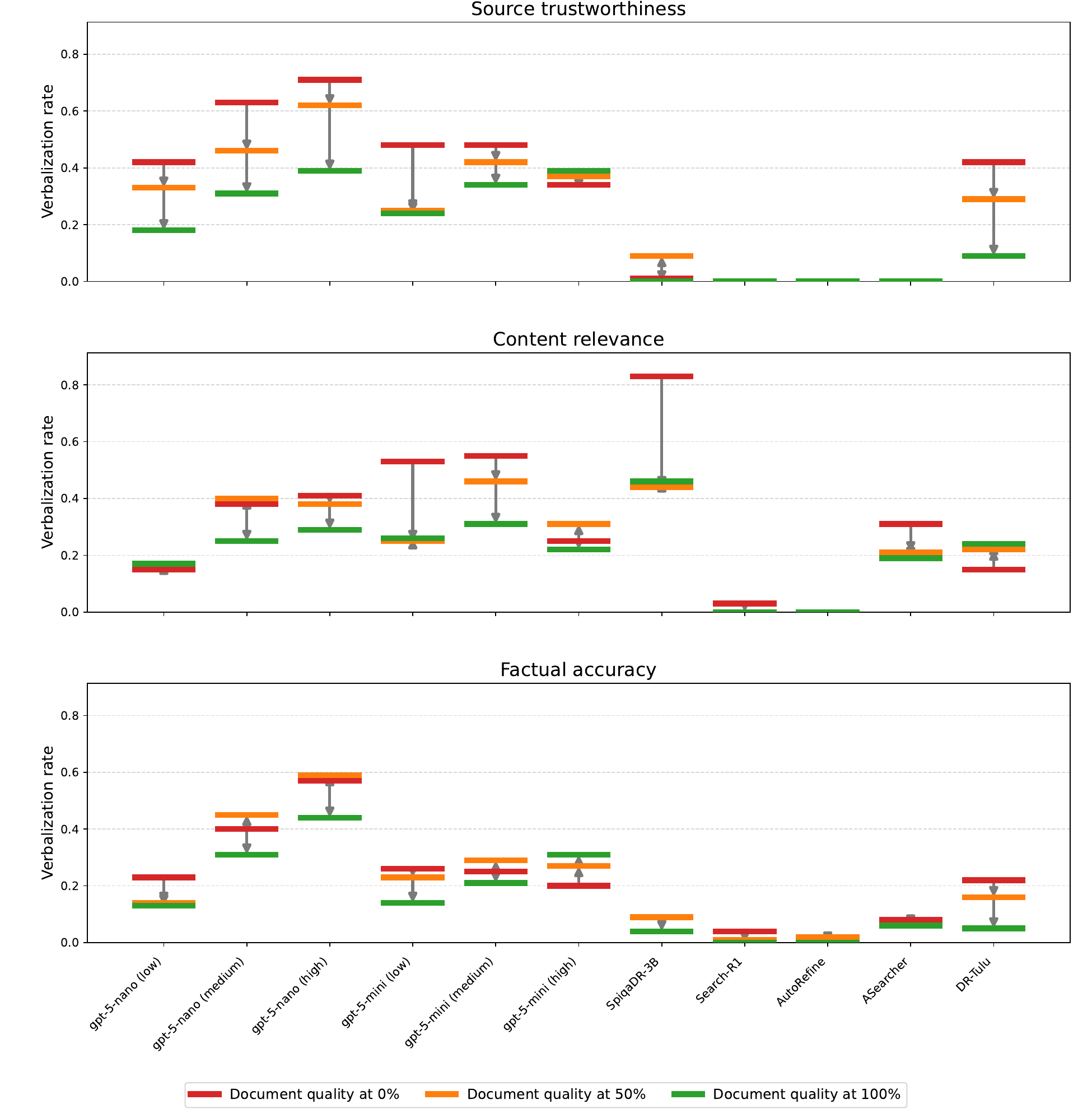}
        \caption{\BrowseCompPlus{}}
        \label{fig:bc_verbalization}
    \end{subfigure}

    \caption{Mean verbalization rate per dataset, per model, for values at 0\%, 50\% and 100\% on all axes.}
    \label{fig:verbalization}
\end{figure*}

Figure~\ref{fig:verbalization} shows large differences across systems and degradation axes. Search-R1 and AutoRefine almost never verbalize concerns about document quality. SpiqaDR and ASearcher mostly verbalize relevance issues, while rarely mentioning trustworthiness or factuality. This effect is especially strong for SpiqaDR: when relevance is set to 0\%, it flags irrelevant documents in nearly 80\% of sampled steps on \BrowseCompPlus{} and nearly 100\% on \HotpotQA. Among GPT-based agents, verbalization generally increases with reasoning effort for \texttt{GPT-5-nano}, although the pattern is less clear for \texttt{GPT-5-mini}.

Across datasets, the difference between degraded and non-degraded conditions is smaller on \BrowseCompPlus{} than on \HotpotQA. One possible explanation is that \HotpotQA{} facts are more often available in parametric knowledge, making contradictions or irrelevant evidence easier for the model to detect. However, verbalization does not clearly translate into better final answers. In particular, removing explicit reasoning from DR-Tulu changes its interaction pattern, including the number of turns, and shifts its failures from format errors toward confident incorrect answers (cf. Section~\ref{sec:system_behaviors}), but does not substantially change final-answer correctness.

\section{Additional technical details}
\subsection{Synthetic document generation}
\label{sec:synthetic_document_generation}
\subsubsection{Dynamic system prompt}
\label{sec:source_gen_sysprompt}

This section reports the prompt used to generate synthetic documents in our simulator. The prompt contains fixed instructions for query handling and output formatting, together with four dynamic placeholders. These placeholders are filled according to the sampled document attributes for the current search call: source trustworthiness, document relevance, and factuality. Colors indicate which attribute each placeholder controls: \colorbox{blue!20}{\strut blue} for source trustworthiness and \colorbox{orange!20}{\strut orange} for factuality and evidence grounding.

\begin{promptbox}[title={Prompt for synthetic document generation.}, breakable]
\label{prompt:source_gen}
\begin{Verbatim}[breaklines=true, breakanywhere=true,
  breaksymbolleft=, commandchars=\\\{\}]
You are a search engine simulator. Your goal is to generate plausible \colorbox{orange!20}{\textbf{factuality\_label}} documents, formatted in a way that would look like it came from the first paragraph of the following \colorbox{blue!20}{\textbf{source\_quality}} website: \{website\_url\}

---

Query handling:
- Only address what is explicitly asked in the query
- Do not include information that would help answer the broader question below but wasn't requested
- Treat the user message as the search query, even if it is long, broad, underspecified, or written as a full natural-language question
- Never ask the user for a more specific query or for clarification. Always generate the synthetic document using the available context and the query as written

Example:
- Full question: "What is the spot on the second biggest planet of the solar system called?"
- Query: "second biggest planet of the solar system"
- Your answer should describe the planet, but must NOT mention the spot.

Broader question the model is trying to answer: "\{question\}"

\colorbox{blue!20}{\textbf{source\_style}}
\colorbox{orange!20}{\textbf{evidence\_instructions}}

---

Formatting examples:
Bad output: **Bradley Cooper** (born January 5, 1975) is...
Bad output: [Source](https://example.com)
Bad output: Please provide a specific query.
Bad output: I need more information to generate a document.

Good output: Bradley Cooper is an American actor and filmmaker known for performances in comedy, drama, and biographical films. He studied acting before moving into television and film, later becoming known for major studio releases and critically praised roles. His career includes acting, directing, producing, and voice work, with recognition from major film awards.

Follow the good output style: one plain paragraph, no Markdown, no title line, no URLs, no refusals, no clarification requests, no meta-commentary, 80-140 words.

---

You will now generate a new synthetic document.
\end{Verbatim}
\end{promptbox}

\paragraph{Prompt placeholders.}
The dynamic placeholders used in the prompt are described in Table~\ref{tab:source_generation_prompt_placeholders}. The \colorbox{blue!20}{\texttt{source\_quality}} and \colorbox{blue!20}{\texttt{source\_style}} placeholders control the apparent credibility of the generated source, while \colorbox{orange!20}{\texttt{factuality\_label}} and \colorbox{orange!20}{\texttt{evidence\_instructions}} control whether the generated document should be factually correct or misleading.

\begin{table*}[t]
\centering
\small
\setlength{\tabcolsep}{4pt}
\renewcommand{\arraystretch}{1.15}
\begin{tabularx}{\textwidth}{p{0.22\textwidth} p{0.18\textwidth} X}
\toprule
\textbf{Placeholder} & \textbf{Condition} & \textbf{Inserted content} \\
\midrule

\colorbox{blue!20}{\texttt{source\_quality}}
& Trustworthy
& \texttt{"trustworth"} \\

& Untrustworthy
& \texttt{"junk (low quality)"} \\

\midrule

\colorbox{orange!20}{\texttt{factuality\_label}}
& Factually correct
& \textit{Empty.} \\

& Factually incorrect
& \texttt{", but factually incorrect"} \\

\midrule

\colorbox{blue!20}{\texttt{source\_style}}
& Trustworthy
& \textit{Empty.} \\

& Untrustworthy
& ``Remember you're generating a clickbaity junk website which should obviously come from an unreliable source, let's be sensationalist! OpenSources describes \texttt{\{website\_url\}} as the following: \texttt{\{opensources\_info\}}'' \\

\midrule

\colorbox{orange!20}{\texttt{evidence\_instructions}}
& Factually correct and relevant
& \textit{Relevant ground-truth context from the dataset.} \\

& Factually correct but irrelevant
& \textit{Irrelevant context from the dataset.} \\

& Factually incorrect
& ``When generating the document, you will introduce severe factual errors that would make the document completely wrong.'' \\

\bottomrule
\end{tabularx}
\caption{Placeholders used in the synthetic document generation prompt.}
\label{tab:source_generation_prompt_placeholders}
\end{table*}

The dynamic prompt construction process is depicted in the methodology described in Algorithm 1. 

\begin{algorithm}[t]
\small
\caption{Synthetic search-result generation}
\label{alg:source_generation}
\begin{algorithmic}[1]
\Require Original question $q$, agent search query $x$, support context function $G(\cdot)$, scenario probabilities $(\pi_T,\pi_R,\pi_F)$
\Ensure Synthetic search result with source URL, title, document text, and labels $(T,R,F)$
\State Sample $T \sim \mathrm{Bernoulli}(\pi_T)$ \Comment{trustworthy source}
\State Sample $R \sim \mathrm{Bernoulli}(\pi_R)$ \Comment{relevant document}
\State Sample $F \sim \mathrm{Bernoulli}(\pi_F)$ \Comment{factually correct document}

\If{$R = 1$}
    \State $q_{\mathrm{gen}} \gets q$
    \State $x_{\mathrm{gen}} \gets x$
    \State $C \gets G(q)$
\Else
    \State $q_{\mathrm{gen}} \gets$ sample an unrelated question from the dataset
    \State $C \gets G(q_{\mathrm{gen}})$
    \State $x_{\mathrm{gen}} \gets$ sample a document title from $C$
\EndIf

\If{$T = 1$}
    \State $(u, \tau) \gets$ generate a plausible reliable URL and title for $q_{\mathrm{gen}}$
\Else
    \State $u \gets$ sample a low-trust domain from the OpenSources list
    \State $\tau \gets$ generate a plausible title for $u$ and $q_{\mathrm{gen}}$
    \State Add source-quality metadata for $u$ to the prompt
\EndIf

\If{$F = 0$}
    \State $C \gets$ instruction to introduce severe factual errors
\EndIf

\State $P \gets$ system-prompt$(q_{\mathrm{gen}}, C, u, T, R, F)$
\State $d \gets \texttt{Gemma-4-31B}(\text{system}=P,\ \text{user}=x_{\mathrm{gen}})$
\State \Return search result with source URL $u$, title $\tau$, document $d$, and labels $(T,R,F)$
\end{algorithmic}
\end{algorithm}

\subsubsection{Inference}
\label{sec:annex_inference_datagen}
We use \texttt{Gemma-4 31B}\footnote{\url{https://huggingface.co/google/gemma-4-31B-it}} to generate synthetic documents. The model is served with \texttt{vLLM==0.20.1} on 8 AMD MI355X GPUs, each with 288GB of memory. This setup allows us to run several agent evaluations in parallel and to generate documents dynamically within a reasonable time. Across all experiments, we generate 809,205 synthetic documents, corresponding to approximately 104M generated tokens over about one week.

For each vLLM request, we set \texttt{max\_tokens=1024}. All other decoding parameters are left to the vLLM defaults.

\subsection{Agents implementation}
\label{sec:systems_implementation_technical_details}
\begin{SaveVerbatim}{gptTemplate}
Unknown due to API calls
\end{SaveVerbatim}

\begin{SaveVerbatim}{spiqaTemplate}
<tool_call>{tool call json}</tool_call>
<tool_result>{search result}</tool_result>
{reasoning}
...
<answer>{answer}</answer>
\end{SaveVerbatim}

\begin{SaveVerbatim}{searchRoneTemplate}
<think>{reasoning}</think>
<search>{query}</search>
<information>{search result}</information>
...
<answer>{answer}</answer>
\end{SaveVerbatim}

\begin{SaveVerbatim}{asearcherTemplate}
<think>{reasoning}</think>
<search>{query}</search>
<information>{search result}</information>
...
<answer>{answer}</answer>
\end{SaveVerbatim}

\begin{SaveVerbatim}{autoRefineTemplate}
<think>{reasoning}</think>
<search>{query}</search>
<documents>{search result}</documents>
<refine>{reasoning on refined evidence}</refine>
...
<answer>{answer}</answer>
\end{SaveVerbatim}

\begin{SaveVerbatim}{drTuluTemplate}
<think>{reasoning}</think>
<tool_call name="search">{query}</tool_call>
<tool_result>{search result}</tool_result>
...
<answer>{answer with <cite id=...>{claim}</cite>}</answer>
\end{SaveVerbatim}

This section describes the implementation details of the six families of systems evaluated in our experiments. For each family, we report the model checkpoint, training procedure when available, inference framework, decoding configuration, and tool-use format. The full orchestration code and prompts are available in our repository\footnote{Repository link to be provided after peer review.}.

Across systems, we aim to stay as close as possible to the original inference logic, prompts, and action format released by the authors. The main common modification is that the original search or retrieval backend is replaced by our simulator, which returns one controlled document per search call. All systems are evaluated with the same maximum budget of 100 tool calls. We deliberately exclude systems whose behavior depends on additional hard-coded pipeline components, such as separate summarization or post-processing modules, when these components would make it difficult to isolate the behavior of the search agent itself.

\paragraph{Tool-use formats.}
The systems use different conventions to delimit reasoning, tool calls, tool responses, and final answers. We preserve these conventions whenever possible. The tool-use formats used in our implementation are shown below.

\begin{systempromptbox}[title={SpiqaDR}, breakable]
\UseVerbatim{spiqaTemplate}
\end{systempromptbox}

\begin{systempromptbox}[title={Search-R1}, breakable]
\UseVerbatim{searchRoneTemplate}
\end{systempromptbox}

\begin{systempromptbox}[title={ASearcher}, breakable]
\UseVerbatim{asearcherTemplate}
\end{systempromptbox}

\begin{systempromptbox}[title={AutoRefine}, breakable]
\UseVerbatim{autoRefineTemplate}
\end{systempromptbox}

\begin{systempromptbox}[title={DR-Tulu}, breakable]
\UseVerbatim{drTuluTemplate}
\end{systempromptbox}

\paragraph{GPT-based systems.}
We evaluate two closed-weight OpenAI reasoning models, \texttt{openai/gpt-5-nano} and \texttt{openai/gpt-5-mini}, each with reasoning effort set to \texttt{low}, \texttt{medium}, or \texttt{high}. Since these are proprietary models, the training data and post-training procedure are not available. We use the OpenAI Responses API through \texttt{openai==2.21.0}, with \texttt{reasoning.summary=auto}, \texttt{tool\_choice=auto}, and \texttt{parallel\_tool\_calls=false}. Unlike the open-weight agents below, GPT-based systems use native API function calling rather than textual tags. At each tool call, the orchestrator samples the next document condition, generates one simulated result, appends it as a function-call output, and continues until a final answer or the tool-call budget is reached.

We use the following system prompt:
\begin{systempromptbox}[title={System prompt (GPT-5)}, breakable]
\begin{Verbatim}[breaklines=true, breakanywhere=true]
You are a Deep Research system whose goal is to search for information online in order to respond to the user question.
\end{Verbatim}
\end{systempromptbox}

\paragraph{SpiqaDR.}
We evaluate SpiqaDR, our custom Deep Research model based on Qwen2.5-3B-Instruct. Its training procedure is described in Section~\ref{sec:spiqa_training}. Inference is served locally with \texttt{sglang==0.5.6}. We use greedy decoding with \texttt{temperature=0}, a 32k-token context budget, and \texttt{max\_new\_tokens=2048} per generation step, capped by the remaining context length. Tool calls are emitted as JSON objects inside \texttt{<tool\_call>} tags, and simulated search results are returned inside \texttt{<tool\_result>} tags. The wrapper streams generation until it detects either a complete tool call or a final answer, then records the step in the shared trace format and resumes generation when needed.

We use the following system prompt:
\begin{systempromptbox}[title={System prompt (SpiqaDR)}, breakable]
\begin{Verbatim}[breaklines=true, breakanywhere=true]
You are a helpful assistant and must answer the user question. You are able to call a search tool in order to get information from the web. In order to call the search tool, format your answer this way (note that the <tool_call> </tool_call> brackets are NOT optional, you MUST write them.): 
<tool_call>
{"name": "search", "arguments": {"query": "What is the role of RL and what are the key points that make exploration-exploitation difficult?"}}
</tool_call>

It can sometimes be useful to do multiple tool call searches if you feel like you don't get the answer needed.

# Tools

You may call one or more functions to assist with the user query.

You are provided with function signatures within <tools></tools> XML tags:
<tools>
{"type": "function", "function": {"name": "search", "description": "Searches against the web", "parameters": {"type": "object", "properties": {"query": {"type": "string", "description": "The query that will be searched against the web"}}, "required": ["query"]}}}
</tools>

For each function call, return a json object with function name and arguments within <tool_call></tool_call> XML tags:
<tool_call>
{"name": <function-name>, "arguments": <args-json-object>}
</tool_call>
\end{Verbatim}
\end{systempromptbox}

\paragraph{Search-R1.}
We use \path{PeterJinGo/SearchR1-nq_hotpotqa_train-qwen2.5-7b-em-ppo-v0.3}. Search-R1 is a Qwen2.5-based search agent trained with reinforcement learning to interleave reasoning and search. The released checkpoint is trained on Natural Questions and HotpotQA, with rewards based on answer correctness and output format. We adapt the public inference script\footnote{\url{https://github.com/PeterGriffinJin/Search-R1/blob/main/infer.py}} while preserving its \texttt{<think>}, \texttt{<search>}, \texttt{<information>}, and \texttt{<answer>} protocol. Inference uses \texttt{transformers==4.57.1} and \texttt{torch==2.9.1}, with the model loaded in \texttt{bfloat16} using \texttt{device\_map=auto}. We use \texttt{temperature=0.7}, \texttt{max\_new\_tokens=1024}, and stop generation on \texttt{</search>}.

\paragraph{ASearcher-Local.}
We use \path{inclusionAI/ASearcher-Local-7B}, a Qwen2.5-7B search agent released for local knowledge-base search with RAG. In the original ASearcher evaluation code, this corresponds to the \texttt{local-rag} prompt setting.\footnote{\url{https://github.com/inclusionAI/ASearcher/blob/main/evaluation/utils.py}} In this setting, the agent issues \texttt{<search>} queries and receives search results inside \texttt{<information>} blocks. The original implementation connects these queries to a local RAG backend, whereas our implementation routes them to the same scenario-controlled simulator used for all systems.

We preserve the native ASearcher tool-use format, with \texttt{<think>}, \texttt{<search>}, \texttt{<information>}, and \texttt{<answer>} tags. Inference uses \texttt{transformers==4.57.1} and \texttt{torch==2.9.1}, with the model loaded in \texttt{bfloat16} using \texttt{device\_map=auto}. We use \texttt{temperature=0.6}, \texttt{top\_p=0.95}, \texttt{max\_new\_tokens=4096}, and a 32k-token context budget.

\paragraph{AutoRefine.}
We use \path{yrshi/AutoRefine-Qwen2.5-3B-Base}, a Qwen2.5-3B retrieval-augmented reasoning model trained with GRPO under a search-and-refine paradigm. AutoRefine introduces an explicit \texttt{<refine>} step: after receiving documents, the model is encouraged to summarize and filter the collected evidence before issuing another search or producing a final answer. This mechanism is intended to make multi-hop reasoning more robust to noisy or irrelevant information.

Our wrapper follows the public inference script\footnote{\url{https://github.com/syr-cn/AutoRefine/blob/main/infer.py}} and preserves the original tool-use format with \texttt{<think>}, \texttt{<search>}, \texttt{<documents>}, \texttt{<refine>}, and \texttt{<answer>} tags. The original implementation fills the \texttt{<documents>} block using a local retrieval server; we instead fill it with the simulated document returned by our scenario-controlled environment. Inference uses \texttt{transformers==4.57.1} and \texttt{torch==2.9.1}, with \texttt{bfloat16}, \texttt{device\_map=auto}, \texttt{temperature=0.7}, \texttt{top\_p=1.0}, \texttt{max\_new\_tokens=1024}, and a 32k-token context budget. If the search budget is exhausted, we run one final generation step without adding a new document.

\paragraph{DR-Tulu.}
We use \path{rl-research/DR-Tulu-8B}. DR-Tulu is an open deep-research model trained for long-form, citation-grounded answers with supervised fine-tuning followed by GRPO using evolving rubrics. We use \texttt{SYSTEM\_PROMPT\_V20250824} from the public DR-Tulu unified tool-calling prompt\footnote{\url{https://github.com/rlresearch/dr-tulu/blob/main/agent/dr_agent/shared_prompts/unified_tool_calling.py}}, but expose only the \textit{search} tool described in the prompt. The original system supports a richer MCP-based tool stack; our setting uses the same delimiter structure but restricts the interface to a single simulated search tool. Inference is served through a local OpenAI-compatible vLLM server with \texttt{vllm==0.16.0}. Our wrapper calls \texttt{/v1/completions} with \texttt{temperature=0}, \texttt{top\_p=1}, \texttt{top\_k=1}, \texttt{repetition\_penalty=1}, a 32k-token context budget, and per-step generations capped at 2048 tokens.
We also implement a \textit{no-reasoning} ablation using the same model, system prompt, search tool, and decoding parameters. In this variant, we add \texttt{<think>} as a stop sequence: whenever the model starts a reasoning block, the wrapper immediately appends \texttt{</think>} and resumes generation. This produces empty reasoning spans while preserving the model's native think--search/answer control flow.

\subsection{Training details about SpiqaDR}
\label{sec:spiqa_training}

\texttt{SpiqaDR} is a search-agent fine-tuned from \texttt{Qwen2.5-3B-Instruct} for deep-research-style question answering over scientific literature, using agentic reinforcement learning with \texttt{verl}. 
The checkpoint is released in the HuggingFace catalog\footnote{The link will be shared after peer review.} for reproducibility.

\paragraph{SPIQA-derived training data.}
The training data is derived from the \texttt{SPIQA}~\cite{pramanick2024spiqa} collection of machine-learning papers, from which the model takes its name. We start from the raw paper texts and enrich them with Semantic Scholar metadata using arXiv identifiers, including titles, authors, venues, years, abstracts, citations, and references.

Our clustering procedure is inspired by the citation-graph community detection approach used in MDA-QA~\citep{huang-etal-2025-towards-multi}, but we adapt the graph construction and filtering criteria to our setting. We build an undirected graph where papers are nodes, and edges connect papers that cite or reference each other within our corpus. We then apply SLPA community detection with 50 iterations and a filtering threshold of 0.1, and keep only communities containing 2 to 10 papers. This removes isolated papers, which cannot support multi-document questions, and overly large communities, which can exceed the context budget used for generation.

For each selected community, we generate a long-form QA example with \texttt{GPT-4.1}. The prompt provides the full text of all papers in the community and asks the model to identify facts specific to each document before generating an overarching question that requires information from all documents. The output contains the question, a detailed answer with sources, and document-level annotations describing each paper's contribution. The final dataset contains 1,000 examples, split into 800 training, 100 validation, and 100 test examples.

\paragraph{Search environment.}
During training, the model interacts with a search tool over the SPIQA paper collection. The tool takes a natural-language query and returns the most similar paper chunk from a vector database, with \(top\_k=1\). The returned passage is inserted into the conversation using a fixed format containing the query and the passage. When the model emits a valid tool call, generation is interrupted, the tool result is appended to the context, and generation resumes. Tool responses are masked from the policy loss, so the model is optimized only on its own generated tokens.

\paragraph{Reward function.}
The reward combines three signals: retrieval, answer quality, and format compliance. The retrieval component rewards the model for finding reference papers associated with the question.

The answer-quality component is computed with METEOR between the generated final report and the reference answer. Finally, the format component enforces the expected agent protocol: the model must use valid tool calls and produce a final answer after the tool responses. The final reward is the product of these components:
\[
R = R_{\mathrm{retrieval}} \cdot R_{\mathrm{meteor}} \cdot R_{\mathrm{format}}.
\]

This product form makes the reward strict: the model must retrieve relevant papers, produce an answer close to the reference, and follow the expected tool-use format. In practice, this was useful because optimizing only answer similarity did not reliably induce search behavior, while optimizing only retrieval did not produce useful final reports.

\paragraph{Optimization setup.}
We train the model with Reinforce++ using \texttt{verl}. The base model is \texttt{Qwen2.5-3B-Instruct}, with a context length of up to 24k tokens. Training is run for 60 epochs on a single 80GB A100/H100 GPU. During development, we also explored alternative reward variants, including rubric-based answer-quality rewards with an LLM judge. The \texttt{SpiqaDR} checkpoint used in this paper, however, is trained with the retrieval, METEOR, and format components described above.

\clearpage
\section{LLM-as-a-judge configuration}
\label{sec:llm_judge_config}
In this section, we detail the different configurations used for our LLM-as-a-judge setups. All decoding hyper-parameters are kept at their default values. We use the OpenAI Python package v2.21.0 and generate outputs through the Structured Outputs mode.

\subsection{Final-answer outcomes}
We use \texttt{GPT-4.1} as the LLM judge for final-answer outcomes.

\Needspace{25\baselineskip}
\begin{lstlisting}[style=python]
class Evaluation(BaseModel):
    reasoning: str = Field(description="""Explain why the extracted_final_answer is correct, incorrect or expresses uncertainty based on [correct_answer], focusing only on if there are meaningful differences between [correct_answer] and the extracted_final_answer or if the model clearly expresses it doesn't know, does not have enough information or asks the user for more information. Do not comment on any background to the problem, do not attempt to solve the problem, do not argue for any answer different than [correct_answer], focus only on whether the answers match""")
    answer: Literal["is_correct", "is_incorrect", "is_uncertain"] = Field(description="""Answer "is_correct" if extracted_final_answer matches the [correct_answer] given above, or is within a small margin of error for numerical problems. Answer "is_incorrect" if there if there is any inconsistency, ambiguity, non-equivalency, or if the extracted answer is incorrect. Answer "is_uncertain" if the answer says that it doesn't know, does not have enough information or asks the user for more information.""")
\end{lstlisting}

\captionof{schema}{Structured output schema used for final-answer evaluation.}
\label{schema:final_answer_evaluation}
    
\begin{systempromptbox}[title={System Prompt}, breakable]
\begin{Verbatim}[breaklines=true, breakanywhere=true]
Judge whether the following [response] to [question] is correct, incorrect or expresses uncertainty based on the precise
and unambiguous [correct_answer] below.
\end{Verbatim}
\end{systempromptbox}

\begin{systempromptbox}[title={User Prompt}, breakable]
\begin{Verbatim}[breaklines=true, breakanywhere=true]
[question]: {question}\n\n[response]: {response}

[correct_answer]: {correct_answer}
\end{Verbatim}
\end{systempromptbox}

\subsection{Verbalization}
We use \texttt{GPT-5-mini} as the LLM judge for verbalization evaluation.

\Needspace{15\baselineskip}
\begin{lstlisting}[style=python]
class ReasoningFeatures(BaseModel):
    mentions_low_trustworthiness: bool = Field(
        description="Whether or not the reasoning step mentions that the source is not trustworthy."
    )
    mentions_low_relevance: bool = Field(
        description="Whether or not the reasoning step mentions that the source does not answer the query."
    )
    mentions_low_factuality: bool = Field(
        description="Whether or not the reasoning step mentions that the source contains factual errors."
    )
\end{lstlisting}

\captionof{schema}{Structured output schema used for reasoning features annotation.}
\label{schema:final_answer_evaluation}

\begin{systempromptbox}[title={System Prompt}, breakable]
\begin{Verbatim}[breaklines=true, breakanywhere=true]
You are an evaluator whose objective is to accurately analyze certain characteristics of the reasoning of a language model that acts as an agent and performs web searches to answer questions.

More specifically, your role is to identify and highlight whether or not the reasoning steps verbalize irregularities concerning the sources used, particularly in terms of low trustworthiness (source not reliable), low relevance (source not directly related to the question), or low factuality (inaccurate, unverifiable, or misleading information).

Example of verbalizations:
Low trustworthiness: 'I see that the results came from a site that wasn't credible, this seems like nonsense!'
Low relevance: 'The search results discuss X but fail to address the actual question about Y.'
Low factuality: 'This source contains factual errors on the topic, it claims X when the correct information is Y.'
\end{Verbatim}
\end{systempromptbox}

\section{Additional evaluation breakdowns}
\label{sec:evaluation_breakdown_by_dataset}

This section provides complementary analyses to the main results. We first evaluate models without search to estimate how much each dataset can be solved from parametric knowledge alone. We then compare outcome distributions across datasets, before reporting additional reliability-aware score breakdowns over the evaluated scenarios.

\subsection{Direct-inference baselines}
\label{sec:direct_inference_results}

We first remove access to the search tool and ask each system to answer directly. This provides an estimate of how much of each dataset can be solved without external evidence. Table~\ref{tab:direct_inference_results} shows that the contrast between datasets is large: several systems answer a substantial fraction of \HotpotQA{} questions correctly, while performance on \BrowseCompPlus{} is close to zero. The lower part of the table further breaks down \HotpotQA{} performance by question difficulty and type.
\begin{table}[!t]
\centering

\small
\setlength{\tabcolsep}{4pt}

\textbf{Dataset comparison}
\vspace{0.25em}

\begin{tabular}{lcc}
\toprule
\textbf{Model} & \textbf{\HotpotQA{}} & \textbf{\BrowseCompPlus{}} \\
\midrule
gpt-5-nano (low) & 0.54 & 0.00 \\
gpt-5-nano (medium) & 0.56 & 0.00 \\
gpt-5-nano (high) & 0.54 & 0.00 \\
\midrule
gpt-5-mini (low) & 0.61 & \textbf{0.02} \\
gpt-5-mini (medium) & 0.66 & \textbf{0.02} \\
gpt-5-mini (high) & \textbf{0.69} & \textbf{0.02} \\
\midrule
SpiqaDR & 0.21 & 0.00 \\
Search-R1 & 0.32 & 0.00 \\
ASearcher & 0.33 & 0.00 \\
AutoRefine & 0.25 & 0.01 \\
DR-Tulu & 0.19 & 0.00 \\
\bottomrule
\end{tabular}

\vspace{0.9em}

\textbf{\HotpotQA{} breakdown}
\vspace{0.25em}

\scriptsize
\setlength{\tabcolsep}{3pt}
\resizebox{\columnwidth}{!}{%
\begin{tabular}{lcccccc}
\toprule
\textbf{System} & \textbf{All} & \multicolumn{3}{c}{\textbf{Question Difficulty}} & \multicolumn{2}{c}{\textbf{Question type}} \\
\cmidrule(lr){3-5} \cmidrule(lr){6-7}
 & \textbf{all} & \textbf{easy} & \textbf{medium} & \textbf{hard} & \textbf{comparison} & \textbf{bridge} \\
\midrule
gpt-5-nano (low) & 53.5\% & 71.8\% & 50.4\% & 44.1\% & 86.4\% & 44.2\% \\
gpt-5-nano (medium) & 56.0\% & 74.4\% & 54.3\% & 41.2\% & 90.9\% & 46.2\% \\
gpt-5-nano (high) & 54.5\% & 71.8\% & 51.2\% & 47.1\% & 88.6\% & 44.9\% \\
\midrule
gpt-5-mini (low) & 61.0\% & 82.1\% & 55.9\% & 55.9\% & 79.5\% & 55.8\% \\
gpt-5-mini (medium) & 65.5\% & 87.2\% & 59.8\% & 61.8\% & 86.4\% & 59.6\% \\
gpt-5-mini (high) & 68.5\% & 84.6\% & 66.1\% & 58.8\% & 88.6\% & 62.8\% \\
\midrule
SpiqaDR & 21.0\% & 20.5\% & 21.3\% & 20.6\% & 56.8\% & 10.9\% \\
Search-R1 & 32.0\% & 46.2\% & 29.1\% & 26.5\% & 72.7\% & 20.5\% \\
ASearcher & 33.0\% & 48.7\% & 30.7\% & 23.5\% & 72.7\% & 21.8\% \\
AutoRefine & 25.0\% & 33.3\% & 22.8\% & 23.5\% & 59.1\% & 15.4\% \\
DR-Tulu & 18.5\% & 35.9\% & 13.4\% & 17.6\% & 34.1\% & 14.1\% \\
\bottomrule
\end{tabular}%
}

\caption{Direct-inference baselines without search tools. The top block compares performance on \HotpotQA{} and \BrowseCompPlus{}; the bottom block breaks down \HotpotQA{} performance by question difficulty and type.}
\label{tab:direct_inference_results}
\vspace{-0.75em}
\end{table}

\subsection{Extended outcome analysis}
\label{sec:extended_outcome_analysis}
This section provides additional analyses of final-answer outcomes. 
Section~\ref{sec:confidence_intervals} reports detailed outcome distributions with confidence intervals, while Section~\ref{sec:detailed_outcomes_dataset} examines how these outcomes differ across datasets.

\subsubsection{Outcome distributions with confidence intervals}
\label{sec:confidence_intervals}

Tables~\ref{tab:browsecomp-plus--hotpotqa-detailed-outcomes-factuality} to \ref{tab:browsecomp-plus--hotpotqa-detailed-outcomes-trustworthiness} report the complete outcome distributions for every system and degradation level, together with 95\% Wilson confidence intervals.

\begin{table*}[t]
\centering
\scriptsize
\setlength{\tabcolsep}{4pt}
\begin{tabular}{llccccc}
\toprule
\textbf{Level} & \textbf{System} & \textbf{Correct} & \textbf{Abstention} & \textbf{Missing answer tag} & \textbf{Exceed tool budget} & \textbf{Incorrect} \\
\midrule
\multirow{12}{*}{\rotatebox[origin=c]{90}{\textbf{100\%}}} & gpt-5-nano (low) & $72.8\,[68.2, 76.9]$ & $7.8\,[5.5, 10.8]$ & $0.0\,[0.0, 1.0]$ & $0.0\,[0.0, 1.0]$ & $19.5\,[15.9, 23.7]$ \\
 & gpt-5-nano (medium) & $82.8\,[78.7, 86.1]$ & $2.8\,[1.5, 4.9]$ & $0.5\,[0.1, 1.8]$ & $0.2\,[0.0, 1.4]$ & $13.8\,[10.7, 17.5]$ \\
 & gpt-5-nano (high) & $79.0\,[74.7, 82.7]$ & $2.0\,[1.0, 3.9]$ & $1.8\,[0.9, 3.6]$ & $5.0\,[3.3, 7.6]$ & $12.2\,[9.4, 15.8]$ \\
 & gpt-5-mini (low) & $71.0\,[66.4, 75.2]$ & $14.0\,[10.9, 17.7]$ & $0.0\,[0.0, 1.0]$ & $0.0\,[0.0, 1.0]$ & $15.0\,[11.8, 18.8]$ \\
 & gpt-5-mini (medium) & $81.0\,[76.9, 84.5]$ & $5.0\,[3.3, 7.6]$ & $0.0\,[0.0, 1.0]$ & $2.0\,[1.0, 3.9]$ & $12.0\,[9.2, 15.6]$ \\
 & gpt-5-mini (high) & $63.2\,[58.4, 67.8]$ & $1.0\,[0.4, 2.5]$ & $0.0\,[0.0, 1.0]$ & $32.0\,[27.6, 36.7]$ & $3.8\,[2.3, 6.1]$ \\
 & SpiqaDR-3B & $61.0\,[56.1, 65.7]$ & $8.5\,[6.1, 11.6]$ & $0.0\,[0.0, 1.0]$ & $0.0\,[0.0, 1.0]$ & $30.5\,[26.2, 35.2]$ \\
 & Search-R1 & $80.2\,[76.1, 83.9]$ & $0.2\,[0.0, 1.4]$ & $0.0\,[0.0, 1.0]$ & $0.0\,[0.0, 1.0]$ & $19.5\,[15.9, 23.7]$ \\
 & AutoRefine & $67.4\,[62.7, 71.9]$ & $0.0\,[0.0, 1.0]$ & $18.2\,[14.7, 22.3]$ & $0.0\,[0.0, 1.0]$ & $14.4\,[11.3, 18.2]$ \\
 & ASearcher & $16.5\,[13.2, 20.5]$ & $0.0\,[0.0, 1.0]$ & $42.0\,[37.3, 46.9]$ & $0.0\,[0.0, 1.0]$ & $41.5\,[36.8, 46.4]$ \\
 & DR-Tulu & $59.8\,[54.9, 64.5]$ & $0.0\,[0.0, 1.0]$ & $10.8\,[8.1, 14.2]$ & $1.0\,[0.4, 2.6]$ & $28.4\,[24.2, 33.0]$ \\
 & DR-Tulu (no reasoning) & $55.1\,[50.2, 59.9]$ & $0.3\,[0.0, 1.4]$ & $3.0\,[1.7, 5.2]$ & $21.1\,[17.3, 25.3]$ & $20.6\,[16.9, 24.8]$ \\
\midrule
\multirow{12}{*}{\rotatebox[origin=c]{90}{\textbf{50\%}}} & gpt-5-nano (low) & $54.5\,[49.6, 59.3]$ & $17.0\,[13.6, 21.0]$ & $0.0\,[0.0, 1.0]$ & $0.0\,[0.0, 1.0]$ & $28.5\,[24.3, 33.1]$ \\
 & gpt-5-nano (medium) & $66.5\,[61.7, 70.9]$ & $10.8\,[8.1, 14.2]$ & $0.2\,[0.0, 1.4]$ & $0.5\,[0.1, 1.8]$ & $22.0\,[18.2, 26.3]$ \\
 & gpt-5-nano (high) & $67.0\,[62.2, 71.4]$ & $7.0\,[4.9, 9.9]$ & $2.5\,[1.4, 4.5]$ & $9.2\,[6.8, 12.5]$ & $14.2\,[11.2, 18.0]$ \\
 & gpt-5-mini (low) & $55.0\,[50.1, 59.8]$ & $26.2\,[22.2, 30.8]$ & $0.0\,[0.0, 1.0]$ & $0.0\,[0.0, 1.0]$ & $18.8\,[15.2, 22.9]$ \\
 & gpt-5-mini (medium) & $68.2\,[63.5, 72.6]$ & $8.8\,[6.4, 11.9]$ & $0.0\,[0.0, 1.0]$ & $5.5\,[3.7, 8.2]$ & $17.5\,[14.1, 21.5]$ \\
 & gpt-5-mini (high) & $40.8\,[36.0, 45.6]$ & $1.5\,[0.7, 3.2]$ & $0.0\,[0.0, 1.0]$ & $53.0\,[48.1, 57.8]$ & $4.8\,[3.1, 7.3]$ \\
 & SpiqaDR-3B & $33.5\,[29.1, 38.3]$ & $13.0\,[10.1, 16.7]$ & $0.0\,[0.0, 1.0]$ & $0.0\,[0.0, 1.0]$ & $53.5\,[48.6, 58.3]$ \\
 & Search-R1 & $51.0\,[46.1, 55.9]$ & $0.0\,[0.0, 1.0]$ & $0.0\,[0.0, 1.0]$ & $0.0\,[0.0, 1.0]$ & $49.0\,[44.1, 53.9]$ \\
 & AutoRefine & $36.0\,[31.5, 40.9]$ & $0.0\,[0.0, 1.0]$ & $22.2\,[18.4, 26.5]$ & $0.0\,[0.0, 1.0]$ & $41.8\,[37.1, 46.7]$ \\
 & ASearcher & $14.0\,[10.9, 17.7]$ & $0.0\,[0.0, 1.0]$ & $40.5\,[35.8, 45.4]$ & $0.0\,[0.0, 1.0]$ & $45.5\,[40.7, 50.4]$ \\
 & DR-Tulu & $43.9\,[39.1, 48.9]$ & $0.0\,[0.0, 1.0]$ & $15.7\,[12.4, 19.6]$ & $1.8\,[0.9, 3.6]$ & $38.6\,[34.0, 43.5]$ \\
 & DR-Tulu (no reasoning) & $39.8\,[35.2, 44.7]$ & $0.0\,[0.0, 1.0]$ & $2.8\,[1.5, 4.9]$ & $20.1\,[16.4, 24.3]$ & $37.3\,[32.7, 42.2]$ \\
\midrule
\multirow{12}{*}{\rotatebox[origin=c]{90}{\textbf{0\%}}} & gpt-5-nano (low) & $23.0\,[19.1, 27.4]$ & $31.8\,[27.4, 36.5]$ & $0.0\,[0.0, 1.0]$ & $0.0\,[0.0, 1.0]$ & $45.2\,[40.4, 50.1]$ \\
 & gpt-5-nano (medium) & $23.8\,[19.8, 28.2]$ & $30.8\,[26.4, 35.4]$ & $0.0\,[0.0, 1.0]$ & $1.5\,[0.7, 3.2]$ & $44.0\,[39.2, 48.9]$ \\
 & gpt-5-nano (high) & $22.5\,[18.7, 26.8]$ & $27.0\,[22.9, 31.6]$ & $1.5\,[0.7, 3.2]$ & $18.5\,[15.0, 22.6]$ & $30.5\,[26.2, 35.2]$ \\
 & gpt-5-mini (low) & $29.0\,[24.8, 33.6]$ & $45.0\,[40.2, 49.9]$ & $0.0\,[0.0, 1.0]$ & $0.0\,[0.0, 1.0]$ & $26.0\,[21.9, 30.5]$ \\
 & gpt-5-mini (medium) & $30.5\,[26.2, 35.2]$ & $30.2\,[26.0, 34.9]$ & $0.0\,[0.0, 1.0]$ & $6.5\,[4.5, 9.4]$ & $32.8\,[28.3, 37.5]$ \\
 & gpt-5-mini (high) & $24.2\,[20.3, 28.7]$ & $4.0\,[2.5, 6.4]$ & $0.0\,[0.0, 1.0]$ & $63.5\,[58.7, 68.1]$ & $8.2\,[5.9, 11.4]$ \\
 & SpiqaDR-3B & $4.8\,[3.1, 7.3]$ & $15.2\,[12.1, 19.1]$ & $0.2\,[0.0, 1.4]$ & $0.0\,[0.0, 1.0]$ & $79.8\,[75.5, 83.4]$ \\
 & Search-R1 & $10.5\,[7.9, 13.9]$ & $0.0\,[0.0, 1.0]$ & $0.0\,[0.0, 1.0]$ & $0.0\,[0.0, 1.0]$ & $89.5\,[86.1, 92.1]$ \\
 & AutoRefine & $6.5\,[4.5, 9.4]$ & $0.0\,[0.0, 1.0]$ & $28.5\,[24.2, 33.1]$ & $0.0\,[0.0, 1.0]$ & $65.0\,[60.2, 69.5]$ \\
 & ASearcher & $11.5\,[8.7, 15.0]$ & $0.0\,[0.0, 1.0]$ & $42.8\,[38.0, 47.6]$ & $0.0\,[0.0, 1.0]$ & $45.8\,[40.9, 50.6]$ \\
 & DR-Tulu & $11.1\,[8.4, 14.6]$ & $0.0\,[0.0, 1.0]$ & $27.3\,[23.1, 31.9]$ & $2.5\,[1.4, 4.6]$ & $59.1\,[54.2, 63.8]$ \\
 & DR-Tulu (no reasoning) & $7.3\,[5.1, 10.2]$ & $0.3\,[0.0, 1.4]$ & $3.0\,[1.7, 5.2]$ & $22.6\,[18.7, 26.9]$ & $66.9\,[62.2, 71.4]$ \\
\bottomrule
\end{tabular}%
\caption{Detailed outcome breakdown by system for \textbf{factuality}.}
\label{tab:browsecomp-plus--hotpotqa-detailed-outcomes-factuality}
\end{table*}

\begin{table*}[t]
\centering
\scriptsize
\setlength{\tabcolsep}{4pt}
\begin{tabular}{llccccc}
\toprule
\textbf{Level} & \textbf{System} & \textbf{Correct} & \textbf{Abstention} & \textbf{Missing answer tag} & \textbf{Exceed tool budget} & \textbf{Incorrect} \\
\midrule
\multirow{12}{*}{\rotatebox[origin=c]{90}{\textbf{100\%}}} & gpt-5-nano (low) & $72.8\,[68.2, 76.9]$ & $7.8\,[5.5, 10.8]$ & $0.0\,[0.0, 1.0]$ & $0.0\,[0.0, 1.0]$ & $19.5\,[15.9, 23.7]$ \\
 & gpt-5-nano (medium) & $82.8\,[78.7, 86.1]$ & $2.8\,[1.5, 4.9]$ & $0.5\,[0.1, 1.8]$ & $0.2\,[0.0, 1.4]$ & $13.8\,[10.7, 17.5]$ \\
 & gpt-5-nano (high) & $79.0\,[74.7, 82.7]$ & $2.0\,[1.0, 3.9]$ & $1.8\,[0.9, 3.6]$ & $5.0\,[3.3, 7.6]$ & $12.2\,[9.4, 15.8]$ \\
 & gpt-5-mini (low) & $71.0\,[66.4, 75.2]$ & $14.0\,[10.9, 17.7]$ & $0.0\,[0.0, 1.0]$ & $0.0\,[0.0, 1.0]$ & $15.0\,[11.8, 18.8]$ \\
 & gpt-5-mini (medium) & $81.0\,[76.9, 84.5]$ & $5.0\,[3.3, 7.6]$ & $0.0\,[0.0, 1.0]$ & $2.0\,[1.0, 3.9]$ & $12.0\,[9.2, 15.6]$ \\
 & gpt-5-mini (high) & $63.2\,[58.4, 67.8]$ & $1.0\,[0.4, 2.5]$ & $0.0\,[0.0, 1.0]$ & $32.0\,[27.6, 36.7]$ & $3.8\,[2.3, 6.1]$ \\
 & SpiqaDR-3B & $61.0\,[56.1, 65.7]$ & $8.5\,[6.1, 11.6]$ & $0.0\,[0.0, 1.0]$ & $0.0\,[0.0, 1.0]$ & $30.5\,[26.2, 35.2]$ \\
 & Search-R1 & $80.2\,[76.1, 83.9]$ & $0.2\,[0.0, 1.4]$ & $0.0\,[0.0, 1.0]$ & $0.0\,[0.0, 1.0]$ & $19.5\,[15.9, 23.7]$ \\
 & AutoRefine & $67.4\,[62.7, 71.9]$ & $0.0\,[0.0, 1.0]$ & $18.2\,[14.7, 22.3]$ & $0.0\,[0.0, 1.0]$ & $14.4\,[11.3, 18.2]$ \\
 & ASearcher & $16.5\,[13.2, 20.5]$ & $0.0\,[0.0, 1.0]$ & $42.0\,[37.3, 46.9]$ & $0.0\,[0.0, 1.0]$ & $41.5\,[36.8, 46.4]$ \\
 & DR-Tulu & $59.8\,[54.9, 64.5]$ & $0.0\,[0.0, 1.0]$ & $10.8\,[8.1, 14.2]$ & $1.0\,[0.4, 2.6]$ & $28.4\,[24.2, 33.0]$ \\
 & DR-Tulu (no reasoning) & $55.1\,[50.2, 59.9]$ & $0.3\,[0.0, 1.4]$ & $3.0\,[1.7, 5.2]$ & $21.1\,[17.3, 25.3]$ & $20.6\,[16.9, 24.8]$ \\
\midrule
\multirow{12}{*}{\rotatebox[origin=c]{90}{\textbf{50\%}}} & gpt-5-nano (low) & $67.2\,[62.5, 71.7]$ & $11.8\,[9.0, 15.3]$ & $0.2\,[0.0, 1.4]$ & $0.0\,[0.0, 1.0]$ & $20.8\,[17.1, 25.0]$ \\
 & gpt-5-nano (medium) & $79.8\,[75.5, 83.4]$ & $4.2\,[2.7, 6.7]$ & $0.5\,[0.1, 1.8]$ & $0.8\,[0.3, 2.2]$ & $14.8\,[11.6, 18.6]$ \\
 & gpt-5-nano (high) & $74.8\,[70.3, 78.8]$ & $4.8\,[3.1, 7.3]$ & $4.0\,[2.5, 6.4]$ & $6.0\,[4.1, 8.8]$ & $10.5\,[7.9, 13.9]$ \\
 & gpt-5-mini (low) & $62.7\,[57.9, 67.3]$ & $19.8\,[16.1, 23.9]$ & $0.0\,[0.0, 1.0]$ & $0.0\,[0.0, 1.0]$ & $17.5\,[14.1, 21.5]$ \\
 & gpt-5-mini (medium) & $75.8\,[71.3, 79.7]$ & $8.2\,[5.9, 11.4]$ & $0.0\,[0.0, 1.0]$ & $1.8\,[0.9, 3.6]$ & $14.2\,[11.2, 18.0]$ \\
 & gpt-5-mini (high) & $57.2\,[52.4, 62.0]$ & $0.8\,[0.3, 2.2]$ & $0.0\,[0.0, 1.0]$ & $36.0\,[31.4, 40.8]$ & $6.0\,[4.1, 8.8]$ \\
 & SpiqaDR-3B & $49.8\,[44.9, 54.6]$ & $14.5\,[11.4, 18.3]$ & $0.0\,[0.0, 1.0]$ & $0.0\,[0.0, 1.0]$ & $35.8\,[31.2, 40.6]$ \\
 & Search-R1 & $67.0\,[62.2, 71.4]$ & $0.0\,[0.0, 1.0]$ & $0.0\,[0.0, 1.0]$ & $0.0\,[0.0, 1.0]$ & $33.0\,[28.6, 37.8]$ \\
 & AutoRefine & $38.4\,[33.8, 43.3]$ & $0.0\,[0.0, 1.0]$ & $41.5\,[36.7, 46.4]$ & $0.0\,[0.0, 1.0]$ & $20.1\,[16.5, 24.3]$ \\
 & ASearcher & $14.0\,[10.9, 17.7]$ & $0.2\,[0.0, 1.4]$ & $38.8\,[34.1, 43.6]$ & $0.0\,[0.0, 1.0]$ & $47.0\,[42.2, 51.9]$ \\
 & DR-Tulu & $55.8\,[50.9, 60.6]$ & $0.0\,[0.0, 1.0]$ & $10.4\,[7.7, 13.7]$ & $0.8\,[0.3, 2.2]$ & $33.1\,[28.6, 37.9]$ \\
 & DR-Tulu (no reasoning) & $57.4\,[52.5, 62.2]$ & $0.0\,[0.0, 1.0]$ & $3.8\,[2.3, 6.1]$ & $14.3\,[11.2, 18.1]$ & $24.6\,[20.6, 29.0]$ \\
\midrule
\multirow{12}{*}{\rotatebox[origin=c]{90}{\textbf{0\%}}} & gpt-5-nano (low) & $27.3\,[23.1, 31.8]$ & $35.5\,[31.0, 40.3]$ & $0.0\,[0.0, 1.0]$ & $0.0\,[0.0, 1.0]$ & $37.2\,[32.7, 42.1]$ \\
 & gpt-5-nano (medium) & $23.2\,[19.4, 27.6]$ & $26.8\,[22.6, 31.3]$ & $1.2\,[0.5, 2.9]$ & $25.0\,[21.0, 29.5]$ & $23.8\,[19.8, 28.2]$ \\
 & gpt-5-nano (high) & $19.0\,[15.5, 23.1]$ & $25.8\,[21.7, 30.3]$ & $6.0\,[4.1, 8.8]$ & $35.2\,[30.7, 40.1]$ & $14.0\,[10.9, 17.7]$ \\
 & gpt-5-mini (low) & $31.2\,[26.9, 36.0]$ & $44.0\,[39.2, 48.9]$ & $0.0\,[0.0, 1.0]$ & $0.0\,[0.0, 1.0]$ & $24.8\,[20.8, 29.2]$ \\
 & gpt-5-mini (medium) & $31.2\,[26.9, 36.0]$ & $23.5\,[19.6, 27.9]$ & $0.0\,[0.0, 1.0]$ & $22.0\,[18.2, 26.3]$ & $23.2\,[19.4, 27.6]$ \\
 & gpt-5-mini (high) & $24.0\,[20.1, 28.4]$ & $0.8\,[0.3, 2.2]$ & $0.0\,[0.0, 1.0]$ & $70.0\,[65.3, 74.3]$ & $5.2\,[3.5, 7.9]$ \\
 & SpiqaDR-3B & $4.5\,[2.9, 7.0]$ & $38.0\,[33.4, 42.8]$ & $0.0\,[0.0, 1.0]$ & $5.2\,[3.5, 7.9]$ & $52.2\,[47.4, 57.1]$ \\
 & Search-R1 & $14.5\,[11.4, 18.3]$ & $0.0\,[0.0, 1.0]$ & $0.0\,[0.0, 1.0]$ & $6.2\,[4.3, 9.1]$ & $79.2\,[75.0, 82.9]$ \\
 & AutoRefine & $7.1\,[4.9, 10.0]$ & $0.0\,[0.0, 1.0]$ & $61.9\,[57.0, 66.5]$ & $0.0\,[0.0, 1.0]$ & $31.1\,[26.7, 35.8]$ \\
 & ASearcher & $7.0\,[4.9, 9.9]$ & $0.5\,[0.1, 1.8]$ & $49.8\,[44.9, 54.6]$ & $0.0\,[0.0, 1.0]$ & $42.8\,[38.0, 47.6]$ \\
 & DR-Tulu & $11.6\,[8.8, 15.1]$ & $3.5\,[2.1, 5.8]$ & $36.1\,[31.5, 41.0]$ & $14.9\,[11.7, 18.7]$ & $33.8\,[29.4, 38.6]$ \\
 & DR-Tulu (no reasoning) & $11.8\,[9.0, 15.3]$ & $3.0\,[1.7, 5.2]$ & $3.3\,[1.9, 5.5]$ & $8.0\,[5.7, 11.1]$ & $73.9\,[69.4, 78.0]$ \\
\bottomrule
\end{tabular}%
\caption{Detailed outcome breakdown by system for \textbf{relevance}.}
\label{tab:browsecomp-plus--hotpotqa-detailed-outcomes-relevance}
\end{table*}

\begin{table*}[t]
\centering
\scriptsize
\setlength{\tabcolsep}{4pt}
\begin{tabular}{llccccc}
\toprule
\textbf{Level} & \textbf{System} & \textbf{Correct} & \textbf{Abstention} & \textbf{Missing answer tag} & \textbf{Exceed tool budget} & \textbf{Incorrect} \\
\midrule
\multirow{12}{*}{\rotatebox[origin=c]{90}{\textbf{100\%}}} & gpt-5-nano (low) & $72.8\,[68.2, 76.9]$ & $7.8\,[5.5, 10.8]$ & $0.0\,[0.0, 1.0]$ & $0.0\,[0.0, 1.0]$ & $19.5\,[15.9, 23.7]$ \\
 & gpt-5-nano (medium) & $82.8\,[78.7, 86.1]$ & $2.8\,[1.5, 4.9]$ & $0.5\,[0.1, 1.8]$ & $0.2\,[0.0, 1.4]$ & $13.8\,[10.7, 17.5]$ \\
 & gpt-5-nano (high) & $79.0\,[74.7, 82.7]$ & $2.0\,[1.0, 3.9]$ & $1.8\,[0.9, 3.6]$ & $5.0\,[3.3, 7.6]$ & $12.2\,[9.4, 15.8]$ \\
 & gpt-5-mini (low) & $71.0\,[66.4, 75.2]$ & $14.0\,[10.9, 17.7]$ & $0.0\,[0.0, 1.0]$ & $0.0\,[0.0, 1.0]$ & $15.0\,[11.8, 18.8]$ \\
 & gpt-5-mini (medium) & $81.0\,[76.9, 84.5]$ & $5.0\,[3.3, 7.6]$ & $0.0\,[0.0, 1.0]$ & $2.0\,[1.0, 3.9]$ & $12.0\,[9.2, 15.6]$ \\
 & gpt-5-mini (high) & $63.2\,[58.4, 67.8]$ & $1.0\,[0.4, 2.5]$ & $0.0\,[0.0, 1.0]$ & $32.0\,[27.6, 36.7]$ & $3.8\,[2.3, 6.1]$ \\
 & SpiqaDR-3B & $61.0\,[56.1, 65.7]$ & $8.5\,[6.1, 11.6]$ & $0.0\,[0.0, 1.0]$ & $0.0\,[0.0, 1.0]$ & $30.5\,[26.2, 35.2]$ \\
 & Search-R1 & $80.2\,[76.1, 83.9]$ & $0.2\,[0.0, 1.4]$ & $0.0\,[0.0, 1.0]$ & $0.0\,[0.0, 1.0]$ & $19.5\,[15.9, 23.7]$ \\
 & AutoRefine & $67.4\,[62.7, 71.9]$ & $0.0\,[0.0, 1.0]$ & $18.2\,[14.7, 22.3]$ & $0.0\,[0.0, 1.0]$ & $14.4\,[11.3, 18.2]$ \\
 & ASearcher & $16.5\,[13.2, 20.5]$ & $0.0\,[0.0, 1.0]$ & $42.0\,[37.3, 46.9]$ & $0.0\,[0.0, 1.0]$ & $41.5\,[36.8, 46.4]$ \\
 & DR-Tulu & $59.8\,[54.9, 64.5]$ & $0.0\,[0.0, 1.0]$ & $10.8\,[8.1, 14.2]$ & $1.0\,[0.4, 2.6]$ & $28.4\,[24.2, 33.0]$ \\
 & DR-Tulu (no reasoning) & $55.1\,[50.2, 59.9]$ & $0.3\,[0.0, 1.4]$ & $3.0\,[1.7, 5.2]$ & $21.1\,[17.3, 25.3]$ & $20.6\,[16.9, 24.8]$ \\
\midrule
\multirow{12}{*}{\rotatebox[origin=c]{90}{\textbf{50\%}}} & gpt-5-nano (low) & $69.8\,[65.1, 74.0]$ & $11.5\,[8.7, 15.0]$ & $0.0\,[0.0, 1.0]$ & $0.0\,[0.0, 1.0]$ & $18.8\,[15.2, 22.9]$ \\
 & gpt-5-nano (medium) & $81.5\,[77.4, 85.0]$ & $3.5\,[2.1, 5.8]$ & $0.5\,[0.1, 1.8]$ & $0.2\,[0.0, 1.4]$ & $14.2\,[11.2, 18.0]$ \\
 & gpt-5-nano (high) & $77.2\,[72.9, 81.1]$ & $3.0\,[1.7, 5.2]$ & $2.0\,[1.0, 3.9]$ & $5.8\,[3.9, 8.5]$ & $12.0\,[9.2, 15.6]$ \\
 & gpt-5-mini (low) & $71.0\,[66.4, 75.2]$ & $14.2\,[11.2, 18.0]$ & $0.0\,[0.0, 1.0]$ & $0.0\,[0.0, 1.0]$ & $14.8\,[11.6, 18.6]$ \\
 & gpt-5-mini (medium) & $82.8\,[78.7, 86.1]$ & $2.8\,[1.5, 4.9]$ & $0.0\,[0.0, 1.0]$ & $2.8\,[1.5, 4.9]$ & $11.8\,[9.0, 15.3]$ \\
 & gpt-5-mini (high) & $56.0\,[51.1, 60.8]$ & $0.5\,[0.1, 1.8]$ & $0.0\,[0.0, 1.0]$ & $39.0\,[34.3, 43.9]$ & $4.5\,[2.9, 7.0]$ \\
 & SpiqaDR-3B & $60.0\,[55.1, 64.7]$ & $6.8\,[4.7, 9.6]$ & $1.0\,[0.4, 2.5]$ & $0.0\,[0.0, 1.0]$ & $32.2\,[27.9, 37.0]$ \\
 & Search-R1 & $78.2\,[73.9, 82.0]$ & $0.0\,[0.0, 1.0]$ & $0.0\,[0.0, 1.0]$ & $0.0\,[0.0, 1.0]$ & $21.8\,[18.0, 26.1]$ \\
 & AutoRefine & $66.8\,[62.0, 71.2]$ & $0.0\,[0.0, 1.0]$ & $18.1\,[14.7, 22.2]$ & $0.0\,[0.0, 1.0]$ & $15.1\,[11.9, 19.0]$ \\
 & ASearcher & $23.5\,[19.6, 27.9]$ & $0.2\,[0.0, 1.4]$ & $37.0\,[32.4, 41.8]$ & $0.0\,[0.0, 1.0]$ & $39.2\,[34.6, 44.1]$ \\
 & DR-Tulu & $56.1\,[51.1, 60.9]$ & $0.0\,[0.0, 1.0]$ & $15.2\,[12.0, 19.0]$ & $0.5\,[0.1, 1.8]$ & $28.3\,[24.1, 32.9]$ \\
 & DR-Tulu (no reasoning) & $53.6\,[48.7, 58.5]$ & $0.0\,[0.0, 1.0]$ & $2.8\,[1.5, 4.9]$ & $26.3\,[22.2, 30.8]$ & $17.3\,[13.9, 21.3]$ \\
\midrule
\multirow{12}{*}{\rotatebox[origin=c]{90}{\textbf{0\%}}} & gpt-5-nano (low) & $59.2\,[54.4, 64.0]$ & $17.5\,[14.1, 21.5]$ & $0.0\,[0.0, 1.0]$ & $0.2\,[0.0, 1.4]$ & $23.0\,[19.1, 27.4]$ \\
 & gpt-5-nano (medium) & $76.0\,[71.6, 79.9]$ & $11.8\,[9.0, 15.3]$ & $0.5\,[0.1, 1.8]$ & $1.5\,[0.7, 3.2]$ & $10.2\,[7.6, 13.6]$ \\
 & gpt-5-nano (high) & $69.5\,[64.8, 73.8]$ & $7.0\,[4.9, 9.9]$ & $2.0\,[1.0, 3.9]$ & $14.5\,[11.4, 18.3]$ & $7.0\,[4.9, 9.9]$ \\
 & gpt-5-mini (low) & $62.0\,[57.2, 66.6]$ & $22.5\,[18.7, 26.8]$ & $0.0\,[0.0, 1.0]$ & $0.0\,[0.0, 1.0]$ & $15.5\,[12.3, 19.4]$ \\
 & gpt-5-mini (medium) & $68.0\,[63.3, 72.4]$ & $4.8\,[3.1, 7.3]$ & $0.0\,[0.0, 1.0]$ & $15.8\,[12.5, 19.6]$ & $11.5\,[8.7, 15.0]$ \\
 & gpt-5-mini (high) & $33.0\,[28.6, 37.8]$ & $0.2\,[0.0, 1.4]$ & $0.0\,[0.0, 1.0]$ & $62.7\,[57.9, 67.3]$ & $4.0\,[2.5, 6.4]$ \\
 & SpiqaDR-3B & $63.7\,[58.9, 68.3]$ & $6.5\,[4.5, 9.4]$ & $0.0\,[0.0, 1.0]$ & $0.0\,[0.0, 1.0]$ & $29.8\,[25.5, 34.4]$ \\
 & Search-R1 & $79.0\,[74.7, 82.7]$ & $0.0\,[0.0, 1.0]$ & $0.0\,[0.0, 1.0]$ & $0.0\,[0.0, 1.0]$ & $21.0\,[17.3, 25.3]$ \\
 & AutoRefine & $70.2\,[65.5, 74.5]$ & $0.0\,[0.0, 1.0]$ & $13.8\,[10.7, 17.5]$ & $0.0\,[0.0, 1.0]$ & $16.0\,[12.8, 20.0]$ \\
 & ASearcher & $17.8\,[14.3, 21.8]$ & $0.0\,[0.0, 1.0]$ & $38.2\,[33.6, 43.1]$ & $0.0\,[0.0, 1.0]$ & $44.0\,[39.2, 48.9]$ \\
 & DR-Tulu & $40.9\,[36.2, 45.8]$ & $0.3\,[0.0, 1.4]$ & $34.6\,[30.1, 39.4]$ & $3.8\,[2.3, 6.2]$ & $20.5\,[16.8, 24.7]$ \\
 & DR-Tulu (no reasoning) & $41.9\,[37.1, 46.8]$ & $0.3\,[0.0, 1.4]$ & $2.8\,[1.5, 4.9]$ & $33.6\,[29.1, 38.4]$ & $21.6\,[17.8, 25.8]$ \\
\bottomrule
\end{tabular}%
\caption{Detailed outcome breakdown by system for \textbf{trustworthiness}.}
\label{tab:browsecomp-plus--hotpotqa-detailed-outcomes-trustworthiness}
\end{table*}

\subsubsection{Dataset-level analysis}
\label{sec:detailed_outcomes_dataset}
\begin{figure*}[t]
    \centering

    \includegraphics[width=\linewidth]{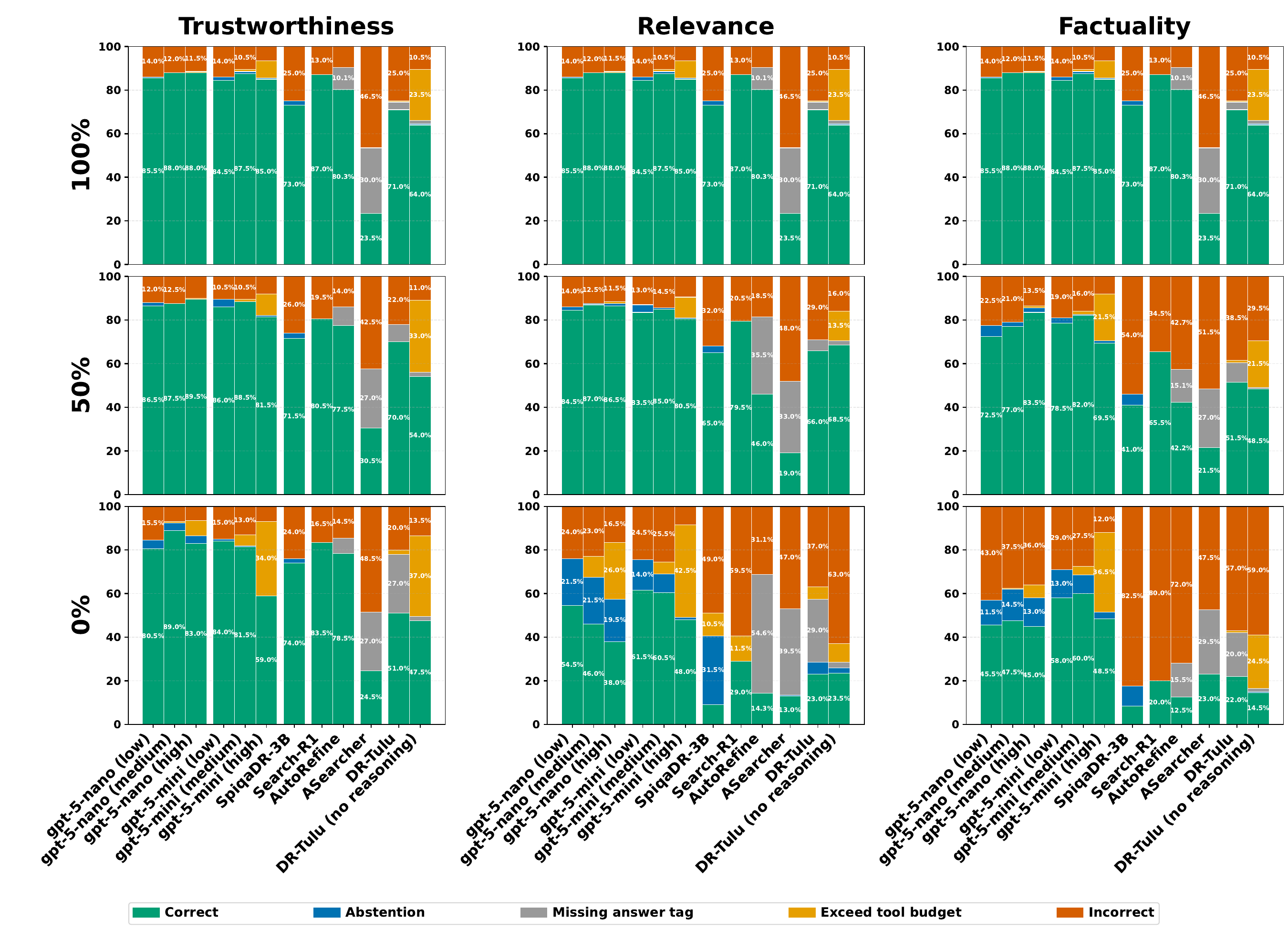}
    \caption{Detailed final-answer outcomes: \HotpotQA{}}
    \label{fig:hpqa_detailed_outcomes}

    \vspace{1em}

    \includegraphics[width=\linewidth]{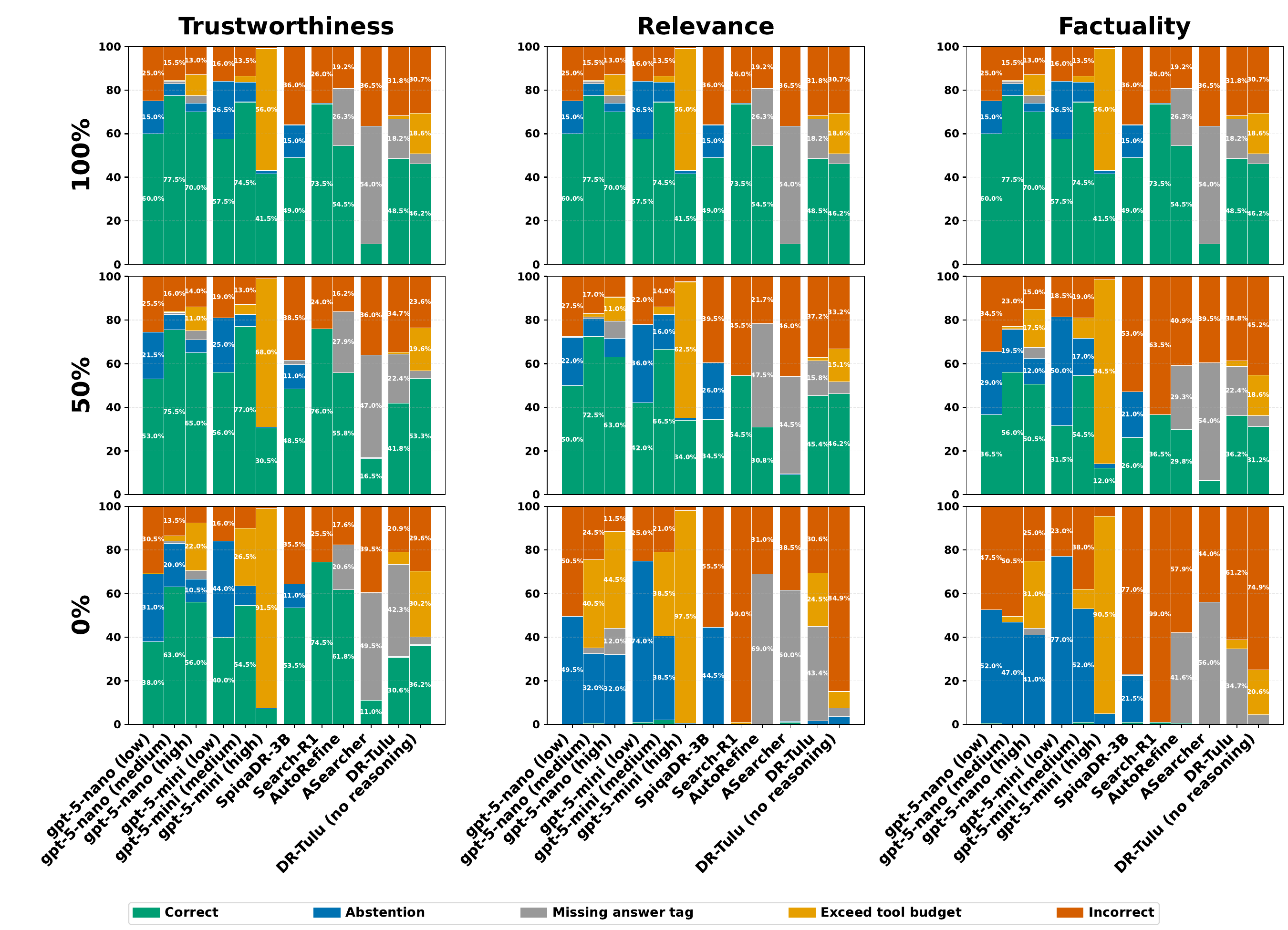}
    \caption{Detailed final-answer outcomes: \BrowseCompPlus{}}
    \label{fig:bc_detailed_outcomes}
\end{figure*}

We next analyze how final-answer outcomes vary across datasets and document-quality axes. Table~\ref{tab:average_outcome_breakdown_by_quality_cross_dataset} reports average outcomes for \HotpotQA{} and \BrowseCompPlus{} at the two extremes of each quality axis, 0\% and 100\%. Figures~\ref{fig:hpqa_detailed_outcomes} and~\ref{fig:bc_detailed_outcomes} gives the corresponding per-system breakdown across all quality levels.

\begin{table}[!htbp]
\centering
\scriptsize
\setlength{\tabcolsep}{3pt}
\renewcommand{\arraystretch}{1.05}
\begin{tabular}{llcccc}
\toprule
\textbf{Axis} & \textbf{Category} & \multicolumn{2}{c}{\textbf{0\%}} & \multicolumn{2}{c}{\textbf{100\%}} \\
\cmidrule(lr){3-4} \cmidrule(lr){5-6}
 & & \textbf{HPQA} & \textbf{BC+} & \textbf{HPQA} & \textbf{BC+} \\
\midrule
\multirow{5}{*}{Trust.}
& Correct & 69.67 & 43.84 & 55.45 & 27.11 \\
& Abstention & 2.83 & 14.08 & 6.77 & 21.02 \\
& Missing tag & 5.25 & 10.08 & 6.07 & 11.30 \\
& Tool limit & 7.12 & 14.90 & 4.93 & 12.16 \\
& Incorrect & 15.12 & 17.10 & 26.78 & 28.42 \\
\midrule
\multirow{5}{*}{Rel.}
& Correct & \cellcolor{blue!20}35.02 & \cellcolor{red!20}0.38 & 63.25 & 37.48 \\
& Abstention & 15.38 & 32.38 & 4.18 & 16.63 \\
& Missing tag & 10.47 & 15.91 & 4.53 & 9.88 \\
& Tool limit & 9.96 & 21.17 & 4.71 & 11.06 \\
& Incorrect & 29.18 & 30.17 & 23.34 & 24.95 \\
\midrule
\multirow{5}{*}{Fact.}
& Correct & \cellcolor{blue!20}33.75 & \cellcolor{red!20}0.33 & 65.48 & 39.83 \\
& Abstention & 11.00 & 33.54 & 4.73 & 15.67 \\
& Missing tag & 5.54 & 11.70 & 5.90 & 10.84 \\
& Tool limit & 6.04 & 13.14 & 5.12 & 12.03 \\
& Incorrect & 43.67 & 41.29 & 18.77 & 21.64 \\
\bottomrule
\end{tabular}
\caption{Average outcome breakdown by dataset and document-quality axis. Values are percentages. Colored cells highlight the contrast between \HotpotQA{} and \BrowseCompPlus{} at 0\% relevance and factuality.}
\label{tab:average_outcome_breakdown_by_quality_cross_dataset}
\end{table}

\subsection{Reliability-aware score}
\label{sec:additional_ras_results}

\begin{figure*}[t]
    \centering

    \begin{subfigure}[t]{0.32\textwidth}
        \centering
        \includegraphics[width=\linewidth]{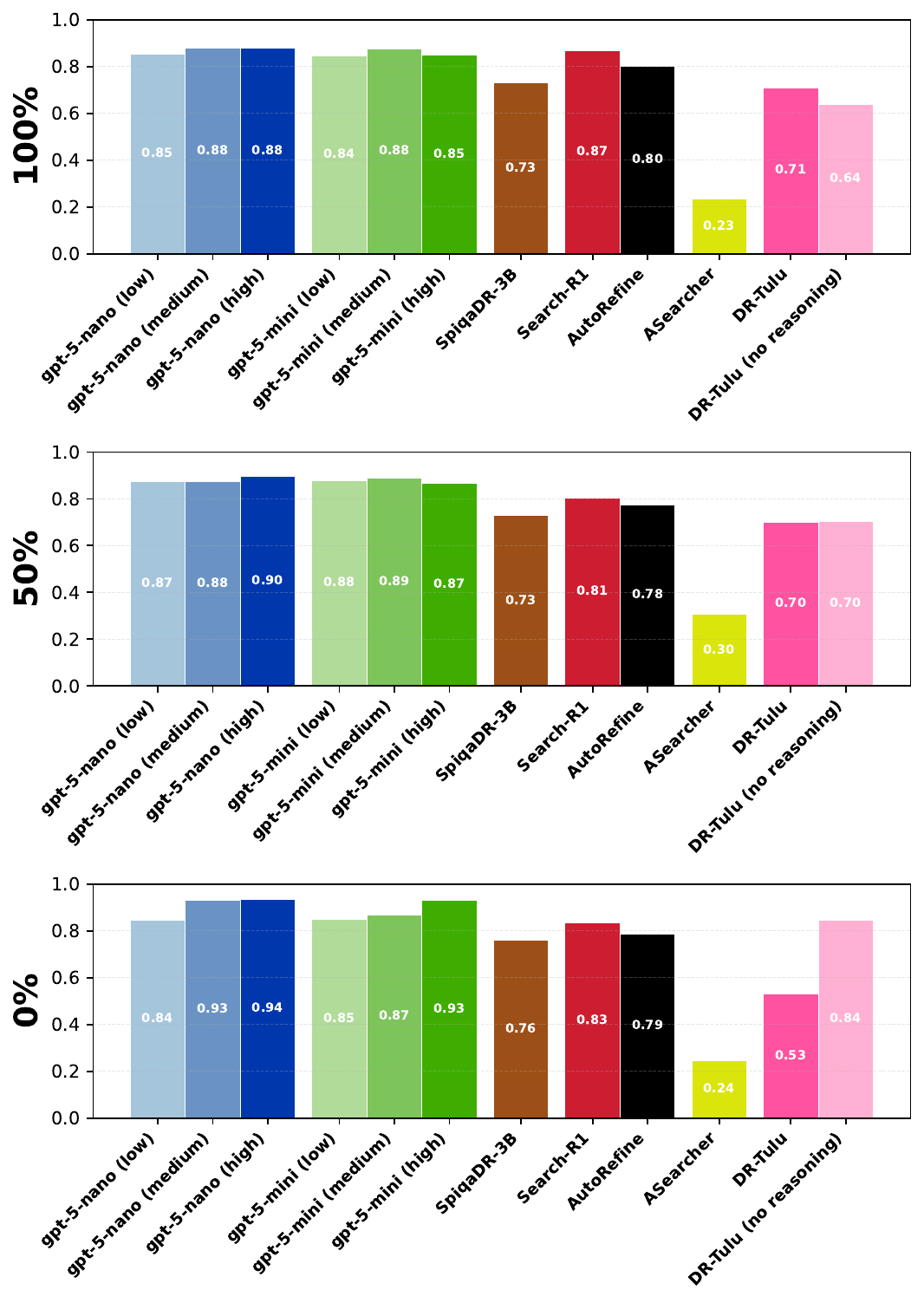}
        \caption{\HotpotQA{} -- Trustworthiness}
        \label{fig:hpqa_ras_trustworthiness}
    \end{subfigure}
    \hfill
    \begin{subfigure}[t]{0.32\textwidth}
        \centering
        \includegraphics[width=\linewidth]{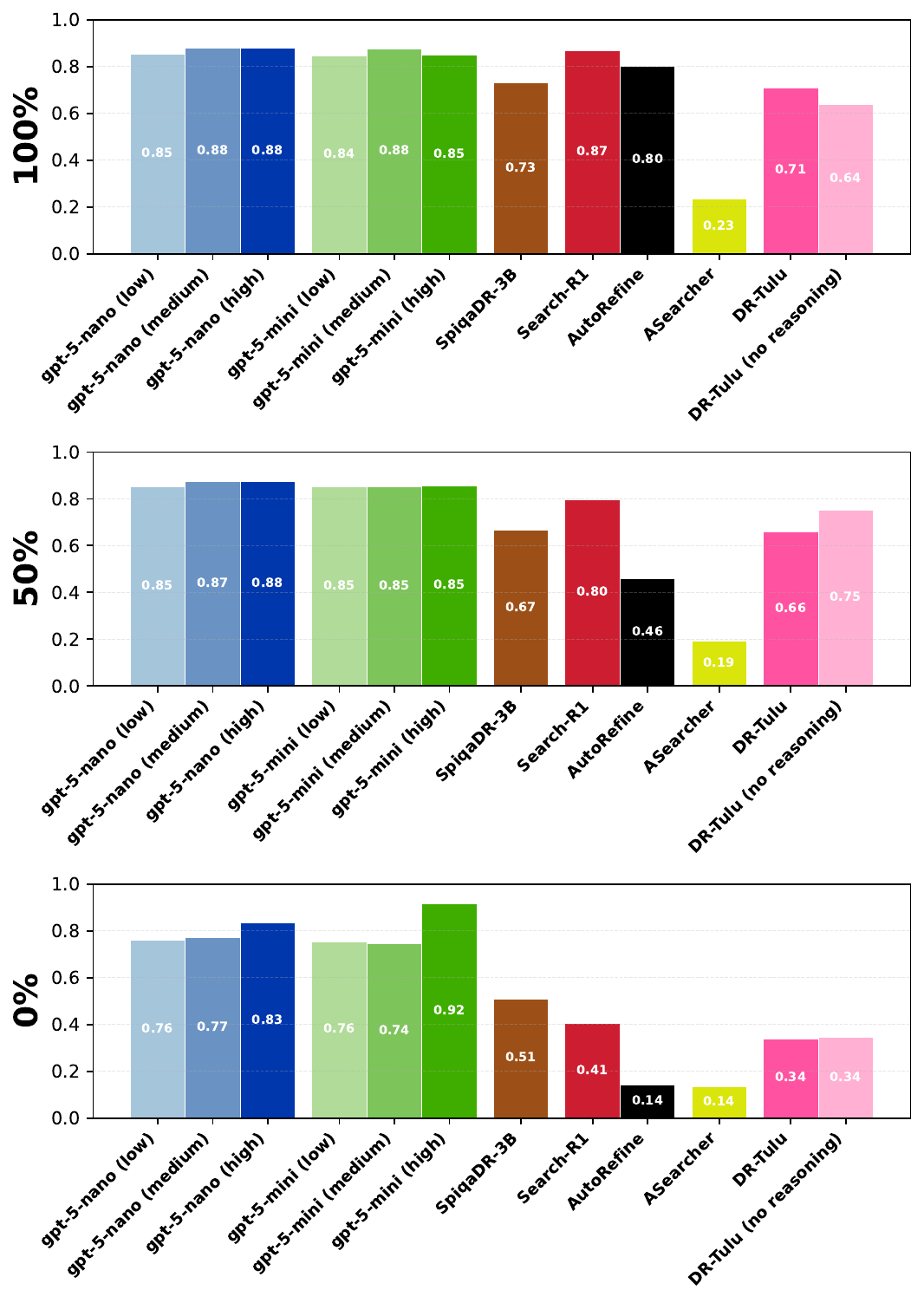}
        \caption{\HotpotQA{} -- Relevance}
        \label{fig:hpqa_ras_relevance}
    \end{subfigure}
    \hfill
    \begin{subfigure}[t]{0.32\textwidth}
        \centering
        \includegraphics[width=\linewidth]{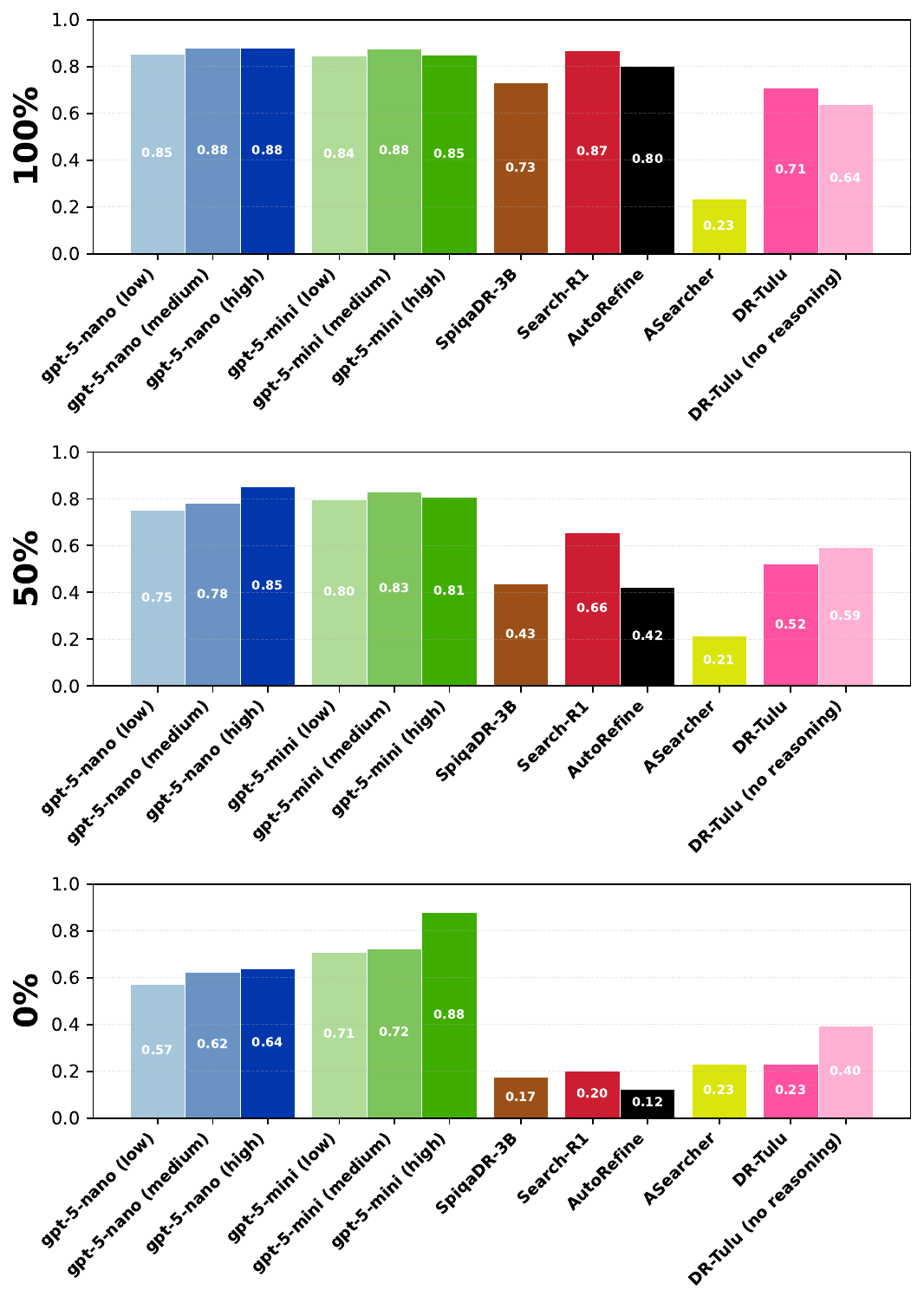}
        \caption{\HotpotQA{} -- Factuality}
        \label{fig:hpqa_ras_factuality}
    \end{subfigure}

    \vspace{2mm}

    \begin{subfigure}[t]{0.32\textwidth}
        \centering
        \includegraphics[width=\linewidth]{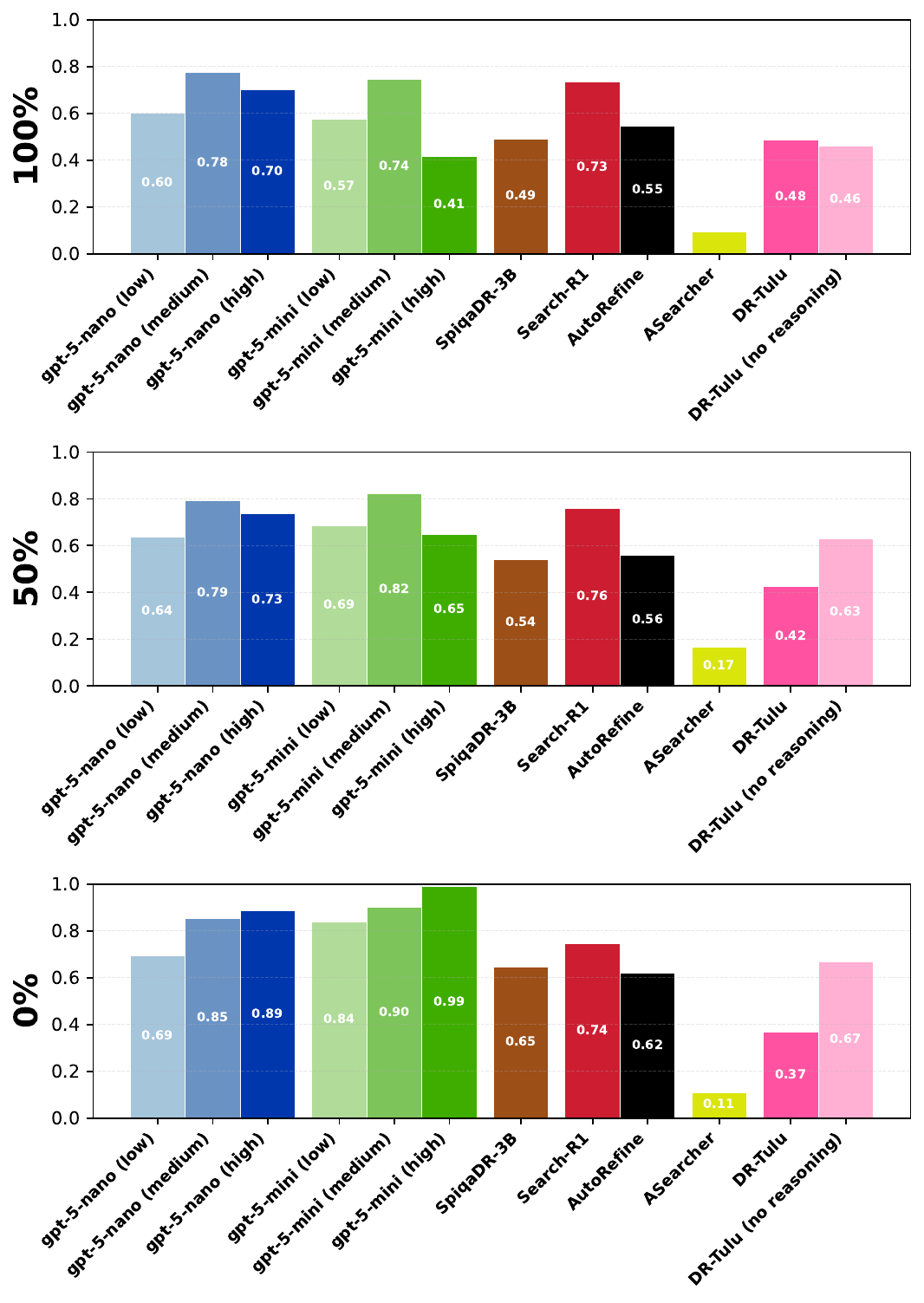}
        \caption{\BrowseCompPlus{} -- Trustworthiness}
        \label{fig:bc_ras_trustworthiness}
    \end{subfigure}
    \hfill
    \begin{subfigure}[t]{0.32\textwidth}
        \centering
        \includegraphics[width=\linewidth]{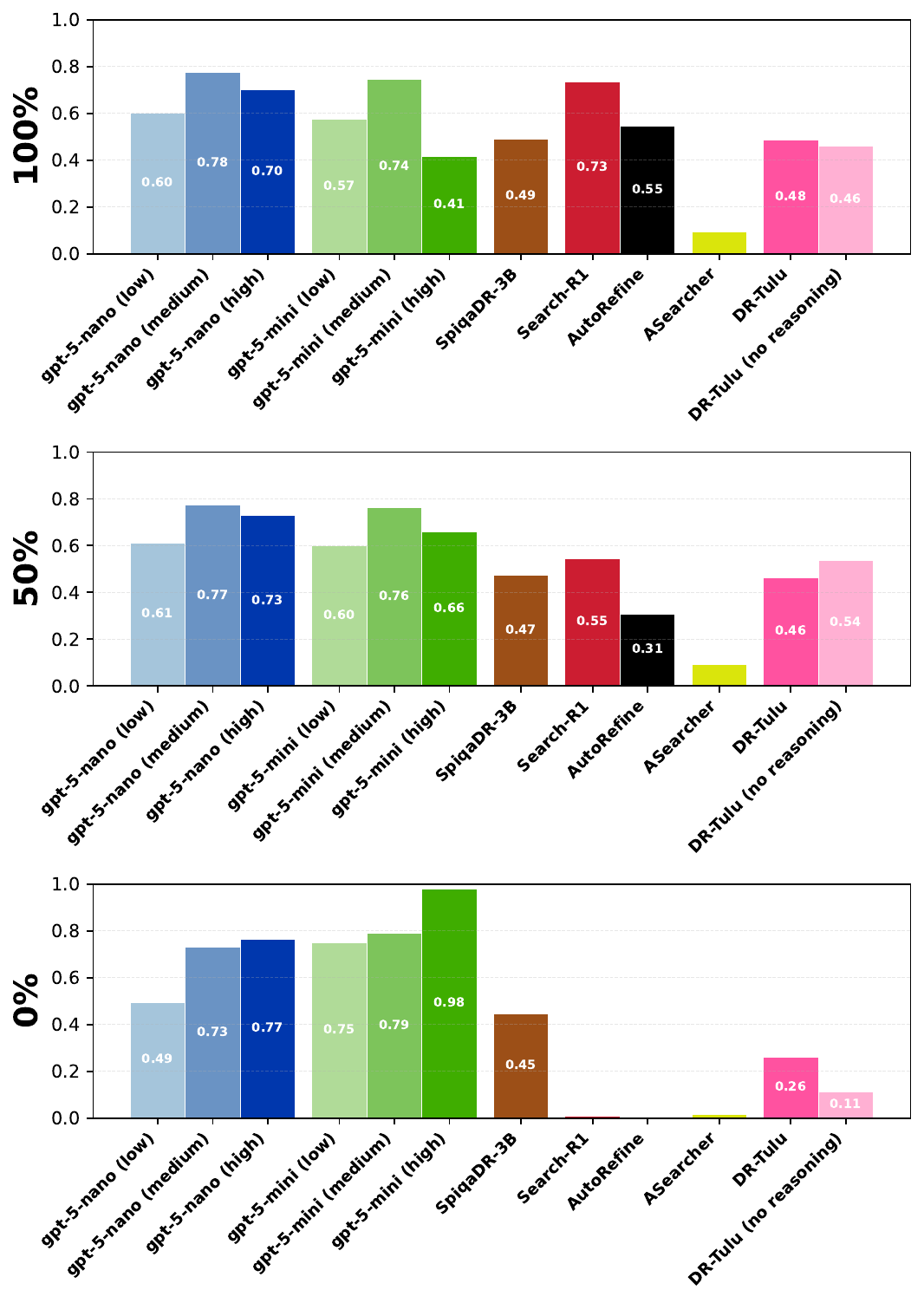}
        \caption{\BrowseCompPlus{} -- Relevance}
        \label{fig:bc_ras_relevance}
    \end{subfigure}
    \hfill
    \begin{subfigure}[t]{0.32\textwidth}
        \centering
        \includegraphics[width=\linewidth]{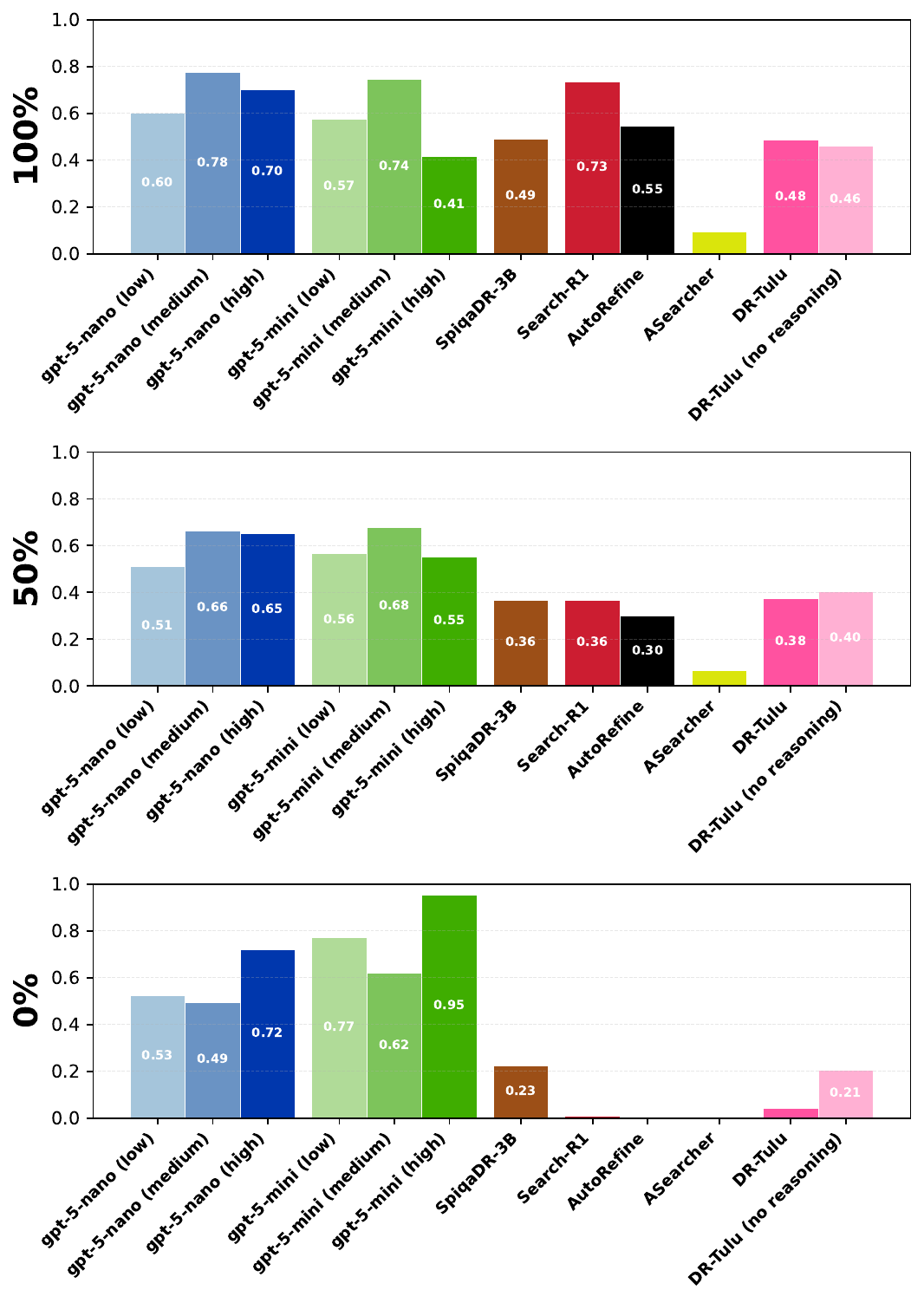}
        \caption{\BrowseCompPlus{} -- Factuality}
        \label{fig:bc_ras_factuality}
    \end{subfigure}

    \caption{Reliability-Aware Score details by system and scenario. Each subplot reports RAS for one dataset and one degraded document-quality axis.}
    \label{fig:ras_per_scenario}
\end{figure*}

Figure~\ref{fig:ras_per_scenario} breaks down RAS by dataset, system, and degraded document-quality axis. This complements the aggregate results in the main paper by showing how reliability-aware performance varies across scenarios.

\begin{table*}[t]
\centering
\scriptsize
\setlength{\tabcolsep}{2pt}
\resizebox{\textwidth}{!}{%
\begin{tabular}{lccccccccccccccc}
\toprule
\textbf{system} & \multicolumn{5}{c}{\textbf{0\%}} & \multicolumn{5}{c}{\textbf{50\%}} & \multicolumn{5}{c}{\textbf{100\%}} \\
\cmidrule(lr){2-6} \cmidrule(lr){7-11} \cmidrule(lr){12-16}
 & \textbf{Cost} & \textbf{Correct} & \textbf{RAS} & \textbf{TPC} & \textbf{CoPRAS} & \textbf{Cost} & \textbf{Correct} & \textbf{RAS} & \textbf{TPC} & \textbf{CoPRAS} & \textbf{Cost} & \textbf{Correct} & \textbf{RAS} & \textbf{TPC} & \textbf{CoPRAS} \\
\midrule
Qwen2.5-0.5B-Instruct & \textbf{0.5k} & 0.18 & 0.18 & \textbf{2.9k} & 2.8k & \textbf{0.5k} & 0.30 & 0.30 & 1.8k & 1.7k & \textbf{0.5k} & 0.43 & 0.43 & \textbf{1.2k} & \textbf{1.2k} \\
Qwen2.5-1.5B-Instruct & 0.6k & 0.15 & 0.32 & 3.7k & \textbf{1.7k} & 0.6k & 0.25 & 0.34 & 2.2k & 1.6k & 0.6k & 0.40 & 0.40 & 1.4k & 1.4k \\
Qwen2.5-3B-Instruct & 2.6k & 0.22 & 0.44 & 11.6k & 5.8k & 1.1k & 0.47 & 0.52 & 2.4k & 2.2k & 1.0k & 0.60 & 0.60 & 1.7k & 1.7k \\
Qwen2.5-7B-Instruct & 2.0k & 0.29 & 0.42 & 6.8k & 4.7k & 1.0k & 0.53 & 0.56 & 2.0k & 1.9k & 0.8k & 0.63 & 0.63 & 1.3k & 1.3k \\
Qwen2.5-14B-Instruct & 3.8k & 0.28 & 0.44 & 13.4k & 8.6k & 1.4k & 0.53 & 0.56 & 2.7k & 2.6k & 1.2k & 0.58 & 0.58 & 2.0k & 2.0k \\
Qwen2.5-32B-Instruct & 1.6k & \textbf{0.36} & \textbf{0.66} & 4.4k & 2.4k & 1.1k & \textbf{0.62} & \textbf{0.71} & \textbf{1.7k} & \textbf{1.5k} & 0.9k & \textbf{0.68} & \textbf{0.68} & 1.3k & 1.3k \\
\midrule
\textbf{Average} & 1.8k & 0.25 & 0.41 & 7.1k & 4.4k & 1.0k & 0.45 & 0.50 & 2.1k & 1.9k & 0.8k & 0.55 & 0.55 & 1.5k & 1.5k \\
\bottomrule
\end{tabular}
}
\caption{Effective reliability-aware score (RAS; higher is better), tokens per correctness (TPC; lower is better), average cost, correctness, and cost per effective RAS (CoPRAS; lower is better), aggregated by isolated document quality level. Reliability is computed per trace from the observed proportions of trustworthy, relevant, and factually correct retrieved documents.}
\label{tab:logs_gemma_effective_ras_tpc_by_quality}
\end{table*}

\begin{table*}[t]
\centering
\scriptsize
\resizebox{\textwidth}{!}{%
\begin{tabular}{lcccccccc}
\toprule
\textbf{system} & \multicolumn{2}{c}{\textbf{0\%}} & \multicolumn{2}{c}{\textbf{50\%}} & \multicolumn{2}{c}{\textbf{100\%}} & \multicolumn{2}{c}{\textbf{Average}} \\
\cmidrule(lr){2-3} \cmidrule(lr){4-5} \cmidrule(lr){6-7} \cmidrule(lr){8-9}
 & \textbf{Correct} & \textbf{RAS} & \textbf{Correct} & \textbf{RAS} & \textbf{Correct} & \textbf{RAS} & \textbf{Correct} & \textbf{RAS} \\
\midrule
Qwen2.5-0.5B-Instruct & 0.18 & 0.36 & 0.30 & 0.39 & 0.43 & 0.43 & 0.30 & 0.39 \\
Qwen2.5-1.5B-Instruct & 0.15 & 0.50 & 0.25 & 0.39 & 0.40 & 0.40 & 0.27 & 0.43 \\
Qwen2.5-3B-Instruct & 0.22 & 0.46 & 0.47 & 0.52 & 0.60 & 0.60 & 0.43 & 0.53 \\
Qwen2.5-7B-Instruct & 0.29 & 0.45 & 0.53 & 0.56 & 0.63 & 0.63 & 0.48 & 0.55 \\
Qwen2.5-14B-Instruct & 0.28 & 0.48 & 0.53 & 0.57 & 0.58 & 0.58 & 0.46 & 0.54 \\
Qwen2.5-32B-Instruct & \textbf{0.36} & \textbf{0.67} & \textbf{0.62} & \textbf{0.69} & \textbf{0.68} & \textbf{0.68} & \textbf{0.56} & \textbf{0.68} \\
\midrule
\textbf{Average} & 0.25 & 0.49 & 0.45 & 0.52 & 0.55 & 0.55 & 0.42 & 0.52 \\
\bottomrule
\end{tabular}
}
\caption{Correctness and reliability-aware score (RAS) by isolated document quality. Average is the arithmetic mean across the 0\%, 50\%, and 100\% quality columns.}
\label{tab:correct_ras_by_isolated_quality}
\end{table*}

\begin{table*}[t]
\centering
\scriptsize
\setlength{\tabcolsep}{4pt}
\resizebox{0.93\textwidth}{!}{%
\begin{tabular}{@{}lcccccc@{}}
\toprule
\textbf{System} &
\multicolumn{3}{c}{\textbf{0\%}} &
\multicolumn{3}{c}{\textbf{100\%}} \\
\cmidrule(lr){2-4} \cmidrule(lr){5-7}
& \textbf{Cost} & \textbf{RAS} & \textbf{CoPRAS}
& \textbf{Cost} & \textbf{RAS} & \textbf{CoPRAS} \\
\midrule
gpt-5-nano (low)  & 2.5k & 0.83 & \textbf{3.0k} & 1.2k & 0.73 & 1.7k \\
gpt-5-mini (high) & 55.6k & \textbf{0.96} & 58.0k & 34.3k & 0.63 & 54.3k \\
Search-R1         & 5.2k & 0.37 & 14.0k & 2.8k & \textbf{0.80} & 3.5k \\
AutoRefine        & \textbf{1.1k} & 0.28 & 3.8k & \textbf{1.1k} & 0.67 & \textbf{1.6k} \\
DR-Tulu           & 29.0k & 0.30 & 97.6k & 10.6k & 0.60 & 17.8k \\
\midrule
\textit{Average (12 systems)} & 20.5k & 0.60 & 36.9k & 10.9k & 0.66 & 18.1k \\
\bottomrule
\end{tabular}
}
\caption{Cost, RAS ($\nearrow$) and CoPRAS ($\searrow$) at 0\% and 100\% document quality.}
\label{tab:aggregate_metrics}
\end{table*}
\end{document}